\pdfoutput=1
\documentclass{article}

\usepackage{microtype}
\usepackage{graphicx}
\usepackage{booktabs}
\usepackage{multirow}
\usepackage{array}
\usepackage{enumitem}
\usepackage{xurl}
\usepackage{hyperref}

\usepackage[preprint]{icml2026}
\hypersetup{pdfsubject={Preprint}}

\usepackage{amsmath}
\usepackage{amssymb}
\usepackage{mathtools}
\usepackage{amsthm}

\theoremstyle{plain}
\newtheorem{theorem}{Theorem}
\newtheorem{proposition}{Proposition}
\newtheorem{lemma}{Lemma}
\newtheorem{corollary}{Corollary}
\theoremstyle{definition}

\theoremstyle{remark}
\newtheorem{remark}{Remark}

\newcommand{\bfsixteen}{\texttt{bf16}}
\newcommand{\fpthirtytwo}{\texttt{fp32}}
\newcommand{\bfsr}{\texttt{bf16+SR}}
\newcommand{\VerbMem}{\mathrm{VerbMem}}
\newcommand{\KnowMemF}{\mathrm{KnowMem}_f}
\newcommand{\KnowMemR}{\mathrm{KnowMem}_r}
\newcommand{\PrivLeak}{\mathrm{PrivLeak}}
\newcommand{\hwzero}{\mathrm{hw}_0}
\newcommand{\ulp}{\mathrm{ulp}}
\newcommand{\LMS}{\mathrm{LMS}}
\newcommand{\LMSnorm}{\widetilde{\mathrm{LMS}}}

\icmltitlerunning{BFloat16 Writeback Freezing Makes LLM Unlearning Fragile to Wide-Gap Four-Bit Quantization}

\begin{document}

\twocolumn[
\icmltitle{BFloat16 Writeback Freezing Makes LLM Unlearning Fragile\texorpdfstring{\\}{ }to Wide-Gap Four-Bit Quantization}

\icmlsetsymbol{corr}{*}

\begin{icmlauthorlist}
\icmlauthor{Weiye Yuan}{bjtu}
\icmlauthor{Xue Yuan}{bjtu,corr}
\end{icmlauthorlist}

\icmlaffiliation{bjtu}{School of Automation and Intelligence, Beijing Jiaotong University, Beijing, China}
\icmlcorrespondingauthor{Xue Yuan}{xyuan@bjtu.edu.cn}

\icmlkeywords{Machine Unlearning, Post-Training Quantization, Magnitude Pruning, BFloat16, Stochastic Rounding, Parameter-Movement Support}

\vskip 0.25in
]

\printAffiliationsAndNotice{\textsuperscript{*}Corresponding author. }

\begin{abstract}
Four-bit post-training quantization (PTQ) can restore supposedly forgotten text after LLM unlearning \citep{zhang2025catastrophic}.
Auditing seven public repositories reveals an upstream precision defect: both \citet{zhang2025catastrophic}'s MUSE trainer and single-GPU \texttt{open-unlearning} default to pure-\bfsixteen{} round-to-nearest (RN) storage, freezing $83\%$--$95\%$ of weights even with \fpthirtytwo{} optimizer moments and concentrating updates inside $|w_0|{\le}2^{-8}$ where wide-gap four-bit quantizers (\texttt{bnb}-FP4, per-channel GPTQ) and magnitude pruning erase them.
On MUSE-News (Llama-2-7B, $n{=}3$), switching \emph{only} writeback to stochastic rounding (\bfsr{}) eliminates severe Ep.~10 FP4 and per-channel GPTQ verbatim rebound ($30.2{\to}0.09$ and $19.0{\to}0.21$) and cuts non-saturated early-stop FP4 rebound by $92\%$ across two independent runs ($30$ neighboring checkpoints across $3$ seeds).
A single-seed same-checkpoint GPTQ sweep is consistent with median zero-cell half-width $\hwzero$ crossing $\tau{=}2^{-8}$, while strict zero-\bfsixteen{} \fpthirtytwo{} 1B surgeries establish near-zero support sufficiency alongside update-magnitude co-mediation on natural 7B runs.
Combining sufficient arithmetic bounds, $1\mathrm{B}$--$8\mathrm{B}$ rebound replications (Llama, Qwen2.5, Gemma-4), and localized $27\mathrm{B}$ freezing checks (Qwen3.8), we recommend checking trained-layer weight movement ($f_{\mathrm{move}}^{\mathrm{tr}}$) before jointly tuning writeback precision and learning rate.
\end{abstract}

\begin{figure*}[t]
  \centering
  \includegraphics[width=0.86\linewidth]{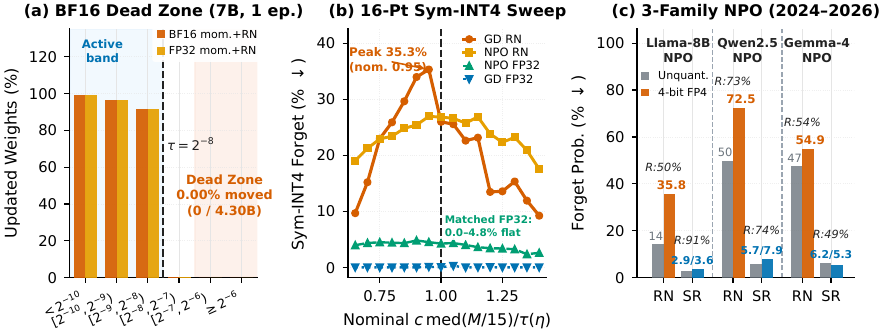}
  \caption{\textbf{Pure-\bfsixteen{} RN writeback freezing drives wide-gap four-bit rebound.}
  (a)~Llama-2-7B (1~ep., 51 steps, $\eta{=}10^{-5}$): both \bfsixteen{} and \fpthirtytwo{} moments leave $100\%$ of $|w_0|\ge 2^{-7}$ weights unchanged (in the profiled \fpthirtytwo{}-moment arm at Ep.~10, $<0.005\%$ move in $[2^{-7},2^{-6})$).
  (b)~Single-seed 16-point Sym-INT4 ($g{=}128$, matched $s{=}cM/7.0$) sweeps on TOFU-1B indexed by nominal $r_{\mathrm{nom}}{=}c\,\mathrm{med}(M/15)/\tau$ (actual $\mathrm{med}(\hwzero/\tau){=}\tfrac{15}{14}r_{\mathrm{nom}}$, placing the $0.95$ peak at $1.02$) while matched \fpthirtytwo{} controls stay flat ($0.0\%$--$4.8\%$, App.~\ref{app:details}).
  (c)~Across three model families under NPO (Llama-3.1-8B, Qwen2.5-1.5B, 2026 Gemma-4-E4B), unbiased \bfsr{} eliminates severe RN post-FP4 rebound (leaving $+0.7$, $+2.2$, and $-0.9$\,pp residuals).}
  \label{fig:hero}
\end{figure*}

\section{Introduction}
\label{sec:intro}

Machine unlearning aims to remove target knowledge from LLMs without full retraining \citep{maini2024tofu,shi2024muse,zhang2024npo}.
Yet \citet{zhang2025catastrophic} found that four-bit post-training quantization (PTQ) can restore forgotten text while eight-bit does not, spurring objective- and architecture-level defenses \citep{mishra2026quail,abitante2026lora,sadhu2026mansu,khan2026durableun,wang2026landscape}.

We isolate a simpler upstream precision defect inside the unlearning trainers themselves.
Small dense updates normally survive coarse quantization because diverse weight phases act as self-dither \citep{schuchman1964dither,gray1993dithered}.
Auditing seven public unlearning repositories (five distinct trainers; Table~\ref{tab:ecosystem_audit}) reveals that both \citet{zhang2025catastrophic}'s MUSE codebase (\texttt{10131ae25f}) and single-GPU \texttt{open-unlearning} default to pure-\bfsixteen{} round-to-nearest (RN) weight storage.
Because RN absorbs steps below half a local ULP, pure-\bfsixteen{} training freezes $83\%$--$95\%$ of trained weights even with \fpthirtytwo{} AdamW moments \citep{micikevicius2018mixed,weiss2026bf16}.
Updates pile up near zero, precisely where wide-gap four-bit quantizers (\texttt{bnb}-FP4, per-channel GPTQ) and magnitude pruning erase them.

\paragraph{Terminology and three distinct empirical outcomes.}
Throughout this paper, we reserve \emph{snap-back} strictly for weight-level quantization erasure (Theorem~\ref{thm:snapback}) and use \emph{rebound} for post-compression behavioral forgetting loss (measured by $\VerbMem$ on MUSE, Forget Prob.\ on TOFU, and Cyber accuracy on WMDP).
Crucially, pure-\bfsixteen{} RN freezing manifests in three distinct practical outcomes that must be separated:
(i)~\emph{pre-PTQ training stall}, where freezing prevents unlearning even before compression (e.g., Qwen2.5, Gemma-4 GradDiff, Qwen3.8-27B, low-LR WMDP);
(ii)~\emph{post-compression rebound}, where unquantized forgetting succeeds via near-zero weights that wide-gap PTQ or pruning erases (e.g., MUSE-7B, Llama-8B, TOFU-1B, Gemma-4 NPO); and
(iii)~\emph{fixed-LR utility/privacy trade-offs}, where restoring unbiased writeback removes RN's implicit shrinkage and can over-forget unless the learning rate or stopping point is re-tuned.

\paragraph{Contributions.}
We isolate an upstream precision confounder in LLM unlearning PTQ robustness and provide a practical workflow:
(1)~\emph{Audit \& writeback isolation (\S\ref{sec:audit}--\ref{sec:causal}):} auditing seven repositories and running matched MUSE-News ablations ($n{=}3$) isolates writeback rounding---not moments---as driving Ep.~10 FP4/per-channel GPTQ rebound, two independent non-saturated early-stop runs ($30$ neighboring steps), and a threshold-compatible single-seed GPTQ sweep.
(2)~\emph{Arithmetic bounds \& zero-\bfsixteen{} surgeries (\S\ref{sec:mechanism}, \S\ref{sec:surgeries}):} we derive sufficient snap-back bounds (noting limited natural coverage and non-coverage of GPTQ) and exact FP4 cluster projection, and show via strict \fpthirtytwo{} 1B surgeries that near-zero support suffices for FP4/pruning rebound while 7B runs show magnitude co-mediation.
(3)~\emph{Cross-scale checks \& protocol (\S\ref{sec:ecosystem}):} across $1\mathrm{B}$--$8\mathrm{B}$ rebound replications and localized $27\mathrm{B}$ freezing checks, we separate freezing from rebound and fixed-LR trade-offs, recommending an $f_{\mathrm{move}}^{\mathrm{tr}}$ check before jointly tuning writeback precision and LR.

\begin{table*}[t]
\centering
\caption{\textbf{Precision audit summary of seven public LLM unlearning repositories (five distinct trainers).}
Five of seven default to pure-\bfsixteen{} RN on a single GPU (all seven if DeepSpeed is omitted on \texttt{tofu}/\texttt{npo}); \texttt{open-unlearning} switches between 1-GPU RN and multi-GPU ZeRO-3 \fpthirtytwo{} masters via the launcher alone. \textbf{[Meas.]} executes the pinned repo; [Repro.] reproduces its default config; \emph{[Inf.]} infers from shared/inspected source. Unchanged percentages measure endpoint non-movement. Full commits and \texttt{file:line} references appear in Appendix Table~\ref{tab:app_audit_full}.}
\label{tab:ecosystem_audit}
\footnotesize
\renewcommand{\arraystretch}{1.05}
\setlength{\tabcolsep}{2.4pt}
\resizebox{\textwidth}{!}{%
\begin{tabular}{lllll}
\toprule
\textbf{Trainer Family (Repos)} & \textbf{Load / Mom.} & \textbf{Default $\eta$} & \textbf{Default Storage} & \textbf{Unchanged (Endpoint)} \\
\midrule
MUSE \texttt{IterativeUnlearner} (3) & \texttt{bf16} / \texttt{bf16} & $10^{-5}$ & Pure BF16 RN & \textbf{82.9\%} \textbf{[Meas., 7B]} (2 repos \emph{[Inf.]}) \\
\texttt{open-unlearning} (1G / DS-Z3) & \texttt{bf16} / \texttt{paged32} & $10^{-5}$+dec. & RN (1G) / FP32 (DS) & \textbf{88--95\%} [Repro.] / No stall \emph{[Inf.]} \\
WMDP (\texttt{Unlearn-Simple} / \texttt{wmdp}) & \texttt{bf16} / \texttt{bf16} & $10^{-5}$ / $5{\times}10^{-5}$ & Pure BF16 RN & 21.8\% / 1.3\% \textbf{[Meas., L5--7]} \\
\texttt{tofu} / \texttt{npo} (DS \texttt{forget.py}) & \texttt{bf16} / \texttt{paged32} & $10^{-5}$ & FP32 Master (DS) & No BF16 stall \emph{[Inf.]} (RN w/o DS) \\
\bottomrule
\end{tabular}}
\end{table*}

\section{Mechanism: Movement Support, Snap-Back, and Alignment}
\label{sec:mechanism}

Let $W_0 \in \mathbb{R}^d$ denote the flattened 2D linear weights of a target model, $W_u = W_0 + \Delta W$ the unlearned weights, and $Q(\cdot)$ a PTQ operator producing $\hat{W} = Q(W)$.
Define the quantized update $\Delta Q \triangleq Q(W_u) - Q(W_0)$ and projection ratio $\rho \triangleq \langle \Delta Q, \Delta W \rangle / \|\Delta W\|_2^2$ (proofs in Appendix~\ref{app:proofs}).

\subsection{Why Small Dense Updates Survive Coarse Quantization}
\label{sec:dither}

Under the heuristic modeling assumption of an unbounded uniform scalar quantizer $Q$ (step $\Delta_{\mathrm{q}}$) with uniform, update-independent sub-cell weight phases $u \in [0,1)$, small updates $|\Delta w| < \Delta_{\mathrm{q}}$ satisfy $\mathbb{E}_{u}[Q(w_0{+}\Delta w){-}Q(w_0)] = \Delta w$ \citep{schuchman1964dither,gray1993dithered}.
Across $d_{\mathrm{eff}}$ active coordinates with conditionally independent phases and comparable contributions, the directional signal-to-noise ratio along a probe $g \in \mathbb{R}^d$ aligned with $\Delta W$ scales as $\sqrt{d_{\mathrm{eff}}|\Delta w|/\Delta_{\mathrm{q}}} \gg 1$ even when $|\Delta w| \ll \Delta_{\mathrm{q}}$ (Appendix Lemma~\ref{prop:dither}, Corollary~\ref{cor:snr}).
This distributed self-dither fails when updates collapse inside a single near-zero quantization cell.

\subsection{The BFloat16 Dead Zone and Exact Group-Wise Snap-Back}
\label{sec:tau_eta}

In standard \bfsixteen{} (7 explicit mantissa bits, truncated binary32), any finite interior normal number $w$ with $2^e \le |w| < 2^{e+1}$ ($e \in [-125, 126]$) has outward spacing $\ulp^+(w) = 2^{e-7}$, while at $|w| = 2^e$ inward spacing halves to $\ulp^-(2^e) = 2^{e-8}$.
Under round-to-nearest-even (RN), adding an effective pre-writeback \fpthirtytwo{} candidate shift $\delta w_t \triangleq \mathrm{fl}_{32}(w_t + \Delta^{\mathrm{cand}}_t) - w_t$ leaves $w_t$ unchanged whenever $|\delta w_t| < \tfrac{1}{2}\ulp_{\mathrm{dir}}(w_t)$.
For AdamW with learning rate $\eta_t$ and normalized moment ratio $\kappa_t = |\hat{m}_t / (\sqrt{\hat{v}_t} + \epsilon)|$, $|\delta w_t| \approx \eta_t \kappa_t$ when weight decay is small.

\begin{proposition}[Directional \bfsixteen{} RN stall criterion and cutoff $\tau(B)$]
\label{prop:tau_eta}
Under pure-\bfsixteen{} storage with one deterministic RN writeback per step over finite normal \bfsixteen{} values ($k \in [-125, 126]$; $[-\tau,+\tau]$ invariance also covers $0$/subnormals):
(1)~Coordinate $i$ stalls at step $t$ whenever $|\delta w_{t,i}| < \tfrac{1}{2}\ulp_{\mathrm{dir}}(w_{t,i})$.
(2)~For $\tau = 2^k$ and $\tau_+ \triangleq \tau + \ulp^+(\tau) = 2^k(1 + 2^{-7})$, if $|\delta w_{t,i}| < \tfrac{1}{2}\ulp^+(\tau) = 2^{k-8}$ holds for all $t \in \{1,\dots,T\}$, then every coordinate with $|w_{0,i}| \ge \tau_+$ freezes permanently ($w_{T,i} = w_{0,i}$), while $[-\tau, +\tau]$ is forward-invariant ($|w_{0,i}| \le \tau \Rightarrow |w_{T,i}| \le \tau$).
(3)~Given a uniform trajectory step bound $|\delta w_{t,i}| \le B$ ($B > 0$) with $k = \lfloor \log_2(256 B) \rfloor + 1 \in [-125,126]$, strict inequality $2^{k-8} > B$ yields
\begin{equation}
  \tau(B) \triangleq 2^{\lfloor \log_2(256 B) \rfloor + 1}.
  \label{eq:tau_eta}
\end{equation}
This final writeback criterion applies identically whether optimizer moments $(m_t, v_t)$ use \bfsixteen{} or \fpthirtytwo{}.
\end{proposition}

Because of the discrete exponent floor in Eq.~\eqref{eq:tau_eta}, the empirical scaling proxy $\tau_{\mathrm{emp}}(\eta) \triangleq \tau(\kappa\eta)$ stays constant across $\kappa \in [0.763, 1.525]$ at $\eta \in \{5{\times}10^{-6}, 10^{-5}, 2{\times}10^{-5}\}$ ($\tau_{\mathrm{emp}} \in \{2^{-9}, 2^{-8}, 2^{-7}\}$; Table~\ref{tab:app_kappa} logs $\kappa_t \le 1.29$, whereas a worst-case Adam bound gives a looser $\tau_{\mathrm{guar}}$) and predicts an intra-binade step at $\eta^* = 2^{-16} \approx 1.53{\times}10^{-5}$.
(PyTorch's separate decoupled weight-decay pass applies relative shrinkage $\eta\lambda \approx 10^{-7} \ll 2^{-9}$, which is a no-op or strictly inward and preserves $[-\tau, +\tau]$ invariance.)
Because individual coordinates can occasionally exceed a global proxy and later-step medians of $\kappa_t$ drop to $0.17$--$0.44$ at $10^{-5}$, $\tau_{\mathrm{emp}}(\eta)$ serves as an early-training guide, whereas our online consistency check verifies $\max_{t,i\in G}|\delta w_{t,i}| < 2^{-16}$ per group directly (\S\ref{sec:thm1_exp}).

\begin{theorem}[Sufficient Group-Wise Snap-Back]
\label{thm:snapback}
Let $G \subset \{1,\dots,d\}$ be a quantization group with initially \bfsixteen{}-representable weights, $M_0 = \max_{i \in G} |w_{0,i}|$, and a deterministic coordinatewise single-quantization scalar encoder $Q_G$ (\texttt{double\_quant=False}) whose scale $s(M)$ depends solely on $M = \max_{i \in G} |w_i|$ and fixes the scalar codebook and thresholds, with zero Voronoi cell containing $(-\hwzero^{(G)}, +\hwzero^{(G)})$ ($\hwzero^{(G)}{=}c_0 M$, $c_0{\le}(1{+}2^{-7})^{-1}$; or frozen-scale $\hwzero(M_0)$ for staircase scales such as MXFP4).
Under pure-\bfsixteen{} RN with $|\delta w_{t,i}| < \tau / 256$ for all $t{\le}T, i{\in}G$ ($\tau{=}2^k, k{\in}[-125,126]$), if $\hwzero^{(G)} > \tau$ and $M_0 \ge \tau_+$, then:
(i)~$M_T{=}M_0$ freezes permanently and all Voronoi cells stay fixed;
(ii)~every active coordinate stays in $[-\tau,+\tau] \subset (-\hwzero^{(G)},+\hwzero^{(G)})$ and decodes to $0$;
(iii)~hence $Q_G(W_T){=}Q_G(W_0)$ numerically ($\Delta Q_G{=}0$, $\rho_G{=}0$ if $\Delta W_G{\neq}0$): bit-for-bit under a canonical zero code, and in decoded value under signed-zero formats ($\pm 0$).
\end{theorem}

\begin{corollary}[FP4 Central-Cluster Containment and Exact $\rho_G^{\mathrm{FP4}}$]
\label{cor:fp4_cluster}
Under Proposition~\ref{prop:tau_eta}(2)'s step bound on single-quantized \texttt{bnb} FP4 ($g{=}64$, \texttt{double\_quant=False}), whose central cluster $\{0,\pm\hat{c}_{1,+}(M)\}$ (nibbles $\{\texttt{0,1,8,9}\}$, ideal $M/192$) spans $\pm c_{\mathrm{clus}}M$ ($c_{\mathrm{clus}}{=}\tfrac{11}{128}$): whenever $c_{\mathrm{clus}}M_0>\tau$, $M_T{=}M_0$ freezes, $|w_{0,i}|\ge\tau_+$ keeps its nibble, and $|w_{0,i}|\le\tau$ stays in $\{\texttt{0,1,8,9}\}$ (containment, not exact snap-back).
Because only two inner thresholds ($\pm M_0/384$ ideally) separate $\{0,\pm\hat{c}_{1,+}(M_0)\}$, any $\Delta W_G{\neq}0$ obeys $\rho_G^{\mathrm{FP4}} = \hat{c}_{1,+}(M_0)\sum_{i\in G}|\Delta w_i|N_i / \|\Delta W_G\|_2^2$, where $N_i{\in}\{0,1,2\}$ counts value-changing threshold crossings between $Q_G(w_{0,i})$ and $Q_G(w_{T,i})$ (decoded $\hat{c}_{1,+}(M_0)$ in finite precision; $M_0/192$ ideally).
Hence $|\Delta Q_{G,i}| \le 2\hat{c}_{1,+}(M_0)$: $M_0/96$ ideally, and for normal $M_0 \ge 192{\cdot}2^{-126}$, $\le (1{+}2^{-23})M_0/96$ (\fpthirtytwo{} RN) and $\le (1{+}2^{-7})M_0/96 \approx 0.01050M_0$ (\bfsixteen{} dequantization; post-hoc refinement covering excesses up to $0.010444 M_0$, \S\ref{sec:thm1_exp}).
\end{corollary}

\paragraph{Wide-gap blind zones and movement sanity check.}
We call a quantizer \emph{wide-gap} relative to training precision when its zero-cell or central-cluster half-width $\hwzero$ approaches or exceeds $\tau(\eta)$.
Near-zero compression spans four regimes: exact zero-bin snap-back (Sym-INT4 $\hwzero{=}M/15$, \texttt{bnb}-NF4 $0.03979M$, OCP MXFP4 \citep{rouhani2023microscaling} $g{=}32$), FP4 central-cluster containment (\texttt{bnb}-FP4 \citep{dettmers2023qlora} $\pm\tfrac{11}{128}M$), layerwise magnitude pruning (quantile $q_s(|W_0|)$), and calibrated PTQ (GPTQ \citep{frantar2023gptq}, AWQ \citep{lin2024awq}, \texttt{LLM.int8()} \citep{dettmers2022llmint8}).
To catch silent freezing during training with negligible compute overhead, we log the moved fraction over trained linear weights, $f_{\mathrm{move}}^{\mathrm{tr}} \triangleq |\mathcal{M}|/d_{\mathrm{tr}}$ (evaluated against $W_0$ or a sampled reference).

\begin{proposition}[Fixed-order scale-shift decomposition and crossing identity]
\label{prop:rho_q}
For a group-wise quantizer $Q_s(\cdot)$ with scales $s{=}s(W_0)$ and $s'{=}s(W_u)$, $\Delta Q = \Delta Q_{\mathrm{fix}} + \Delta Q_{\mathrm{scl}} \triangleq [Q_s(W_u){-}Q_s(W_0)] + [Q_{s'}(W_u){-}Q_s(W_u)]$, giving $\rho {=} \rho_{\mathrm{fix}} + \rho_{\mathrm{scl}}$ ($\Delta W{\neq}0$).
With ordered levels $c_{G,k} < c_{G,k+1}$, thresholds $b_{G,k} < b_{G,k+1}$, and $|\Delta w_i| < \min_k(b_{G,k+1}{-}b_{G,k})$ for $i{\in}G$,
\begin{equation}
  \rho_{\mathrm{fix}} = \frac{1}{\|\Delta W\|_2^2}\sum_G \sum_k (c_{G,k+1}{-}c_{G,k})\!\sum_{i\in G} |\Delta w_i|\,\mathbb{1}_{G,k}(i),
  \label{eq:rho_exact}
\end{equation}
where $\mathbb{1}_{G,k}(i)$ indicates exact encoder crossing of $b_{G,k}$.
Under a locally smooth globally energy-normalized density $\tilde{p}_G$ of $w_{0,i}$ ($\sum_G \int \tilde{p}_G(w)\,dw{=}1$, weighted by $\Delta w_i^2/\|\Delta W\|_2^2$) with step signs locally independent of position near $b_{G,k}$, Eq.~\eqref{eq:rho_exact} heuristically yields $\hat{\rho}_{\mathrm{smooth}} = \sum_G \sum_k (c_{G,k+1}{-}c_{G,k})\,\tilde{p}_G(b_{G,k})$ (Table~\ref{tab:app_prop2_mediation}).
\end{proposition}

\begin{figure*}[t]
  \centering
  \includegraphics[width=0.92\linewidth]{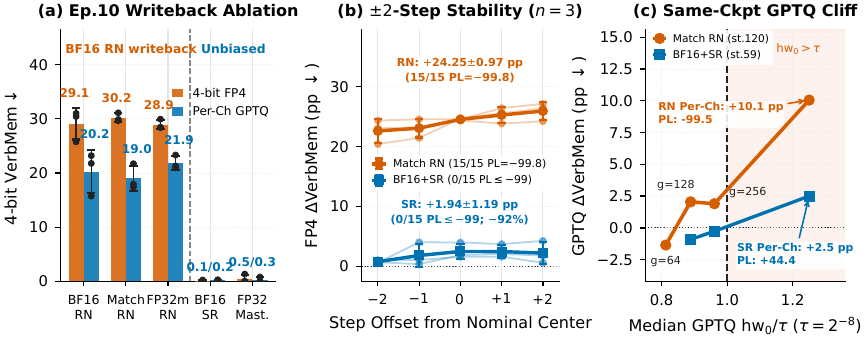}
  \caption{\textbf{Writeback causation, independent re-run $\pm 2$-step stability, and single-seed GPTQ group-size sweep on MUSE-News (Llama-2-7B).}
  (a)~Ep.~10 five-arm ablation ($n{=}3$): \fpthirtytwo{} moments under RN leave rebound intact, whereas toggling writeback to \bfsr{} cuts FP4 $\VerbMem$ by $99.7\%$ and per-channel GPTQ by $98.9\%$ vs.\ matched RN.
  (b)~Independent fresh retraining pass across $\pm 2$-step neighborhoods around nominal early-stop steps ($3$ seeds, $30$ within-seed correlated checkpoints): all $15/15$ RN steps rebound $+20.4$--$+27.1$\,pp and reach Target-oriented privacy scores (FP4 $\PrivLeak{=}{-}99.77$, $\Delta\PrivLeak$ from $-35.9$ to $-153.4$), versus $+0.4$--$+4.2$\,pp ($0/15$ reaching $\PrivLeak{\le}{-}99.0$) for \bfsr{}.
  (c)~Single-seed, same-checkpoint GPTQ sweep (independent re-run $s_0$, WikiText-2 $128{\times}1024$): RN $\VerbMem$ stays near-flat for $g{\in}\{64,128,256\}$ ($\hwzero{<}\tau$) while $\PrivLeak$ shifts progressively ($-50.6$ at $g{=}256$) and jumps $+10.1$\,pp ($\Delta\PrivLeak{=}{-}126.3$) at per-channel ($1.25\tau$), versus $+2.5$\,pp ($\Delta\PrivLeak{=}{-}6.1$) on \bfsr{}.}
  \label{fig:factorial}
\end{figure*}

\begin{table*}[t]
\centering
\caption{\textbf{Causal writeback isolation on MUSE-News (Llama-2-7B \citep{touvron2023llama2}, NPO+GDR, $\eta{=}10^{-5}$, \texttt{FailureLLMUnlearning}, $n{=}3$): Ep.~10 ablation (Panel~A) and two independent early-stop controls (Panel~B).}
$\PrivLeak{=}100(\mathrm{AUC}{-}\mathrm{AUC}_{\mathrm{Retrain}})/\mathrm{AUC}_{\mathrm{Retrain}}$ (via Min-$K\%$ \citep{shi2024detecting}): official unq.\ Retrain reports $-0.02$ (empirical $0.00$ from logged AUC $47.71\%$; GPTQ-Ch Retrain $-3.8$); $0$ matches Retrain, $-100$ reflects Target-oriented leakage, $+109.6$ reflects extreme over-unlearning ($\mathrm{AUC}{\approx}1.0$, still separable in the reversed orientation; avoiding $\le{-}99$ is not calibrated privacy).
$^*$Raw \fpthirtytwo{} basis ($100.0\%$ moved, $17.50\%$ in $|w_0|{\le}2^{-8}$; after \bfsixteen{} export at Ep.~10: $87.5\%$ moved, $19.70\%$ in $\le 2^{-8}$).
$^\dagger$Independent fresh retraining pass (Run~2) at Run-1 nominal centers; effective unit is $n{=}3$ seeds ($15/15$ RN reach FP4 $\PrivLeak{\le}{-}99.0$ [$-99.77$; $13/15$ with $\Delta\PrivLeak{\le}{-}45$] vs.\ $0/15$ SR [worst $-96.27$ from $-95.10$]). Full tables in Appendix Tables~\ref{tab:app_paired}--\ref{tab:app_w16_neighbor_gptq}.}
\label{tab:factorial}
\scriptsize
\renewcommand{\arraystretch}{1.04}
\setlength{\tabcolsep}{3.0pt}
\resizebox{\textwidth}{!}{%
\begin{tabular}{lcccccc}
\toprule
\textbf{Arm / Control ($n{=}3$ seeds)} & \textbf{Chg./$\le\!2^{-8}$\%} & \textbf{Unq.\ VM ($\KnowMemR$; PL)} & \textbf{FP4 VM ($\Delta$)} & \textbf{FP4 $\PrivLeak$} & \textbf{GPTQ-Ch VM (PL)} & \textbf{NF4 VM} \\
\midrule
\texttt{Target} / \texttt{Retrain} (ref.) & --- & 56.97 (54.4; $-$99.8) / 20.00 (55.4; $-$0.02) & 46.82 / 19.99 & $-$99.8 / $-$2.1 & 43.05 ($-$99.8) / 19.39 ($-$3.8) & 50.28 / 20.81 \\
\midrule
\multicolumn{7}{l}{\emph{Panel A: Standard Epoch-10 Five-Arm Writeback Ablation ($n{=}3$)}} \\
\texttt{official\_bf16\_rn} (native) & 17.07 / 98.94 & 0.00 (40.98; +109.5) & \textbf{29.07$\pm$2.88} & \textbf{$-$99.75} & \textbf{20.23$\pm$3.97} ($-$61.9) & 1.61 \\
\texttt{matched\_rn} (wrapper base) & 17.06 / 98.99 & 0.00 (40.85; +109.5) & \textbf{30.21$\pm$0.79} & \textbf{$-$99.76} & \textbf{19.01$\pm$2.36} ($-$70.7) & 1.40 \\
\texttt{fp32\_mom\_bf16\_rn} & 17.13 / 98.58 & 0.00 (41.49; +109.4) & \textbf{28.92$\pm$0.88} & \textbf{$-$99.77} & \textbf{21.86$\pm$1.27} ($-$82.0) & 1.56 \\
\texttt{official\_bf16\_sr} (\textbf{SR}) & 91.96 / 18.75 & 0.00 (36.61; +109.6) & \textbf{0.09$\pm$0.12} & \textbf{+109.56} & \textbf{0.21$\pm$0.07} (+109.6) & 0.00 \\
\texttt{official\_fp32\_master} & 100.0$^*$ / 17.50$^*$ & 0.00 (38.67; +109.6) & \textbf{0.45$\pm$0.77} & \textbf{+109.56} & \textbf{0.26$\pm$0.45} (+109.6) & 0.10 \\
\midrule
\multicolumn{7}{l}{\emph{Panel B: Early-Stop Controls --- Primary Run (Run~1) \& Independent Re-Run$^\dagger$ (Run~2)}} \\
\texttt{matched\_rn} (Run~1, Ep.\ 2.42; RMS $0.874$) & 16.63 / \textbf{99.83} & 9.12 (39.49; +5.2) & \textbf{34.02$\pm$0.27} (\textbf{+24.9}) & \textbf{$-$99.77} & \textbf{20.18$\pm$0.99} (\textbf{$-$99.2}) & 14.43 \\
\texttt{official\_bf16\_sr} (Run~1, Ep.\ 1.10; RMS $1.564$) & 76.84 / \textbf{21.89} & 10.76 (36.04; $-$10.1) & \textbf{12.83$\pm$3.89} (\textbf{+2.1}) & $-$26.09 & \textbf{10.95$\pm$4.17} (+10.1) & 13.18 \\
\texttt{official\_fp32\_master} (Run~1, Ep.\ 1.14; RMS $1.096$) & 70.59 / \textbf{23.80} & 12.70 (35.38; $-$12.1) & \textbf{14.76$\pm$3.67} (\textbf{+2.1}) & $-$32.70 & \textbf{12.64$\pm$3.08} (+4.9) & 15.47 \\
Run~2$^\dagger$ $\pm 2$-step ($15$ vs.\ $15$ ckpts; matched-VM) & RN vs.\ \textbf{SR} & Unq.\ PL: $-64$..$+54$ vs.\ $-95$..$+96$ & \textbf{+24.25} vs.\ \textbf{+1.94} (\textbf{+25.1}/\textbf{+2.2}) & \textbf{15/15} vs.\ \textbf{0/15} & --- & --- \\
\bottomrule
\end{tabular}}
\end{table*}

\section{Ecosystem Precision Audit of Selected Codebases}
\label{sec:audit}

Five of seven audited public repositories---spanning five distinct trainers, as three MUSE repos share \texttt{IterativeUnlearner}---default to pure-\bfsixteen{} RN under single-GPU launch (all seven if DeepSpeed is omitted on \texttt{tofu} and \texttt{npo}; Table~\ref{tab:ecosystem_audit}; echoing \texttt{transformers} \#50581 \citep{transformers50581}).
Severe freezing strikes the two most widely used benchmarks: the pinned MUSE trainer freezes $82.9\%$ of weights, and reproduced single-GPU \texttt{open-unlearning} freezes $88\%$--$92\%$ on Llama and $94.8\%$ on Qwen2.5.
Crucially, \texttt{open-unlearning} \citep{dorna2025openunlearning} switches from pure-\bfsixteen{} RN (1~GPU) to ZeRO-3 \fpthirtytwo{} masters (multi-GPU) via the launcher alone, so multi-GPU tests can hide a precision failure that triggers deterministically on single-GPU setups.

\section{Matched Writeback Intervention on MUSE-News}
\label{sec:causal}

\subsection{Writeback Isolation at Epoch 10}
\label{sec:factorial}

Using the pinned \texttt{FailureLLMUnlearning} codebase (\texttt{10131ae25f}, Llama-2-7B, NPO+GDR, $\eta{=}10^{-5}$, 10~ep., $n{=}3$), we isolate weight writeback rounding from optimizer moment precision (Table~\ref{tab:factorial} Panel~A, Figure~\ref{fig:factorial}a).
Upgrading AdamW moments from \bfsixteen{} to \fpthirtytwo{} under RN leaves rebound unchanged ($28.92$ vs.\ $29.07$ native RN and $30.21$ single-axis \texttt{matched\_rn} on FP4).
Toggling \emph{only} writeback in the matched wrapper to \bfsr{} unlocks $92.0\%$ of weights, cutting FP4 $\VerbMem$ by $99.7\%$ ($30.21{\to}0.09$) and per-channel GPTQ by $98.9\%$ ($19.01{\to}0.21$), matching \fpthirtytwo{} masters.
RN's forget $\KnowMemF$ stays flat under FP4 ($49.4{\to}48.4$), so compression targets verbatim memorization; consistent with $\tau_{\mathrm{emp}}(\eta)$ scaling, single-seed ($s_0$) checks show halving RN LR to $5{\times}10^{-6}$ extends severe rebound to NF4 ($38.4$ $\VerbMem$), whereas raising LR to $2{\times}10^{-5}$ or $10^{-4}$ nearly removes greedy FP4 rebound ($0.90$ and $0.00$; Appendix Tables~\ref{tab:app_pareto},~\ref{tab:app_multig}, and~\ref{tab:app_supp_controls}).

\subsection{Non-Saturated Early-Stop, $\pm 2$-Step Stability, and Matched Pairs}
\label{sec:pareto}

Because Ep.~10 \bfsr{} and \fpthirtytwo{} sit at the over-unlearning privacy ceiling ($\PrivLeak{=}{+}109.6$), we verify that the writeback effect persists outside saturation via three progressively tighter controls (Table~\ref{tab:factorial} Panel~B, Figure~\ref{fig:factorial}b, Appendix Tables~\ref{tab:app_privleak0}--\ref{tab:app_pareto}).

\paragraph{Primary non-saturated early-stop (Run~1, $n{=}3$).}
Blind windowed scans selecting $\arg\min_t|\text{unq.\ }\PrivLeak(t)|$ inside unmodified \texttt{FailureLLMUnlearning} yield non-saturated checkpoints (Ep.~$2.42$ RN vs.\ $1.10$/$1.14$ SR/\fpthirtytwo{}; unq.\ $\KnowMemF$ $41.4$/$36.4$/$26.7$).
Early stopping narrows the update-RMS gap vs.\ Ep.~10 (\fpthirtytwo{} $1.096{\times}10^{-4}$ is $25.4\%$ above RN $0.874{\times}10^{-4}$, vs.\ $1.76\times$ at Ep.~10; SR $1.564{\times}10^{-4}$ is $+79.0\%$, vs.\ $2.5\times$ at Ep.~10), though magnitude differences remain and co-mediate recovery.
RN still rebounds $+24.90{\pm}0.68$\,pp (FP4), $+11.06{\pm}0.53$\,pp (GPTQ-Ch), and $+5.31{\pm}0.47$\,pp (NF4, $\PrivLeak{=}{-}72.4$, showing NF4 is not uniformly benign), hitting Target-oriented privacy on $3/3$ seeds under FP4/GPTQ ($\PrivLeak{=}{-}99.8$/$-99.2$).
\bfsr{} and \fpthirtytwo{} cut FP4 rebound by $91.7\%$ ($+2.07$\,pp; $T{=}0.9$ sampling $+15.41{\to}+1.25$/$+1.52$\,pp over $54$ raw draws) and hold GPTQ-Ch drift near zero ($+0.20$/$-0.06$\,pp; seed range $-1.24$ to $+1.14$), avoiding Target-oriented collapse (though $\PrivLeak$ measures signed attack orientation rather than calibrated privacy: strongly reversed scores remain separable) with seed-level privacy spread (FP4 $-26.1$/$-32.7$; GPTQ-Ch SR $-28.8$ to $+32.5$) and slightly lower retain ($36.0$/$35.4$ vs.\ $39.5$).
(For $n{=}3$, exact sign-flip $p{=}0.25$; parametric paired $t$-test $p{\le}0.0053$ for FP4/GPTQ-Ch, $0.0103$/$0.0352$ for NF4.)

\paragraph{Independent re-run $\pm 2$-step stability, non-determinism, and matched pairs (Run~2, $n{=}3$).}
An \emph{independent fresh retraining pass} evaluates FP4 across offsets $\{-2..+2\}$ around Run~1's nominal centers ($30$ within-seed correlated checkpoints, effective $n{=}3$; Figure~\ref{fig:factorial}b, Appendix Table~\ref{tab:app_w16_neighbor_gptq}).
Non-deterministic \bfsixteen{} training shifts unquantized operating points across runs at fixed steps (e.g., RN $s_1$ st.~$126$: $+6.94{\to}{-}32.17$; SR $s_0$ st.~$59$: $-36.48{\to}{+}50.48$), so $3/6$ centers cease to be local $\min|\PrivLeak|$.
Yet RN's rebound reproduces across both runs: $15/15$ RN steps rebound $+20.38$--$+27.11$\,pp (seed-mean $+24.25{\pm}0.97$\,pp) and reach Target-oriented FP4 $\PrivLeak{=}{-}99.77$ ($\Delta\PrivLeak$ $-35.9$ to $-153.4$; $13/15$ drop $\ge 45$\,pts), while all $15/15$ \bfsr{} steps stay $+0.36$--$+4.24$\,pp (seed-mean $+1.94{\pm}1.19$\,pp, $-92.0\%$; $0/15$ reach $\PrivLeak{\le}{-}99.0$, worst $-96.27$ [AUC $1.78\%$, $98.2\%$ reversed] from $-95.10$ unq.).
Within-seed closest unq.-$\VerbMem$ pairs ($|\Delta|{\le}0.55$\,pp, $n{=}3$) give RN $+25.08{\pm}1.13$ vs.\ \bfsr{} $+2.15{\pm}1.71$\,pp (seed~0: RN st.~$119$ vs.\ SR st.~$59$ match $\VerbMem$/$\KnowMemF$/$\KnowMemR$, though unq.\ $\PrivLeak$ is $-1.03$ vs.\ $+50.48$, rebounding $+24.51$ vs.\ $+1.87$\,pp; Ep.~1--10 trajectories in Appendix Table~\ref{tab:app_pareto}).

\subsection{Same-Checkpoint GPTQ Group-Size Sweep}
\label{sec:gptq_cliff}

In \citet{zhang2025catastrophic}, 4-bit GPTQ and AWQ group sizes are undisclosed; earlier cross-harness checks found minimal RN rebound ($\le 0.21$ $\VerbMem$; Table~\ref{tab:app_supp_controls}) under group-$128$ GPTQ/AWQ despite strong rebound under per-channel GPTQ.
Holding both the Run-2 checkpoint (within each arm) and exact WikiText-2 ($128{\times}1024$) \texttt{llmcompressor} W4A16 calibration fixed (zero max logit parity difference between the quantized-export and evaluation loaders; Table~\ref{tab:gptq_cliff}, Figure~\ref{fig:factorial}c), sweeping GPTQ group size provides threshold-compatible evidence (single seed $s_0$, $4$ points where $\hwzero$ co-varies with group size; Theorem~\ref{thm:snapback} does not cover Hessian-coupled GPTQ):
on \texttt{matched\_rn} (Run~2 $s_0$ st.~$120$, unq.\ $9.14$, $\PrivLeak{=}{+}26.8$), $g{\in}\{64,128,256\}$ keep $\hwzero{<}\tau$ ($0.81\tau$--$0.96\tau$) and near-flat $\VerbMem$ ($-1.35$ to $+2.05$\,pp), while $\PrivLeak$ shifts progressively ($-50.6$ at $g{=}256$, $\Delta\PrivLeak{=}{-}77.4$) before jumping $+10.05$\,pp ($19.19$) and collapsing privacy ($\PrivLeak{=}{-}99.46$, $\Delta\PrivLeak{=}{-}126.3$) once per-channel crosses $\tau$ ($\hwzero{=}0.004883{=}1.25\tau$).
On \texttt{official\_bf16\_sr} (Run~2 $s_0$ st.~$59$, unq.\ $9.56$, $\PrivLeak{=}{+}50.5$) at identical per-channel $\hwzero$, GPTQ drifts $+2.50$\,pp with paired $\Delta\PrivLeak{=}{-}6.1$ ($+44.40$).
Uncalibrated coarse Sym-INT4 RTN still leaks partially under \bfsr{} ($29.5\%$--$50.3\%$ cut at per-channel/$g{=}512$; Table~\ref{tab:app_supp_controls}), confirming that severe uncalibrated noise can leak independently of RN freezing.

\begin{table}[t]
\centering
\caption{\textbf{Same-checkpoint, same-calibration GPTQ group-size sweep (Run~2 $s_0$, WikiText-2 $128{\times}1024$, exact loader logit parity).} Single-seed threshold-compatible evidence ($g{=}512$ skipped: $512\nmid 11008$; SR $g{=}64$ omitted).}
\label{tab:gptq_cliff}
\scriptsize
\renewcommand{\arraystretch}{1.02}
\setlength{\tabcolsep}{2.6pt}
\resizebox{\columnwidth}{!}{%
\begin{tabular}{lcccc}
\toprule
\textbf{Run-2 Arm (Unq.\ VM; PL)} & \textbf{Group} & \textbf{$\hwzero/\tau$} & \textbf{GPTQ VM ($\Delta$)} & \textbf{GPTQ PL ($\Delta$)} \\
\midrule
\multirow{4}{*}{\shortstack[l]{\texttt{matched\_rn}\\($s_0$ st.\ 120:\\9.14; +26.8)}}
 & \textbf{Per-Ch} & \textbf{1.25} & \textbf{19.19 (+10.1)} & \textbf{$-$99.5 ($-$126.3)} \\
 & $g{=}256$ & 0.96 & 11.04 (+1.9) & $-$50.6 ($-$77.4) \\
 & $g{=}128$ & 0.89 & 11.19 (+2.1) & $-$3.1 ($-$29.9) \\
 & $g{=}64$ & 0.81 & 7.79 ($-$1.4) & +30.0 (+3.2) \\
\midrule
\multirow{3}{*}{\shortstack[l]{\texttt{official\_bf16\_sr}\\($s_0$ st.\ 59:\\9.56; +50.5)}}
 & \textbf{Per-Ch} & \textbf{1.25} & \textbf{12.06 (+2.5)} & \textbf{+44.4 ($-$6.1)} \\
 & $g{=}256$ & 0.96 & 9.29 ($-$0.3) & +56.8 (+6.3) \\
 & $g{=}128$ & 0.89 & 8.66 ($-$0.9) & +58.5 (+8.0) \\
\bottomrule
\end{tabular}}
\end{table}

\section{Mechanistic Checks and Pure-FP32 Surgeries}
\label{sec:surgeries}

\subsection{Threshold Cliffs and 130-Step Consistency}
\label{sec:thm1_exp}

Fixed-weight Sym-INT4 clip sweeps tie rebound qualitatively to $\hwzero/\tau(\eta)$: on Llama-2-7B ($\eta{=}10^{-5}$, $g{=}128$), crossing $\hwzero{=}2^{-8}$ jumps $\VerbMem$ from $5.68$ to $30.83$ at intact retain (Appendix Table~\ref{tab:app_multig}), and on TOFU-1B (Figure~\ref{fig:hero}b), RN curves peak near $\hwzero/\tau{\approx}1$ while matched \fpthirtytwo{} controls stay flat ($0.0\%$--$4.8\%$).
Online tracking of $|\delta w_{t,i}| < 2^{-16}$ across all 130 steps of TOFU-1B RN consistency-checks Theorem~\ref{thm:snapback} and Corollary~\ref{cor:fp4_cluster} with zero certified failures across examined groups under exact NPO packed nibbles and the machine bound $(1{+}2^{-7})M_0/96$ (Figure~\ref{fig:theorem1}b--c, Appendix Tables~\ref{tab:app_contingency} and~\ref{tab:app_thm1}; note that natural certificate coverage is limited---$17.8\%$--$25.0\%$ under the stricter $\hwzero{\ge}\tau_+$ filter [$18.1\%$--$25.5\%$ online $\hwzero{>}\tau$] for Sym-INT4 $g{=}64/128$, $2.5\%$ for NF4, and $0\%$ exact-zero FP4---and the ideal real bound $M_0/96$ saw $5{,}646$/$1{,}330$ tiny \bfsixteen{} dequantization excesses up to $0.010444 M_0$).
Importantly, support geometry alone does not explain every natural RN run: in the 10-epoch constant-LR TOFU-1B sweep (Figure~\ref{fig:theorem1}a, Table~\ref{tab:app_supp_controls}, distinct from the 5-epoch sweeps in Tables~\ref{tab:app_warmup} and~\ref{tab:app_universal}), raising LR from $10^{-5}$ to $1.4{\times}10^{-5}$ cuts FP4 rebound ($26.3{\to}7.4$\,pp) while $\le 2^{-8}$ support remains $99.66\%$, and at matched update energy $E_{\Delta W}$ GradDiff FP4 backfill shows no certified-vs-uncertified separation ($22.79$ vs.\ $22.54$; Table~\ref{tab:app_cf}), confirming that update magnitude co-mediates recovery alongside support geometry.

\begin{figure*}[t]
  \centering
  \includegraphics[width=0.90\linewidth]{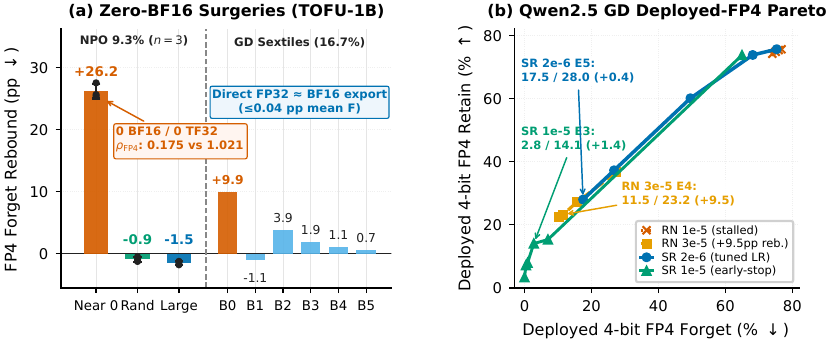}
  \caption{\textbf{Zero-\bfsixteen{} causal support surgeries and deployed-FP4 Qwen2.5 trade-off.}
  (a)~Strict IEEE-\fpthirtytwo{} NPO ($\eta{=}4{\times}10^{-5}$, $9.34\%$ per-layer matched counts, $n{=}3$): near-zero masks rebound $+26.2$\,pp ($\rho_{\mathrm{FP4}}{=}0.175$) vs.\ $-0.9$\,pp (random) and $-1.5$\,pp (large). Among six $16.7\%$ GradDiff sextiles, only $B_0$ exceeds $5$\,pp ($+9.9$\,pp).
  (b)~Qwen2.5-1.5B GradDiff on the \emph{deployed} FP4 (forget, retain) plane ($s_0$): RN $3{\times}10^{-5}$ still incurs $+9.5$\,pp FP4 rebound, while tuned \bfsr{} and high-LR RN remain mutually non-dominated ($18/23$ non-dominated grid points).}
  \label{fig:surgeries}
\end{figure*}

\begin{table*}[t]
\centering
\caption{\textbf{Cross-architecture freezing and rebound ($1\mathrm{B}$--$8\mathrm{B}$), localized $27\mathrm{B}$ freezing check, and Qwen2.5 deployed-FP4 trade-off (\texttt{OpenUnlearning}, default $\eta{=}10^{-5}$ unless noted).}
All rows use \fpthirtytwo{} moments; \bfsr{} toggles only writeback rounding. Post-PTQ cells are bolded only for clear rebound; pre-PTQ RN freezing is tagged \emph{(stall)}, and $0.00\%$ rows are tagged \emph{[floor]} (uninformative for rebound suppression). \fpthirtytwo{} masters match \bfsr{} within $0.14$\,pp forget / $1.9$\,pp retain on classic 8B/Qwen2.5 unq./FP4/NF4 blocks (Qwen2.5 GD GPTQ-128 differs by $2.56$\,pp; on Gemma-4, unq.\ \bfsr{} retain exceeds \fpthirtytwo{} by $6.8$--$11.0$\,pp; Tables~\ref{tab:app_tofu_full},~\ref{tab:app_qwen_pareto},~\ref{tab:app_w17_2026}).}
\label{tab:tofu_ecosystem}
\footnotesize
\renewcommand{\arraystretch}{1.03}
\setlength{\tabcolsep}{2.8pt}
\resizebox{\textwidth}{!}{%
\begin{tabular}{llcccc}
\toprule
\textbf{Model / Method} & \textbf{Arm (FP32 moments)} & \textbf{Chg.\ (\%)} & \textbf{Unquant.\ F / R} & \textbf{FP4 F / R} & \textbf{NF4 F / R} \\
\midrule
\multirow{2}{*}{Llama-3.1-8B NPO ($n{=}2$)}
 & Default \texttt{paged32} RN & 11.14 & 14.09 / 83.78 & \textbf{35.81} / 50.40 & 19.24 / 74.03 \\
 & Single-axis \textbf{BF16+SR} & 78.59 & 2.92 / \textbf{93.77} & 3.63 / \textbf{91.03} & 3.31 / \textbf{92.25} \\
\addlinespace[2pt]
\multirow{2}{*}{Llama-3.1-8B GD ($n{=}2$)}
 & Default \texttt{paged32} RN & 10.98 & 2.21 / 54.48 & \textbf{62.72} / 84.33 & 7.17 / 44.06 \\
 & Single-axis \textbf{BF16+SR} \emph{[floor]} & 73.93 & 0.00 / \textbf{80.53} & 0.00 / 74.60 & 0.00 / \textbf{77.86} \\
\addlinespace[2pt]
\multirow{2}{*}{Qwen2.5-1.5B NPO ($n{=}3$)}
 & Triton \bfsixteen{}-RN \emph{(stall+reb.)} & 5.2$\pm$0.1 & 49.71 / 51.70 & \textbf{72.47} / 72.82 & \textbf{70.61} / 69.90 \\
 & Triton \textbf{BF16+SR} & 67.0$\pm$0.2 & 5.74 / \textbf{89.47} & 7.94 / \textbf{74.23} & 6.97 / \textbf{82.54} \\
\addlinespace[2pt]
\multirow{2}{*}{Qwen2.5-1.5B GD ($n{=}3$)}
 & Triton \bfsixteen{}-RN \emph{(stall)} & 5.30$\pm$0.02 & 68.48 / 86.42 & 73.55 / 73.96 & 84.74 / 84.14 \\
 & Triton \textbf{BF16+SR} \emph{[floor]} & 65.16$\pm$0.06 & 0.00 / 26.09 & 0.03 / 6.23 & 0.02 / 13.30 \\
\midrule
\multicolumn{6}{l}{\emph{2026-Era Architectures (TOFU-forget10, 5~ep., $\eta{=}10^{-5}$; P20 / $\rho$ / \fpthirtytwo{} in Table~\ref{tab:app_w17_2026})}} \\
\addlinespace[2pt]
\multirow{2}{*}{Gemma-4-E4B NPO ($n{=}3$)}
 & \bfsixteen{}-RN \emph{(stall+reb.)} & 6.33$\pm$0.03 & 47.45 / 49.28 & \textbf{54.92} / 54.33 & \textbf{52.49} / 52.68 \\
 & \textbf{BF16+SR} & 59.18$\pm$0.12 & 6.18 / \textbf{52.43} & 5.30 / 48.73 & 5.98 / 51.39 \\
\addlinespace[2pt]
\multirow{2}{*}{Qwen3.8-27B GD (L52--55, $n{=}3/2$)}
 & \bfsixteen{}-RN \emph{(stall)} & 9.89$\pm$0.01 & 58.36 / 60.13 & 58.54 / 57.70 & 57.34 / 58.57 \\
 & \textbf{BF16+SR} & 72.12$\pm$0.07 & 8.38 / 40.74 & 9.04 / 37.03 & 8.99 / 38.98 \\
\midrule
\multicolumn{6}{l}{\emph{Deployed-FP4 Tuned-LR / Early-Stop Trade-off (Qwen2.5-1.5B GradDiff; three non-dominated points, $s_0$ / $n{=}2$)}} \\
\addlinespace[2pt]
\multirow{3}{*}{\shortstack[l]{Qwen2.5-1.5B GD\\Tuned / Early-Stop}}
 & \bfsixteen{}-RN ($3{\times}10^{-5}$, E4, $s_0$) & 15.93 & 2.06 / 33.40 & \textbf{11.52} / 23.23 (\textbf{+9.5}) & 3.34 / 22.37 \\
 & \textbf{BF16+SR} ($10^{-5}$, E3, $s_0$) & 51.02 & 1.34 / 31.63 & 2.77 / 14.05 (+1.4) & 2.16 / 21.47 \\
 & \textbf{BF16+SR} ($2{\times}10^{-6}$, E5, $[s_0,s_2]$) & 28.47 & 16.52 / \textbf{58.95} & 17.25 / \textbf{27.84} (+0.7) & 20.23 / \textbf{42.72} \\
\bottomrule
\end{tabular}}
\end{table*}

\subsection{Pure-FP32 Support-Sufficiency Surgeries}
\label{sec:weight_surgeries}

To test whether near-zero support \emph{suffices} without any \bfsixteen{} arithmetic, we run strict IEEE-\fpthirtytwo{} surgeries (TF32 disabled, $n{=}3$) on TOFU-1B with identical per-layer eligible counts across near-zero, random, and large masks ($9.34\%$ globally across 112 layers; Figure~\ref{fig:surgeries}a, Appendix Tables~\ref{tab:app_eqcard_sextile},~\ref{tab:app_universal}).
Because $20\%$ pruning zeroes the smallest weights and FP4's central cluster spans $\pm\tfrac{11}{128}M$, confining small updates to $|w_0|\le 2^{-9}$ is structurally coupled to snap-back; this test therefore isolates \emph{sufficiency} rather than claiming sole mediation.
Direct \fpthirtytwo{}-to-FP4 rebounds $+26.2$\,pp on near-zero masks ($\rho{=}0.175$, $65$-st.\ $\mathrm{eff\_bs}{=}32$ suite, Table~\ref{tab:app_eqcard_sextile}) versus $-0.9$\,pp (random) and $-1.5$\,pp (large), matching \bfsixteen{} export within $0.04$\,pp arm-mean for NPO ($0.13$\,pp max per-seed; strict GradDiff large-mask differs by $0.77$\,pp, $5.26$ vs.\ $6.03$).
In the separate $125$-st.\ ($\mathrm{eff\_bs}{=}16$) five-operator suite (Table~\ref{tab:app_universal}), near-zero masks rebound $+33.2$\,pp (FP4) and $+39.4$\,pp ($20\%$ pruning), while random and large arm means stay $\le{+}0.73$\,pp under FP4/P20 ($\le{+}1.59$\,pp across all five operators); among six exact-\fpthirtytwo{} GradDiff sextiles only $B_0$ exceeds $5$\,pp ($+9.9$\,pp).
On 7B MUSE-News ($s_0$; Appendix Table~\ref{tab:app_surgeries}), masked \fpthirtytwo{} training ($|w_0|{\le}2^{-8}$) reproduces per-channel GPTQ rebound ($18.85$ $\VerbMem$, though $\PrivLeak{=}{-}28.37$) and reproduces Ep.~10 FP4 partially ($8.02$), confirming partial mediation alongside update magnitude ($\lambda{=}0.75$).

\section{Cross-Scale Generality and Practical Trade-offs}
\label{sec:ecosystem}

\subsection{Cross-Architecture Validation and Fixed-LR Utility Trade-offs}
\label{sec:openunlearning}

Pure-\bfsixteen{} RN writeback freezing generalizes across Llama-3.1-8B ($n{=}2$), Llama-3.2-1B \citep{grattafiori2024llama3} ($n{=}3$: NPO, GradDiff, SimNPO \citep{fan2024simnpo}, RMU \citep{li2024wmdp}), and Qwen2.5-1.5B \citep{qwen2024qwen25} ($n{=}3$) under \texttt{OpenUnlearning} defaults (Table~\ref{tab:tofu_ecosystem}, Figure~\ref{fig:hero}c, Appendix Tables~\ref{tab:app_tofu_full} and~\ref{tab:app_matched_traj}).
RN freezes $88\%$--$95\%$ of trained weights, causing severe FP4 rebound on Llama-8B NPO/GradDiff ($+21.7$/$+60.5$\,pp) and 1B RMU alongside pre-PTQ stall on 1B SimNPO and Qwen2.5.
Toggling only writeback to \bfsr{} unlocks $65\%$--$79\%$ of weights, brings Qwen2.5 NPO forget below $8\%$ across FP4/NF4/GPTQ-128/AWQ-128, and on Llama-8B NPO lifts FP4 retain by $+40.6$\,pp while saving $26.5$\,GiB vs.\ \fpthirtytwo{} masters ($+6.4$\,GiB over paged RN; Table~\ref{tab:app_provenance}).

Writeback freezing recurs across two 2026 architectures ($90\%$--$94\%$ unchanged, $100\%$ moved inside $|w_0|{\le}2^{-8}$), while post-PTQ rebound is objective-dependent (Tables~\ref{tab:tofu_ecosystem} and~\ref{tab:app_w17_2026}).
On \texttt{google/gemma-4-E4B-it} (42 layers, $3.95\mathrm{B}$ 2D weights, $n{=}3$), NPO RN rebounds $+7.47$\,pp (FP4), $+5.04$\,pp (NF4), and $+12.69$\,pp (P20 to $60.13\%$ vs.\ Base $62.06\%$, $\rho_{\mathrm{P20}}{=}0.016$), whereas \bfsr{} unlocks $59.2\%$ ($\rho{\ge}0.997$), lowers forget to $6.18\%$ with higher unq.\ retain ($52.43\%$ vs.\ RN $49.28\%$, \fpthirtytwo{} $45.67\%$), and eliminates rebound ($-0.88$/$-0.20$/$+0.30$\,pp); GradDiff shows RN stall ($58.18\%$, P20 $+3.48$\,pp) and \bfsr{}/\fpthirtytwo{} floor retain drop ($27.0\%$/$16.0\%$).
On hybrid Gated-DeltaNet \texttt{Qwen/Qwen3.8-27B} (localized L52--55, 4/64 layers, $1.52\mathrm{B}$ 2D weights, full 64-layer eval; $n{=}3$ RN / $n{=}2$ SR, no \fpthirtytwo{}), GradDiff RN exhibits pre-PTQ stall ($90.1\%$ unchanged, $\rho_{\mathrm{FP4}}{=}0.196$, $\rho_{\mathrm{P20}}{=}0.006$, unq.\ $58.36\%$, FP4 $+0.18$, P20 $+1.48$\,pp to $59.84\%$ vs.\ Base $59.90\%$), while \bfsr{} unlocks $72.1\%$ ($\rho{\ge}0.995$), reaches $8.38\%$ forget at $40.74\%$ unq.\ retain (down $19.0$\,pp from Base $59.70\%$; FP4 $37.03\%$), with operator-mean drift $-0.02$ to $+0.67$\,pp (max seed $+0.72$\,pp).

Critically, removing RN's implicit shrinkage creates both floor effects and fixed-LR utility trade-offs under \emph{any} unbiased writeback: default-LR GradDiff \bfsr{}/\fpthirtytwo{} drives Llama-8B, Qwen2.5, Gemma-4, and TOFU-1B to the $0.00\%$ forget floor (Llama-8B forget NLL $143.7$: $139.0$/$148.4$; Qwen2.5 FP4 retain $6.2\%$/$5.4\%$), so only non-saturated early-stop MUSE and NPO rows provide informative rebound-suppression evidence.
Our Qwen2.5 GradDiff sweep (Table~\ref{tab:tofu_ecosystem}, Figure~\ref{fig:surgeries}b, Appendix Table~\ref{tab:app_qwen_pareto}) shows that raising RN LR to $3{\times}10^{-5}$ still freezes $84\%$ of weights and incurs $+9.0$--$+9.5$\,pp FP4 rebound, whereas tuned \bfsr{} keeps positive drift $\le +1.43$\,pp; on the deployed FP4 (forget, retain) plane, tuned \bfsr{} and higher-LR RN remain mutually non-dominated.
Unbiased writeback is therefore \emph{not} a drop-in fixed-LR remedy, but an enabler that requires joint LR or early-stopping tuning.

\paragraph{Additional controls in Appendices~\ref{app:details}--\ref{app:tables}.}
Four secondary studies support and bound our findings:
(1)~on Zephyr-7B WMDP-Cyber ($n{=}3$; Table~\ref{tab:app_wmdp}), RN exhibits no severe PTQ rebound ($34.0{\to}33.4$ at $10^{-5}$), and \bfsr{} instead relieves pre-PTQ stall at low LRs ($-3.1$ to $-3.7$\,pp FP4 Cyber) while tuned RN at $1.4$--$1.65{\times}10^{-5}$ sits close to $10^{-5}$ \bfsr{};
(2)~on MUSE-Books ($n{=}3$; Table~\ref{tab:app_supp_controls}), \bfsr{} eliminates FP4 rebound ($28.2{\to}0.0$) but exhibits high unquantized retain variance ($32.2{\pm}21.7$ vs.\ \fpthirtytwo{} $51.5{\pm}10.6$; deployed FP4 retain $41.5{\pm}11.6$ vs.\ $45.0{\pm}5.1$);
(3)~on TOFU-1B, deterministic \texttt{Kahan-BF16} tracks \fpthirtytwo{} closely ($\cos{=}0.902$), Salient-Weight Unlearning (SURE \citep{zhang2025catastrophic}) transitions when LR escapes the dead zone ($92.0\%$ in-mask freeze at $10^{-5}$), and our fused Triton \bfsr{} kernel adds $+0.00$\,GiB weight copy at Triton RN speed ($1.57\times$ faster than naive PyTorch \texttt{autocast} AMP; $1.31\times$ optimizer-step speedup vs.\ fused \fpthirtytwo{}-master; Figure~\ref{fig:triton_latency}, Tables~\ref{tab:app_microbench}--\ref{tab:app_w13_align},~\ref{tab:app_sure}); and
(4)~as a descriptive training sanity check across our correlated checkpoint manifest (Table~\ref{tab:app_lomo_diag}), $f_{\mathrm{move}}^{\mathrm{tr}}{\le}40\%$ correctly classifies $62/67$ TOFU and $23/23$ cross-domain runs, while a Qwen2.5 BeaverTails \citep{ji2023beavertails} refusal-template SFT probe ($n{=}3$; Table~\ref{tab:safety_universal}) confirms the same RN freezing loss under FP4 and $20\%$ pruning.

\section{Related Work}
\label{sec:related}

Following \citet{zhang2025catastrophic}'s observation that 4-bit PTQ restores forgotten text, defenses modify objectives, parameterizations, or quantization pipelines---including quantization-aware unlearning \citep{mishra2026quail}, LoRA adapters \citep{abitante2026lora}, representation edits \citep{fang2026kuda}, circuit sparsity \citep{sadhu2026mansu}, sharpness penalties \citep{khan2026durableun,ranjan2026fomul,wang2026landscape}, optimizer simplifications \citep{lang2026downgrade}, Fisher initialization \citep{openunlearning2026pr212}, and empirical audits \citep{sasi2026bits,balaji2026quantunlearn}.
Standard mixed precision avoids writeback absorption via \fpthirtytwo{} masters \citep{micikevicius2018mixed}, whereas stochastic rounding \citep{gupta2015limited} and Kahan summation \citep{zamirai2020revisiting} cut master-weight memory \citep{yu2024collage,ozkara2025stochastic}; concurrent studies note pure-\bfsixteen{} stalls in RL \citep{miahi2026pulse,gao2026faircast,nazemi2026rl}, ternary export \citep{sahai2026lost}, VLA tuning \citep{singh2026vla}, and SFT \citep{weiss2026bf16,transformers50581,unsloth18192}.
Distinct from these, we connect \bfsixteen{} RN writeback freezing directly to wide-gap four-bit PTQ and pruning rebound in unlearning, characterize the same-checkpoint GPTQ group-size transition, state sufficient snap-back bounds, and separate writeback precision from LR tuning.

\section{Conclusion, Practical Protocol, and Limitations}
\label{sec:limitations}

\paragraph{Conclusion and practical protocol.}
In \citet{zhang2025catastrophic}'s tested MUSE trainer and single-GPU \texttt{open-unlearning}, pure-\bfsixteen{} RN writeback freezing drives wide-gap four-bit verbatim rebound (\texttt{bnb}-FP4, per-channel GPTQ) and magnitude-pruning fragility, whereas group-$128$ GPTQ and AWQ stay below $\hwzero{=}\tau$ on verbatim recovery.
Before blaming an unlearning objective, we recommend:
(1)~log trained-layer movement $f_{\mathrm{move}}^{\mathrm{tr}}$ to detect silent freezing; and
(2)~if frozen, replace pure-\bfsixteen{} RN (\fpthirtytwo{} masters, deterministic compensated Kahan-\bfsixteen{}, or \bfsr{}) \emph{jointly} with LR/stopping re-tuning, evaluating forget, privacy, and retain after compression (since \bfsr{} is not free: Ep.~10 FP4 $\KnowMemR$ is $33.6$ vs.\ RN $45.6$---which reflects RN reverting toward Target---and fixed-LR GradDiff can collapse retain).

\paragraph{Limitations.}
(1)~\emph{Operating points \& variance:} Ep.~10 \bfsr{}/\fpthirtytwo{} sit at $\PrivLeak{=}+109.6$ (reversed-orientation separability, not calibrated privacy); in early-stop runs, $15/15$ RN steps hit FP4 $\PrivLeak{\le}{-}99.0$ vs.\ $0/15$ \bfsr{} (worst $-96.27$), yet \bfsixteen{} non-determinism shifts $\PrivLeak$, \bfsr{} retains seed privacy spread, and GradDiff needs LR tuning (mutually non-dominated vs.\ high-LR RN on Qwen2.5).
(2)~\emph{Theory \& quantizer scope:} Theorem~\ref{thm:snapback} has low natural coverage ($17.8\%$--$25\%$ Sym-INT4, $0\%$ exact FP4; excludes GPTQ); the single-seed GPTQ sweep is threshold-compatible rather than a verified law; near-zero support suffices on 1B but co-mediates 7B recovery with magnitude; coarse RTN retains partial rebound under \bfsr{}.
(3)~\emph{Statistical \& model scope:} the independent unit is a seed ($n{=}3$ MUSE/TOFU/Gemma-4, sign-flip $p{=}0.25$; 8B $n{=}2$; 27B 4-layer GradDiff $n{=}3$/$2$ w/o \fpthirtytwo{}, stall only; Kahan 1B; GPTQ $s_0$); manifest diagnostics pool correlated runs; NVFP4, 70B+, MoE, and sub-4-bit PTQ remain open.

\clearpage
\section*{Impact Statement}
This work identifies pure-\bfsixteen{} RN writeback freezing as a recurring precision defect in widely used LLM unlearning trainers and explains when wide-gap four-bit quantization erases near-zero updates.
By recommending a lightweight movement check ($f_{\mathrm{move}}^{\mathrm{tr}}$) paired with joint writeback-precision and learning-rate tuning, our findings improve the reliability of privacy and copyright unlearning audits without introducing new offensive capabilities.
All experimental scripts, the fused Triton \bfsr{} kernel, and full raw evaluation logs will be released publicly upon publication to support independent verification.

{\setlength{\bibsep}{3.5pt}
\bibliography{references}
\bibliographystyle{icml2026}}

\newpage
\appendix
\onecolumn

\section{Complete Mathematical Proofs}
\label{app:proofs}

For completeness, we state the distributed-support dither baseline summarized in \S\ref{sec:dither} before its proof: for an unbounded uniform scalar quantizer $Q$ of step $\Delta_{\mathrm{q}}$ (thresholds $(m{+}\tfrac{1}{2})\Delta_{\mathrm{q}}$), let $u = ((w_0 - \Delta_{\mathrm{q}}/2) \bmod \Delta_{\mathrm{q}}) / \Delta_{\mathrm{q}} \in [0,1)$ denote coordinate $w_0$'s normalized cell phase.
We treat uniform, update-independent phases across coordinates as a \emph{heuristic modeling assumption} illustrating why dense updates survive coarse quantization, rather than a property derived from model width or adaptive group quantization.

\begin{lemma}[Dithered expectation and variance \citep{schuchman1964dither,gray1993dithered}]
\label{prop:dither}
For fixed $\Delta_{\mathrm{q}}$ and $|\Delta w| < \Delta_{\mathrm{q}}$, if $u \sim \mathrm{Unif}[0,1)$ is independent of $\Delta w$, then $\delta Q \triangleq Q(w_0+\Delta w)-Q(w_0)$ satisfies $\mathbb{E}_{u}[\delta Q] = \Delta w$ and $\mathrm{Var}_{u}[\delta Q] = \Delta_{\mathrm{q}}|\Delta w| - \Delta w^2$.
\end{lemma}

\begin{corollary}[Directional SNR in wide layers]
\label{cor:snr}
Under Lemma~\ref{prop:dither} with conditionally independent phases across $i{=}1,\dots,d$, for any fixed $\Delta W \in (-\Delta_{\mathrm{q}}, \Delta_{\mathrm{q}})^d$ and fixed probe $g \in \mathbb{R}^d$ with $\mathrm{Var}(g^\top \Delta Q) > 0$, $\mathbb{E}[g^\top \Delta Q] = g^\top \Delta W$ and
\begin{equation}
  \mathrm{SNR}_g \triangleq \frac{|\mathbb{E}[g^\top \Delta Q]|}{\sqrt{\mathrm{Var}(g^\top \Delta Q)}}
  \ge \frac{|g^\top \Delta W|}{\sqrt{\Delta_{\mathrm{q}} \sum_{i=1}^d g_i^2 |\Delta w_i|}}.
\end{equation}
\end{corollary}

\begin{proof}[Proof of Lemma~\ref{prop:dither} and Corollary~\ref{cor:snr}]
Write $w_0 = (m + \tfrac{1}{2})\Delta_{\mathrm{q}} + u\Delta_{\mathrm{q}}$ with $m \in \mathbb{Z}$ and $u \sim \mathrm{Unif}[0,1)$ independent of fixed $\Delta w$, so the quantization thresholds lie at $(m+\tfrac{1}{2})\Delta_{\mathrm{q}}$ and $Q(w_0) = (m+1)\Delta_{\mathrm{q}}$.
Without loss of generality assume $0 \le \Delta w < \Delta_{\mathrm{q}}$ (the negative case is symmetric).
Then $Q(w_0 + \Delta w) - Q(w_0) = \Delta_{\mathrm{q}}$ if $u + \Delta w / \Delta_{\mathrm{q}} \ge 1$ (which occurs with probability $p = \Delta w / \Delta_{\mathrm{q}}$) and $0$ otherwise.
Hence $\mathbb{E}[\delta Q \mid \Delta w] = \Delta_{\mathrm{q}} p = \Delta w$ and $\mathrm{Var}(\delta Q \mid \Delta w) = \Delta_{\mathrm{q}}^2 p(1-p) = \Delta_{\mathrm{q}}|\Delta w| - \Delta w^2$.
For Corollary~\ref{cor:snr}, linearity of expectation yields $\mathbb{E}[g^\top \Delta Q] = \sum_{i=1}^d g_i \Delta w_i = g^\top \Delta W$.
By conditional independence of phases across coordinates, $\mathrm{Var}(g^\top \Delta Q) = \sum_{i=1}^d g_i^2 (\Delta_{\mathrm{q}}|\Delta w_i| - \Delta w_i^2) \le \Delta_{\mathrm{q}} \sum_{i=1}^d g_i^2 |\Delta w_i|$, giving the stated SNR lower bound whenever the variance is positive.
\end{proof}

\begin{proof}[Proof of Proposition~\ref{prop:tau_eta}]
In standard \bfsixteen{} (7 explicit mantissa bits), any finite normal number $w$ with $2^e \le |w| < 2^{e+1}$ has neighbors at distance $\ulp^+(w) = 2^{e-7}$ away from zero and $\ulp^-(w) = 2^{e-8}$ toward zero if $|w|=2^e$ (or $2^{e-7}$ if $|w| > 2^e$).
Under round-to-nearest-even (RN), adding $\delta w$ leaves $w$ unchanged whenever $|\delta w| < \tfrac{1}{2}\ulp_{\mathrm{dir}}(w)$, establishing (1).
For (2), let $\tau = 2^k$ and $\tau_+ = \tau + \ulp^+(\tau) = 2^k(1 + 2^{-7})$, so $\tfrac{1}{2}\ulp^+(\tau) = 2^{k-8}$.
Any \bfsixteen{} value with $|w| \ge \tau_+$ lies strictly above $2^k$ or at/above $2^{k+1}$; in all cases both its inward and outward neighbors lie at distance at least $2^{k-7}$, so both directional half-ULPs are at least $2^{k-8}$.
Thus $|\delta w_{t,i}| < 2^{k-8}$ implies $\mathrm{RN}_{\mathrm{bf16}}(w_{t,i}+\delta w_{t,i}) = w_{t,i}$ at every step $t$, freezing every coordinate with $|w_{0,i}| \ge \tau_+$.
For forward-invariance of $[-\tau, +\tau]$, take any \bfsixteen{} value $w$ with $|w| \le \tau$.
If $|w| = \tau$, stepping outward requires $|\delta w| \ge \tfrac{1}{2}\ulp^+(\tau) = 2^{k-8}$, contradicting the hypothesis, while stepping inward stays in $[-\tau, +\tau]$.
If $|w| < \tau$, then $|w| \le \tau - \ulp^-(\tau) = \tau - 2^{k-8}$; adding $|\delta w| < 2^{k-8}$ yields $|w + \delta w| < \tau$, and monotonicity of RN gives $|\mathrm{RN}_{\mathrm{bf16}}(w+\delta w)| \le \tau$.
By induction on $t = 1,\dots,T$, $|w_{T,i}| \le \tau$.
For (3), requiring strict inequality $2^{k-8} > B$ (so an odd-mantissa tie at $w = \tau_+$ with $|\delta w| = 2^{k-8}$ cannot round outward to even) gives $k > \log_2(256 B)$, whose smallest integer solution is $k = \lfloor \log_2(256 B) \rfloor + 1$, yielding $\tau(B) = 2^{\lfloor \log_2(256 B) \rfloor + 1}$.
For any given candidate step $\delta w_t$, stalling occurs strictly inside the final \bfsixteen{} writeback $\mathrm{RN}_{\mathrm{bf16}}(w_t + \delta w_t)$, so moment precision (\bfsixteen{} vs.\ \fpthirtytwo{}) leaves the writeback criterion unchanged.
\end{proof}

\begin{proof}[Proof of Theorem~\ref{thm:snapback}]
Suppose $\hwzero^{(G)} > \tau$ and $M_0 \ge \tau_+$ (which holds automatically whenever $\hwzero^{(G)} = c_0 M_0 > \tau$ with $c_0 \le (1 + 2^{-7})^{-1}$, since $M_0 > \tau / c_0 \ge \tau(1 + 2^{-7}) = \tau_+$; for power-of-two staircase scales such as MXFP4, apply the same argument to $\hwzero^{(G)} = \hwzero(M_0)$).
Let $i^* \in \arg\max_{i \in G} |w_{0,i}|$.
Since $|w_{0,i^*}| = M_0 \ge \tau_+$, Proposition~\ref{prop:tau_eta}(2) implies $w_{T,i^*} = w_{0,i^*}$ permanently.
Moreover, every coordinate $i \in G$ with $|w_{0,i}| \ge \tau_+$ is permanently frozen ($|w_{T,i}| = |w_{0,i}| \le M_0$), while every coordinate with $|w_{0,i}| \le \tau$ remains inside $[-\tau, +\tau]$ by forward-invariance, with $|w_{T,i}| \le \tau < M_0$.
(Note there are no \bfsixteen{} values strictly between $\tau$ and $\tau_+ = \tau + \ulp^+(\tau)$.)
Therefore $M_T = \max_{i \in G} |w_{T,i}| = M_0$, so the single-quantized group scale metadata $s(W_{T,G}) = s(W_{0,G})$ and all quantization Voronoi cells in group $G$ are identical before and after unlearning.
For every coordinate $i \in G$:
if $|w_{0,i}| \ge \tau_+$, then $w_{T,i} = w_{0,i}$, so deterministic encoding yields $Q_G(w_{T,i}) = Q_G(w_{0,i})$ bit-for-bit;
if $|w_{0,i}| \le \tau$, then both $|w_{0,i}| \le \tau < \hwzero^{(G)}$ and $|w_{T,i}| \le \tau < \hwzero^{(G)}$ lie strictly inside the zero Voronoi cell, so both decode to $0$ (and map to the identical code under a unique canonical zero code; under signed-zero formats such as MXFP4 or \texttt{bnb}-FP4 true zero, zero sign bits $\pm 0$ may flip while decoded real values remain identically $0$).
Hence $Q_G(W_T) = Q_G(W_0)$ numerically across all $i \in G$ (and bit-for-bit under a canonical zero code), giving $\Delta Q_G = 0$ unconditionally and $\rho_G = 0$ whenever $\Delta W_G \neq 0$.
\end{proof}

\begin{proof}[Proof of Corollary~\ref{cor:fp4_cluster}]
Since $c_{\mathrm{clus}} = \tfrac{11}{128} < (1+2^{-7})^{-1}$, $c_{\mathrm{clus}} M_0 > \tau$ implies $M_0 > \tau / c_{\mathrm{clus}} > \tau_+$.
By Proposition~\ref{prop:tau_eta}(2), every coordinate with $|w_{0,i}| \ge \tau_+$ (including the group maximum $i^* \in \arg\max_{i\in G}|w_{0,i}|$) is permanently frozen ($w_{T,i}=w_{0,i}$), while every coordinate with $|w_{0,i}| \le \tau$ stays inside $[-\tau, +\tau] \subset (-M_0, +M_0)$.
Thus $M_T = M_0$, preserving the \texttt{bnb} FP4 scale and Voronoi cells identically: frozen coordinates retain their exact FP4 nibbles, and active coordinates ($|w_{0,i}|\le\tau$) remain inside $(-c_{\mathrm{clus}} M_0, +c_{\mathrm{clus}} M_0)$, which maps exclusively to nibbles $\{\texttt{0,1,8,9}\}$.
By symmetry of the decoded levels $\{0, \pm \hat{c}_{1,+}(M_0)\}$, any transition within $\{\texttt{0,1,8,9}\}$ changes the dequantized value by at most $2\hat{c}_{1,+}(M_0)$, and since $\pm 0$ decode identically to $0$, only the two value-changing inner thresholds ($\pm M_0/384$ in real arithmetic) separate $\{0, \pm \hat{c}_{1,+}(M_0)\}$ with step size $\hat{c}_{1,+}(M_0)$.
If coordinate $i\in G$ makes $N_i \in \{0,1,2\}$ monotonic value-changing transitions between $Q_G(w_{0,i})$ and $Q_G(w_{T,i})$ in the direction of $\Delta w_i$, then $\Delta Q_{G,i} = \mathrm{sign}(\Delta w_i) N_i \hat{c}_{1,+}(M_0)$ (exact under both real arithmetic with $\hat{c}_{1,+}(M_0)=M_0/192$ and finite-precision dequantization using the actual decoded level $\hat{c}_{1,+}(M_0)$), so $\langle \Delta Q_G, \Delta W_G \rangle = \hat{c}_{1,+}(M_0)\sum_{i\in G}|\Delta w_i|N_i$, proving the exact identity for $\rho_G^{\mathrm{FP4}}$.
In ideal real arithmetic $\hat{c}_{1,+}(M_0) = M_0/192$, giving $M_0/96 \approx 0.010417 M_0$.
In machine arithmetic, $\mathrm{fl}_{32}(1/192) = \texttt{0x3BAAAAAB}$ exceeds $1/192$ by $\tfrac{1}{3}$ local ULP (relative error $2^{-25}$), so a normal \fpthirtytwo{} RN product stays within $(1+2^{-25})(1+2^{-24})M_0/96 < (1+2^{-23})M_0/96$; similarly $\mathrm{RN}_{\mathrm{bf16}}(1/192)=\texttt{0x3BAB}$ is $\tfrac{1}{3}$ local ULP high (relative error $2^{-9}$), and an additional normal \bfsixteen{} RN dequantization cast stays strictly within $(1+2^{-7})M_0/96 \approx 0.010498 M_0$ over all normal \bfsixteen{} $M_0 \ge 192 \cdot 2^{-126}$ (exhaustively verified over all $128$ significands).
\end{proof}

\begin{proof}[Proof of Proposition~\ref{prop:rho_q}]
Adding and subtracting $Q_s(W_u)$ in $\Delta Q = Q_{s'}(W_u) - Q_s(W_0)$ yields the fixed-order identity $\Delta Q = \Delta Q_{\mathrm{fix}} + \Delta Q_{\mathrm{scl}}$ (changing order reassigns the scale-weight interaction term), and taking the inner product with $\Delta W / \|\Delta W\|_2^2$ (for $\Delta W \neq 0$) gives $\rho = \rho_{\mathrm{fix}} + \rho_{\mathrm{scl}}$.
Under fixed per-group scales $s_G$ with ordered levels $c_{G,k} < c_{G,k+1}$, ordered thresholds $b_{G,k} < b_{G,k+1}$, and $|\Delta w_i| < \min_k(b_{G,k+1}-b_{G,k})$, each coordinate $i \in G$ crosses at most one threshold $b_{G,k}$ under $\mathrm{code}_{s_G}(\cdot)$, jumping by $\mathrm{sign}(\Delta w_i)(c_{G,k+1}-c_{G,k})$ so that $\Delta w_i \cdot \Delta Q_{\mathrm{fix},i} = |\Delta w_i|(c_{G,k+1}-c_{G,k})$, proving Eq.~\eqref{eq:rho_exact}.
Heuristically, let $\tilde{p}_G$ denote the globally energy-normalized smooth density of $w_{0,i}$ ($\sum_G \int \tilde{p}_G(w)\,dw = 1$, weighted by $\Delta w_i^2/\|\Delta W\|_2^2$). When step signs are locally independent of position near $b_{G,k}$, a coordinate of magnitude $|\Delta w_i|$ crosses $b_{G,k}$ with probability $|\Delta w_i|\,p_i(b_{G,k})$; substituting into Eq.~\eqref{eq:rho_exact} and summing over $i\in G$ yields $\hat{\rho}_{\mathrm{smooth}} = \sum_G \sum_k (c_{G,k+1}-c_{G,k})\,\tilde{p}_G(b_{G,k})$.
\end{proof}

\begin{remark}[Counterfactual alignment under fixed scales]
\label{rem:alignment}
For any group subset $S$ with $\Delta W_S \neq 0$, let $c_S \triangleq \Delta W_S - \Delta Q_S$, $\rho_S \triangleq \langle \Delta Q_S, \Delta W_S \rangle / \|\Delta W_S\|_2^2$, and $r_S \triangleq \|\Delta Q_S\|_2 / \|\Delta W_S\|_2$.
Then $\langle c_S, \Delta W_S \rangle = \|\Delta W_S\|_2^2 (1 - \rho_S)$ and $\|c_S\|_2^2 = \|\Delta W_S\|_2^2 (1 - 2\rho_S + r_S^2)$.
On exact-snap groups ($\Delta Q_S{=}0$), $\cos(c_S, \Delta W_S){=}1$.
Under fixed per-group scales (holding identically across our measured RN runs, where $100\%$ of scales freeze), coordinate-wise monotonicity guarantees $\rho_S \ge 0$; whenever $c_S \neq 0$ and $r_S \ge 1$, the one-sided upper bound (not ruling out negative alignment)
\begin{equation}
  \cos(c_S, \Delta W_S) = \frac{1 - \rho_S}{\sqrt{1 - 2\rho_S + r_S^2}} \le \frac{1}{\sqrt{1 + r_S^2}} < \frac{1}{r_S}
  \label{eq:cos_bound}
\end{equation}
holds (empirical measurements in Table~\ref{tab:app_jump_align} further confirm small alignment magnitude and $67\%$--$96\%$ $\Delta Q$ jump share on uncertified controls).
\end{remark}

\begin{proof}[Proof of Remark~\ref{rem:alignment}]
By definition $c_S = \Delta W_S - \Delta Q_S$, $\langle \Delta Q_S, \Delta W_S \rangle = \rho_S \|\Delta W_S\|_2^2$, and $r_S = \|\Delta Q_S\|_2 / \|\Delta W_S\|_2$.
Expanding directly gives $\langle c_S, \Delta W_S \rangle = \|\Delta W_S\|_2^2(1 - \rho_S)$ and $\|c_S\|_2^2 = \|\Delta W_S\|_2^2(1 - 2\rho_S + r_S^2)$.
When $\Delta Q_S = 0$ (exact snap-back), $\rho_S = 0$, $r_S = 0$, and $c_S = \Delta W_S$, so $\cos(c_S, \Delta W_S) = 1$ and $\|\Delta Q_S\|_2^2/\|c_S\|_2^2 = 0$.
Whenever $c_S \neq 0$, dividing the inner-product identity by $\|c_S\|_2\|\Delta W_S\|_2$ yields
$f(\rho_S) \triangleq \cos(c_S, \Delta W_S) = (1 - \rho_S) / \sqrt{1 - 2\rho_S + r_S^2}$.
By Cauchy--Schwarz, $|\rho_S| \le r_S$.
For any $r_S \ge 1$ and $\rho_S \in [0, r_S]$ (since monotonic fixed-scale scalar quantizers satisfy $\Delta Q_i \Delta W_i \ge 0$ coordinate-wise, hence $\rho_S \ge 0$), differentiating with respect to $\rho_S$ gives
$f'(\rho_S) = (\rho_S - r_S^2) / (1 - 2\rho_S + r_S^2)^{3/2} \le 0$ (strictly negative for $\rho_S < r_S^2$).
Hence $f(\rho_S)$ attains its maximum on $[0, r_S]$ at $\rho_S = 0$, proving the one-sided upper bound $\cos(c_S, \Delta W_S) \le 1 / \sqrt{1 + r_S^2} < 1/r_S$ under $r_S \ge 1$ and $\rho_S \ge 0$ (Eq.~\eqref{eq:cos_bound}).
Moreover, if $|\rho_S| \le C = O(1)$ with $1 - 2C + r_S^2 > 0$ (sparse threshold crossings), then $|\cos(c_S, \Delta W_S)| \le (1+C)/\sqrt{1-2C+r_S^2} = O(1/r_S)$.
Finally, $\|\Delta Q_S\|_2^2/\|c_S\|_2^2 = r_S^2/(1 - 2\rho_S + r_S^2) \to 1$ as $r_S \to \infty$ (potentially exceeding $1$ when $2\rho_S > 1$).
\end{proof}

\section{Implementation Details, Fused Triton Kernel, Support Definitions, and Telemetry}
\label{app:details}

\paragraph{Descriptive support metrics: Large-Weight Mobility Score ($\LMS_\tau, \LMSnorm_\tau, \LMS_Q, \LMS_{E,Q}$).}
\label{def:lms}
For linear weights $(W_0, W_u)$ evaluated at a declared dtype with $\mathcal{M}\triangleq\{i:\Delta w_i{\neq}0\}$, $f_{\mathrm{move}}\triangleq|\mathcal{M}|/d$, base empirical near-zero mass $F(\tau) \triangleq \frac{1}{d}|\{i:|w_{0,i}|{\le}\tau\}|$, and operator blind zone $\mathcal{Z}(Q)$ ($|w_{0,i}|{\le}\tfrac{11}{128}M_{0,G}$ for FP4; $[-0.04553,0.03979]M_{0,G}$ for NF4; $|w_{0,i}|{\le}0.25s_{0,G}$ for MXFP4 with zero groups in $\mathcal{Z}$; $|w_{0,i}|{\le}q_s(|W_0|)$ per layer for pruning), define for $\mathcal{M}{\neq}\emptyset$:
$\LMS_\tau \triangleq |\{i{\in}\mathcal{M}:|w_{0,i}|{>}\tau\}|/|\mathcal{M}|$, relative mobility ratio $\LMSnorm_\tau \triangleq \LMS_\tau/(1{-}F(\tau))$ (an enrichment ratio bounded above by $1/(1{-}F(\tau))$ when $1{-}F(\tau){>}0$, with observed maximum $1.23$ in our manifest when large weights move disproportionately), $\LMS_Q \triangleq |\mathcal{M}\setminus\mathcal{Z}(Q)|/|\mathcal{M}|$, and $\LMS_{E,Q} \triangleq \sum_{i\notin\mathcal{Z}(Q)}\Delta w_i^2/\|\Delta W\|_2^2$.
Because base near-zero mass $F(2^{-8})$ spans $12.17\%$ (Qwen2.5-1.5B), $17.50\%$ (Llama-2-7B), $18.48\%$ (Llama-3.2-1B), $24.86\%$ (Llama-3.1-8B), and $80.40\%$ (Zephyr-7B L5--7), $\LMSnorm_{2^{-8}}$ aligns fixed-$\tau$ scales; however, on natural runs simple $f_{\mathrm{move}}$ tracks $\LMS$ closely (pooled AUROC $0.9748$ vs.\ $0.9805$; Table~\ref{tab:app_lomo_diag}).

\paragraph{Fused Triton \bfsr{} AdamW kernel, verbatim writeback, and kernel audits.}
Inside \texttt{\_fused\_bf16\_adamw\_kernel} (Appendix~\ref{app:details}), each Triton thread block loads a \bfsixteen{} weight tile \texttt{p\_bf}, \bfsixteen{} gradient tile \texttt{g\_bf} (cast in-register to \fpthirtytwo{}), and \fpthirtytwo{} moments $(m, v)$ in one HBM pass, forms the \fpthirtytwo{} candidate \texttt{w = p\_bf.to(tl.float32)*(1-lr*wd) - (lr*inv\_bc1)*(m/(sqrt(v)*inv\_sqrt\_bc2+eps))}, and applies unbiased 16-bit mantissa rounding with per-step deterministic seed \texttt{rng\_seed = ((seed*0x9E3779B9) \^{} (t*0x85EBCA6B) \^{} (p\_idx*0xC2B2AE35)) \& 0xFFFFFFFF} before the sole \bfsixteen{} store:
\begin{quote}\scriptsize\ttfamily
x\_u32 = offs.to(tl.uint32) \^{} rng\_seed.to(tl.uint32)\\
x\_u32 = (x\_u32 \^{} (x\_u32 >> 16)) * 0x45D9F3B; \ x\_u32 = (x\_u32 \^{} (x\_u32 >> 16)) * 0x45D9F3B\\
rand16 = (x\_u32 \^{} (x\_u32 >> 16)) \& 0xFFFF\\
w\_out = ((w.to(tl.uint32, bitcast=True) + rand16) \& 0xFFFF0000).to(tl.float32, bitcast=True).to(tl.bfloat16)
\end{quote}
Because IEEE-754 \bfsixteen{} uses sign-magnitude encoding, unsigned bit addition of \texttt{rand16} is exact and unbiased for both positive and negative finite normal values away from overflow/subnormal boundaries, preserves exact zero drift when $r=0$, and adds $0.00$\,GiB persistent weight copy.
Audits on NVIDIA L4 confirm:
(i)~across $24$ signed/binade/residual buckets $\times 200\mathrm{k}$ draws ($4.8\mathrm{M}$ draws), the fused Triton kernel matches $r/65536$ within max error $0.00159$ with exact zero-residual invariance (chunked Python \texttt{sr\_round}: $\le 0.00161$);
(ii)~exact boundary enumeration at $\{\pm\tau, \pm\tau_+\}$ ($\tau=2^{-8}$) yields $0$ violations under effective shift $\delta w_t = \mathrm{fl}_{32}(w_t + \Delta^{\mathrm{cand}}_t) - w_t$ and $6$ concrete \fpthirtytwo{} double-rounding witnesses (e.g., $w{=}\tau_+ (\texttt{0x3B81}) + \Delta^{\mathrm{cand}} (\texttt{0x377FFFFF} < 2^{-16})$ rounds in \fpthirtytwo{} to the exact \bfsixteen{} midpoint $\texttt{0x3B818000}$, causing odd-mantissa RN to move to $\texttt{0x3B82}$);
(iii)~at Step~1 on Llama-3.2-1B-Instruct, pre-rounding \fpthirtytwo{} candidates between custom RN and SR match bit-for-bit across all $146/146$ tensors ($\max |\Delta| = 0.0$); and
(iv)~on real \texttt{bnb.functional.quantize\_4bit} ($20\mathrm{k}$ adversarial $g=64$ groups each), NF4 has $0$ packed \texttt{uint8}/\texttt{absmax}/dequantized violations, FP4 central cluster has $0$ bound violations ($\max |\Delta Q|/M = 0.005222$, exercising single $0{\leftrightarrow}\pm\tfrac{1}{192}$ transitions rather than worst-case $\pm\tfrac{1}{192}$ sign flips seen at $0.010444$ in full-model NPO), and FP4 true-zero has $0$ dequantized violations (with $3{,}103$ signed $\pm 0$ nibble flips \texttt{0x0}$\leftrightarrow$\texttt{0x8} that dequantize identically to $0.0$).

\paragraph{Exact quantizer definitions, 16-point sweep, and eval-split hygiene.}
Symmetric INT4 uses $s = M/7.5$ and $q = \mathrm{clip}(\mathrm{round}(w/s), -7, 7)$ ($\hwzero = M/15$).
In the 16-point TOFU-1B Sym-INT4 ($g{=}128$) sweep (Figure~\ref{fig:hero}b, single seed), both the RN curves and the matched \fpthirtytwo{}-master negative controls run the identical $s = cM/7.0$ ($\hwzero = cM/14$) encoder indexed along the x-axis by the nominal ratio $r_{\mathrm{nom}} = c\,\mathrm{med}(M/15)/\tau$ (so actual $\mathrm{med}(\hwzero/\tau) = \tfrac{15}{14}r_{\mathrm{nom}}$, placing the nominal $0.95$ peak at actual $1.02$); matched \fpthirtytwo{} controls stay flat ($0.00\%$--$0.27\%$ GradDiff [floor at $0.00\%$ unq.]; $2.36\%$--$4.81\%$ NPO), while retain degrades to $6.5\%$ at high $c$.
A separate re-evaluation of the \fpthirtytwo{} negative controls at identical clip scales $c$ under exact $s = cM/7.5$ ($\hwzero = cM/15$) likewise stays flat ($0.00\%$ GradDiff; $3.37\%$--$4.81\%$ NPO, max $+0.76$\,pp vs.\ $4.05\%$ unq.), while heavy clipping degrades retain probability ($8.91\%$--$35.09\%$ GradDiff; $16.36\%$--$48.31\%$ NPO).
Asymmetric min-max INT4 ($g=64$) uses $s = (w_{\max}-w_{\min})/15$ and $q = \mathrm{clip}(\mathrm{round}((w-w_{\min})/s), 0, 15)$.
\texttt{bitsandbytes} FP4 ($g=64$, 15 distinct values $\{0, \pm 0.00521, \pm \tfrac{1}{6}, \pm \tfrac{1}{4}, \pm \tfrac{1}{3}, \pm \tfrac{1}{2}, \pm \tfrac{2}{3}, \pm 1\}M$, central nibbles $\{\texttt{0,1,8,9}\}$) and NF4 ($g=64$, 16-level Gaussian quantiles; zero cell $[-0.04553M, +0.03979M]$, safe inner radius $0.03979 M$) run via \texttt{BitsAndBytesConfig(load\_in\_4bit=True)}.
Group-128 GPTQ \citep{frantar2023gptq} and AWQ \citep{lin2024awq} calibrate strictly on independent WikiText-2 text ($64$ 512-token sequences on TOFU/Qwen; $128$ 1024-token sequences for official MUSE GPTQ); forgetting metrics evaluate training forget sets (measuring memorization erasure), and utility metrics evaluate disjoint held-out retain splits (MUSE retain QA; TOFU \texttt{retain\_perturbed}).

\begin{figure}[htbp]
  \centering
  \includegraphics[width=0.48\linewidth]{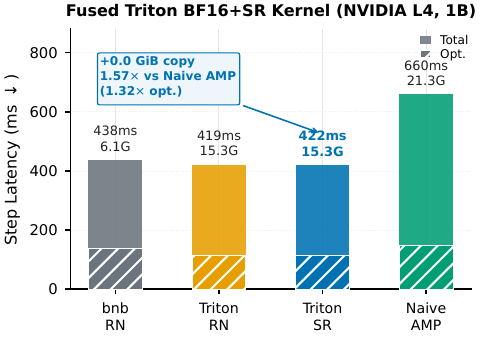}
  \caption{\textbf{Fused Triton \bfsr{} step latency and VRAM on NVIDIA L4 (TOFU-1B NPO).} Matches Triton RN ($422$ vs.\ $419$\,ms, $15.3$\,GiB) and speeds up step time $1.57\times$ ($1.32\times$ opt., $-5.98$\,GiB) over naive PyTorch \texttt{autocast} AMP, trading $+9.2$\,GiB GPU moments vs.\ CPU-paged \texttt{bnb}.}
  \label{fig:triton_latency}
\end{figure}

\begin{table}[htbp]
\centering
\caption{\textbf{Synchronized step-time and VRAM microbenchmark on NVIDIA L4 ($22.03$\,GiB VRAM, TOFU-1B NPO, micro-batch 4, grad.\ checkpointing, 6 warmup + 12 synced steps).}
Top: five-arm naive-\texttt{autocast} AMP suite. Bottom: reference Triton vs.\ fused \fpthirtytwo{}-master ($1.31\times$ opt.) and unfused/CPU/Kahan-OOM rows.}
\label{tab:app_microbench}
\scriptsize
\begin{tabular}{lccccc}
\toprule
\textbf{Optimizer / Precision Arm (True AMP Suite)} & \textbf{Weights (GiB)} & \textbf{Autocast BF16?} & \textbf{Peak VRAM (GiB)} & \textbf{Step Time (ms)} & \textbf{Opt.\ Step (ms)} \\
\midrule
\texttt{bnb\_paged32\_rn} (C++ paged \texttt{bnb}) & 2.30 & Yes & 6.09 & 438.4$\pm$1.4 & 136.8$\pm$0.1 \\
\texttt{triton\_bf16\_rn} (Fused Triton RN) & 2.30 & Yes & 15.30 & 418.7$\pm$2.3 & 113.1$\pm$0.2 \\
\texttt{triton\_bf16\_sr} (\textbf{Fused Triton SR, Ours}) & 2.30 & Yes & \textbf{15.30 (+0.00)} & \textbf{421.6$\pm$2.1} & \textbf{113.0$\pm$0.0} \\
\texttt{amp\_bf16\_fp32master\_fused} (\textbf{True AMP}) & 4.60 & Yes & 21.27 (+5.98) & 660.0$\pm$0.9 & 149.2$\pm$0.1 \\
\texttt{pure\_fp32\_fused} (Pure FP32 compute) & 4.60 & No & 21.37 (+6.07) & 1159.5$\pm$9.2 & 149.3$\pm$0.0 \\
\midrule
\texttt{triton\_bf16\_sr} / \texttt{fp32\_master\_fused} (\texttt{ref}) & 2.30 / 4.60 & Yes & 14.99 / 19.41 & 407.1 / 581.4 & 113.7 / 149.2 \\
\texttt{bf16\_sr\_gpu} (unfused) / \texttt{cpu\_mom} / \texttt{kahan\_gpu} & 2.30 & Yes & 17.98 / 10.73 / \textbf{OOM} & 1061.1 / 5094.0 / --- & 739.6 / 4782.6 / --- \\
\bottomrule
\end{tabular}
\end{table}

\begin{table}[htbp]
\centering
\caption{\textbf{Weight-space alignment, two-seed SR noise floor, and SURE in-mask freeze audit (TOFU-1B NPO, 5~ep., $\mathrm{eff\_bs}{=}16$, 125 st., 112 linear layers, seed~0).}
Top: deterministic \texttt{Kahan-BF16} tracks \fpthirtytwo{} closely ($\cos{=}0.902$, $1.01\times$ RMS), whereas \bfsr{} injects a $1.447\times\|\Delta W_{\mathrm{FP32}}\|_2$ random-walk noise floor (pairwise SR--SR $\cos{=}0.292$; 2-seed average $\cos{=}0.457$, RMS $1.38\times$). Bottom: at $\eta{=}10^{-5}$, BF16 RN freezes $92.0\%$ of SURE's unmasked weights; raising $\eta$ to $3{\times}10^{-5}$ triples in-mask movement ($8.0\%{\to}24.4\%$).}
\label{tab:app_w13_align}
\scriptsize
\resizebox{\textwidth}{!}{%
\begin{tabular}{lccccccc}
\toprule
\textbf{NPO Arm (Identical Batch Order, $s_0$)} & \textbf{$f_{\mathrm{move}}$ (\%)} & \textbf{RMS ($\times 10^{-5}$, vs.\ FP32)} & \textbf{$\cos(\cdot,\Delta W_{\mathrm{FP32}})$} & \textbf{Proj.\ along FP32} & \textbf{Unq.\ F / R} & \textbf{FP4 F / R ($\Delta F$)} & \textbf{NF4 F / R} \\
\midrule
\texttt{fp32\_master} ($\eta{=}10^{-5}$, reference) & 61.20 & 7.33 (1.00$\times$) & \textbf{1.000} & 1.000 & 4.46 / 74.65 & 5.06 / 49.16 (+0.60) & 5.37 / 61.61 \\
\texttt{bf16\_rn} ($\eta{=}10^{-5}$) & 8.62 & 2.97 (0.40$\times$) & 0.155 & 0.063 & 22.80 / 31.50 & \textbf{27.30} / 28.88 (\textbf{+4.50}) & 23.99 / 28.23 \\
\texttt{kahan\_bf16} ($\eta{=}10^{-5}$, zero-noise) & \textbf{61.48} & \textbf{7.40 (1.01$\times$)} & \textbf{0.902} & \textbf{0.910} & 4.32 / 74.20 & \textbf{4.90} / 48.44 (\textbf{+0.58}) & 5.15 / 61.04 \\
\texttt{bf16\_sr\_rng0} ($\eta{=}10^{-5}$, hash seed 0) & 72.52 & 12.60 (1.72$\times$) & 0.374 & 0.643 & 4.01 / 77.26 & \textbf{4.49} / 48.41 (\textbf{+0.48}) & 4.71 / 62.43 \\
\texttt{bf16\_sr\_rng1} ($\eta{=}10^{-5}$, hash seed 1) & 72.52 & 12.61 (1.72$\times$) & 0.360 & 0.620 & 3.78 / 77.50 & \textbf{4.13} / 47.62 (\textbf{+0.35}) & 4.50 / 61.81 \\
\midrule
\textbf{SURE Top-10\% In-Mask Audit} & \textbf{Global $f_{\mathrm{move}}$} & \textbf{Ever-Unmasked (\%)} & \textbf{In-Mask Moved} & \textbf{In-Mask Frozen} & \textbf{Unq.\ F / R} & \textbf{FP4 F / R ($\Delta F$)} & \textbf{NF4 F / R} \\
\midrule
\texttt{sure\_top10\_rn} ($\eta{=}10^{-5}$) & 6.06\% & 76.05\% & \textbf{7.96\%} & \textbf{92.04\%} & 24.16 / 25.09 & \textbf{30.21} / 29.49 (\textbf{+6.05}) & 27.19 / 27.07 \\
\texttt{sure\_top10\_rn} ($\eta{=}3{\times}10^{-5}$) & 19.18\% & 78.52\% & \textbf{24.42\%} & \textbf{75.58\%} & 14.44 / 59.38 & \textbf{14.28} / 36.63 (\textbf{$-$0.16}) & 15.29 / 47.49 \\
\bottomrule
\end{tabular}}
\end{table}

\begin{table}[htbp]
\centering
\caption{\textbf{Code-to-paper provenance, eval-split role, and measured resource telemetry ($95$\,GiB / $80$\,GiB / $22$\,GiB GPUs).}}
\label{tab:app_provenance}
\scriptsize
\resizebox{\textwidth}{!}{%
\begin{tabular}{lllllll}
\toprule
\textbf{Suite / Arm (Table)} & \textbf{Weights / Mom.} & \textbf{Write.} & \textbf{PTQ Backend \& Calib.} & \textbf{Eval Split Roles} & \textbf{Peak VRAM / RSS} & \textbf{Train s} \\
\midrule
MUSE-News \texttt{official\_bf16\_rn} (Tab.~\ref{tab:factorial}) & BF16 / BF16 & RN & bnb FP4/NF4; GPTQ (WikiText-2) & Forget (mem.) / Retain (disjoint) & --- (HF Trainer) & 3608--3613 \\
MUSE-News \texttt{fp32\_mom\_bf16\_rn} (Tab.~\ref{tab:factorial}) & BF16 / FP32 (pg.) & RN & Same as above & Same & 52.8 / 52.7 GiB & 5338 mean (s0: 4709; s1--2: 5653) \\
MUSE-News \texttt{official\_bf16\_sr} (Tab.~\ref{tab:factorial}) & BF16 / BF16 & SR & Same as above & Same & --- (HF Trainer) & 3869--3887 \\
MUSE-News \texttt{official\_fp32\_master} (Tab.~\ref{tab:factorial}) & FP32 / FP32 & FP32 & Same + BF16 export & Same & --- (masked \fpthirtytwo{}: 89.8/17.5 GiB) & 6015--6034 \\
\midrule
TOFU-8B \texttt{default paged32} (Tab.~\ref{tab:tofu_ecosystem}) & BF16 / FP32 & RN & bnb FP4/NF4 ($g=64$) & \texttt{forget10} / \texttt{retain\_pert.} & 41.7 / 63.6 GiB & 956 \\
TOFU-8B \texttt{fp32\_mom\_bf16\_sr} (Tab.~\ref{tab:tofu_ecosystem}) & BF16 / FP32 & SR & Same as above & Same & 48.1 / 63.6 GiB & 1002 \\
TOFU-8B \texttt{fp32\_master} (Tab.~\ref{tab:tofu_ecosystem}) & FP32 / FP32 & FP32 & Same + BF16 export & Same & 74.6 / 63.7 GiB & 1561 \\
\midrule
TOFU-1B \& Qwen2.5-1.5B Multi-Seed Suite (Tables~\ref{tab:tofu_ecosystem},~\ref{tab:w5_controls}) & BF16 / FP32 & Triton RN/SR & FP4, NF4, GPTQ-128, AWQ-128 & Same ($n{=}3$ RN/SR; $n{=}1$ FP32) & 5.78--19.41 GiB (L4) & 300--2028 \\
TOFU-1B Strict Zero-BF16 Causal Suite (Tab.~\ref{tab:app_eqcard_sextile}) & FP32 / FP32 & Direct FP32 & Direct FP32$\to$FP4/NF4 & TOFU $n{=}3$ NPO + $s_0$ GD (\texttt{allow\_tf32=False}) & 24.1 GiB (A100-80G) & 364--419 \\
WMDP-Cyber 5-LR $\times$ 3-Seed Sweep (Tab.~\ref{tab:app_wmdp}) & BF16 (+FP32 oracle) / FP32 & Triton RN/SR; FP32 & bnb FP4/NF4 ($g=64$) & WMDP-Cyber 1,987 + MMLU 500 ($n{=}3$) & 25.8 GiB (A100-80G) & 196--324 \\
Step-1 Writeback \& Warmup Ablation (Tab.~\ref{tab:app_warmup}) & BF16 / FP32 & Pure BF16 RN & bnb FP4 ($g=64$, TOFU) & TOFU-1B (125 st.) \& WMDP-7B L5--7 (100 st.), $s_0$ & 19.2 GiB (L4-24G) & 298--456 \\
Cross-Operator Universal Suite (Suite~B Surgery, Tab.~\ref{tab:app_universal}) & FP32 / FP32 & Direct FP32 & FP4, NF4, MXFP4 ($g{=}32$), Prune 20/30\% & TOFU-1B strict zero-BF16/TF32 ($n{=}3$, $\mathrm{eff\_bs}{=}16$, 125 st., $\mathrm{wd}{=}0$) & 20.3 GiB (L4-24G) & 568--770 \\
Cross-Operator Natural TOFU-1B (Tab.~\ref{tab:app_universal}) & BF16 / FP32 & Triton RN/SR; FP32 & FP4, NF4, MXFP4 ($g{=}32$), Prune 20/30\% & TOFU-1B natural ($n{=}9$, $\mathrm{eff\_bs}{=}16$, 5 ep./125 st., linear layers) & 18.4 GiB (L4-24G) & 209--450 \\
Refusal-Template SFT, WMDP \& Prop.~\ref{prop:rho_q} Audit (Tabs.~\ref{tab:safety_universal},~\ref{tab:app_universal},~\ref{tab:app_prop2_mediation}) & BF16 (+FP32) / FP32 & Triton RN/SR; FP32 & FP4, NF4, MXFP4, Prune 20/30\% & Zephyr-7B ($n{=}5$); Qwen2.5 Clean Refusal SFT ($n{=}3$); Prop.~\ref{prop:rho_q} ($s_0$) & 18.8--25.8 GiB (L4/A100) & 147--620 \\
\bottomrule
\end{tabular}}
\end{table}

\begin{table}[htbp]
\centering
\caption{\textbf{Full commit-pinned precision audit of seven public LLM unlearning repositories (five distinct trainers; companion to Table~\ref{tab:ecosystem_audit}).}}
\label{tab:app_audit_full}
\scriptsize
\renewcommand{\arraystretch}{0.86}
\resizebox{\textwidth}{!}{%
\begin{tabular}{lllllll}
\toprule
\textbf{Repository (Pinned Commit)} & \textbf{Entrypoint (File:Lines)} & \textbf{Load / Mom.} & \textbf{Default $\eta$} & \textbf{FP32 Master?} & \textbf{Default Storage} & \textbf{Unchanged @ Default [Meas./Inf.]} \\
\midrule
\texttt{FailureLLMUnlearning} (\texttt{10131ae25f}) & \texttt{iterative.py:60-62}, \texttt{utils.py:76} & \texttt{bf16} / \texttt{adamw} (BF16) & $10^{-5}$ & No (no DS) & Pure BF16 RN & \textbf{82.9\%} \textbf{[Meas.]} (MUSE-News 7B) \\
\texttt{muse\_bench} (\texttt{fcd6651da8}) & \texttt{iterative.py:54-56}, \texttt{utils.py:76} & \texttt{bf16} / \texttt{adamw} (BF16) & $10^{-5}$ & No (no DS) & Pure BF16 RN & 82.9\% \emph{[Inf.]} (shared trainer) \\
\texttt{Unlearn-Simple} MUSE (\texttt{3083017cb3}) & \texttt{MUSE/.../iterative.py:60-62} & \texttt{bf16} / \texttt{adamw} (BF16) & $10^{-5}$ & No (no DS) & Pure BF16 RN & 82.9\% \emph{[Inf.]} (shared trainer) \\
\texttt{Unlearn-Simple} WMDP (\texttt{3083017cb3}) & \texttt{WMDP/.../unlearn.py:100,177,182} & \texttt{bf16} / \texttt{adamw} (BF16) & $10^{-5}$ & No (no DS) & Pure BF16 RN & 21.8\% \textbf{[Meas.]} (L5--7; $F(2^{-8}){=}80.4\%$) \\
\texttt{wmdp} RMU (\texttt{c0b6c12bb0}) & \texttt{rmu/utils.py:50-54}, \texttt{unlearn.py:26} & \texttt{"auto"}$\to$\texttt{bf16} / raw & $5{\times}10^{-5}$ & No (no DS) & Pure BF16 RN & 1.3\% \textbf{[Meas.]} (L5--7; safe @ $5{\times}10^{-5}$) \\
\texttt{open-unlearning} 1-GPU (\texttt{838cc3c091}) & \texttt{README.md:154}, \texttt{finetune.yaml:6-15} & \texttt{bf16} / \texttt{paged32} & $10^{-5}$+decay & No (no DS) & Pure BF16 RN & \textbf{88.0--92.0\%} [Repro.] (Qwen Triton-RN 94.8\%) \\
\texttt{open-unlearning} Multi-GPU & \texttt{configs/accelerate/default\_config.yaml} & \texttt{bf16} / \texttt{paged32} & $10^{-5}$+decay & \textbf{Yes (ZeRO-3)} & FP32 Master & No BF16 stall \emph{[Inf.]} (launcher split) \\
\texttt{tofu} (\texttt{bdc209ea7c}) / \texttt{npo} (\texttt{1207bbe9a1}) & \texttt{forget.py:100-105} / \texttt{TOFU/forget.py:129-138} & \texttt{bf16} / \texttt{paged32} & $10^{-5}$ & \textbf{Yes (shipped DS)} & FP32 (DS) & No BF16 stall \emph{[Inf.]} (RN w/o DS) \\
\bottomrule
\end{tabular}}
\end{table}

\clearpage
\section{Online Certification, Alignment Audit, SURE Controls, and Trajectories}
\label{app:thm1_kappa}

\begin{figure}[htbp]
  \centering
  \includegraphics[width=0.92\linewidth]{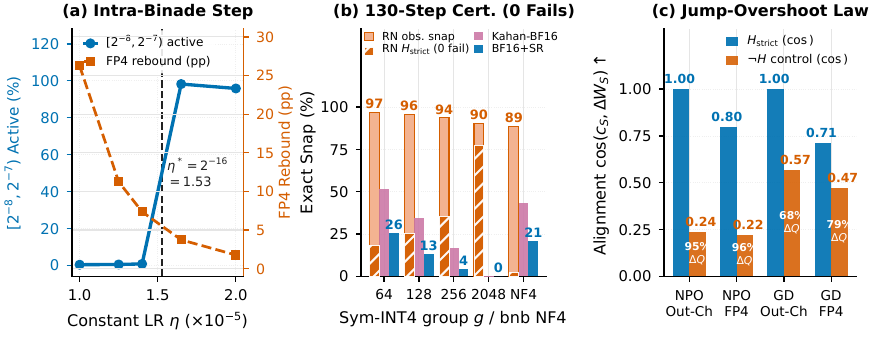}
  \caption{\textbf{Intra-binade jump, online 130-step consistency, and jump-overshoot alignment (TOFU-1B).}
  (a)~Sweeping constant $\eta$ across $\eta^*{=}2^{-16}$ in the 10-epoch sweep (Table~\ref{tab:app_supp_controls}) triggers a $131\times$ activation jump in $[2^{-8},2^{-7})$ ($0.75\%{\to}98.26\%$; distinct from Table~\ref{tab:app_warmup}'s 5-epoch band $(2^{-8},2^{-7}]$) while FP4 rebound decreases across the sweep.
  (b)~Online 130-step implementation consistency check across 112 layers finds zero counterexamples in certified groups (hatched) while \bfsr{} cuts $g{=}2048$ snap-back to $0.00\%$ (NPO) / $0.02\%$ (GradDiff).
  (c)~Certified-group backfills align with $\Delta W$ ($\cos{=}0.71$--$1.00$), while dose-matched uncertified controls mainly cancel $\Delta Q$ overshoot ($\|\Delta Q\|_2^2/\|c\|_2^2{=}0.67$--$0.96$).}
  \label{fig:theorem1}
\end{figure}

\begin{table}[htbp]
\centering
\caption{\textbf{Online 130-step Theorem~\ref{thm:snapback} / Corollary~\ref{cor:fp4_cluster} $2\times 2$ consistency tables on TOFU-1B (\bfsixteen{}-RN, $\eta{=}10^{-5}$, 112 linear layers).}
Tracks online $\max_{t\le 130,i\in G}|\delta w_{t,i}| < 2^{-16}$ and $\hwzero > 2^{-8}$ ($H_{\mathrm{strict}}$) across up to $15.2\mathrm{M}$ examined groups (non-independent parameter groups, consistency-checking the arithmetic bounds) against exact snap-back and real packed \texttt{bnb} codes, yielding zero counterexamples ($N(H_{\mathrm{strict}}\wedge\neg\mathrm{ok}){=}0$) under Corollary~\ref{cor:fp4_cluster}'s post-hoc $(1{+}2^{-7})M_0/96$ machine bound (added after observing $5{,}646$ NPO and $1{,}330$ GradDiff tiny \bfsixteen{} dequantization excesses vs.\ the ideal real bound $M_0/96$, up to $0.010444 M_0$). \emph{Row predicates (seed~0):} Sym-INT4 True Out-Ch omits energy ok\%; FP4 cluster-nib rows report the changed-coordinate predicate (under the stricter all-in-band predicate, NPO $\neg H$ splits $7{,}333{,}633 / 315{,}252$, yielding $97.93\%$ ok under the relaxed machine bound and $97.89\%$ under the un-relaxed $M_0/96$ bound after excluding the $5{,}646$ excesses; standalone exact packed-nibble checks cover NPO only, not GradDiff).}
\label{tab:app_contingency}
\scriptsize
\resizebox{\textwidth}{!}{%
\begin{tabular}{llrrrrrrrr}
\toprule
\textbf{Method} & \textbf{Quantizer Backend} & \textbf{Total Groups} & $N(H_{\mathrm{strict}})$ & $N(H \wedge \mathrm{ok})$ & $N(H \wedge \neg\mathrm{ok})$ & $N(\neg H \wedge \mathrm{ok})$ & $N(\neg H \wedge \neg\mathrm{ok})$ & $N_{\mathrm{never}}$ / $N_{\mathrm{ret}}$ & \textbf{Obs.\ ok (\% / E\%)} \\
\midrule
\multirow{8}{*}{NPO RN}
 & Sym-INT4 $g=64$ & 15,204,352 & 2,751,469 & 2,751,469 & \textbf{0} & 11,977,713 & 475,170 & 110,379 / 4,767 & 96.87 / 93.09 \\
 & Sym-INT4 $g=128$ & 7,602,176 & 1,935,160 & 1,935,160 & \textbf{0} & 5,331,796 & 335,220 & 2,589 / 167 & 95.59 / 91.11 \\
 & Sym-INT4 $g=256$ & 3,801,088 & 1,337,631 & 1,337,631 & \textbf{0} & 2,219,673 & 243,784 & 34 / 3 & 93.59 / 88.14 \\
 & Sym-INT4 $g=2048$ (fixed blocks) & 475,136 & 367,062 & 367,062 & \textbf{0} & 61,848 & 46,226 & 0 / 0 & 90.27 / 84.38 \\
 & Sym-INT4 True Out-Ch (whole row) & 376,832 & 296,531 & 296,531 & \textbf{0} & 46,477 & 33,824 & --- & 91.02 / --- \\
 & Real \texttt{bnb} NF4 $g=64$ (packed) & 15,204,352 & 385,991 & 385,991 & \textbf{0} & 13,096,780 & 1,721,581 & --- & 88.68 / 81.09 \\
 & Real \texttt{bnb} FP4 $g=64$ (true zero) & 15,204,352 & 0 & 0 & \textbf{0} & 10,905,211 & 4,299,141 & --- & 71.72 / 58.46 \\
 & Real \texttt{bnb} FP4 $g=64$ (cluster nib.) & 15,204,352 & 7,555,467 & 7,555,467 & \textbf{0} & 7,369,136 & 279,749 & $\max\frac{|\Delta Q|}{M}{=}0.01044$ & 98.16 / 95.37 \\
\midrule
\multirow{8}{*}{GradDiff RN}
 & Sym-INT4 $g=64$ & 15,204,352 & 2,751,469 & 2,751,469 & \textbf{0} & 12,307,048 & 145,835 & 166,397 / 7,775 & 99.04 / 97.61 \\
 & Sym-INT4 $g=128$ & 7,602,176 & 1,935,160 & 1,935,160 & \textbf{0} & 5,592,870 & 74,146 & 4,234 / 263 & 99.02 / 97.63 \\
 & Sym-INT4 $g=256$ & 3,801,088 & 1,337,631 & 1,337,631 & \textbf{0} & 2,423,318 & 40,139 & 53 / 0 & 98.94 / 97.57 \\
 & Sym-INT4 $g=2048$ (fixed blocks) & 475,136 & 367,062 & 367,062 & \textbf{0} & 100,937 & 7,137 & 0 / 0 & 98.50 / 97.28 \\
 & Sym-INT4 True Out-Ch (whole row) & 376,832 & 296,531 & 296,531 & \textbf{0} & 74,665 & 5,636 & --- & 98.50 / --- \\
 & Real \texttt{bnb} NF4 $g=64$ (packed) & 15,204,352 & 385,991 & 385,991 & \textbf{0} & 13,830,420 & 987,941 & --- & 93.50 / 88.95 \\
 & Real \texttt{bnb} FP4 $g=64$ (true zero) & 15,204,352 & 0 & 0 & \textbf{0} & 11,668,383 & 3,535,969 & --- & 76.74 / 63.49 \\
 & Real \texttt{bnb} FP4 $g=64$ (cluster nib.) & 15,204,352 & 7,555,467 & 7,555,467 & \textbf{0} & 7,590,450 & 58,435 & --- & 99.62 / 98.86 \\
\bottomrule
\end{tabular}}
\end{table}

\begin{table}[htbp]
\centering
\caption{\textbf{Dose-matched group-to-behavior counterfactual backfill on TOFU-1B (\bfsixteen{}-RN $\eta{=}10^{-5}$, seed~0).}
Evaluated in dense \bfsixteen{} dequantized hybrid mode over 112 linear layers ($K{=}5$ draws matching per-layer correction dose $D(S){=}\|\Delta W_S{-}\Delta Q_S\|_2^2/\|\Delta W{-}\Delta Q\|_2^2$ or update energy $E_{\Delta W}(S)$, aggregate imbalance $\le 0.215\%$). Recovery share $\mathrm{sh.}\% \triangleq (F_{\mathrm{None}} - F_{+H}) / (F_{\mathrm{None}} - F_{\mathrm{Full}})$ measures the fraction of the full-backfill drop recovered by restoring only certified $H_{\mathrm{strict}}$ groups. FP4 $+\neg H{\wedge}\mathrm{ok}$ uses the strict all-in-band predicate ($62.6\%$ / $62.8\%$ OK; under Table~\ref{tab:app_contingency}'s changed-coordinate predicate, $+\neg H{\wedge}\mathrm{ok}$ gives $17.27/25.91$ NPO and $6.32/8.80$ GradDiff). Note the important matched-energy exception: at matched $E_{\Delta W}$, GradDiff FP4 gives $22.79$ ($H$) vs.\ $22.54$ ($\neg H$), underscoring that dose-matched vs.\ energy-matched predicates answer distinct mechanistic questions.}
\label{tab:app_cf}
\scriptsize
\resizebox{\textwidth}{!}{%
\begin{tabular}{lccccccc}
\toprule
\textbf{Method / Quantizer} & \textbf{Pure $Q(W_0)$} & \textbf{$Q(W_u)$ None} & \textbf{+$H_{\mathrm{strict}}$ Full} & \textbf{+$\neg H{\wedge}\mathrm{ok}$ Alone} & \textbf{Full $Q(W_0){+}\Delta W$} & \textbf{Match $D(S)$ ($K{=}5$): $H$ vs.\ $\neg H$ [$D\%$, $\|c\|$]} & \textbf{Match $E_{\Delta W}$ ($K{=}5$): $H$ vs.\ $\neg H$ [$E\%$]} \\
 & \textbf{F / R} & \textbf{F / R} & \textbf{F / R [sh.\%]} & \textbf{F / R} & \textbf{F / R} & \textbf{F$\pm$SD / R (both)} & \textbf{F$\pm$SD / R (both)} \\
\midrule
NPO $g{=}64$ (Unq.\ 21.26/41.06) & 50.89 / 50.36 & 24.95 / 27.26 & 23.40 / 26.70 [19.0\%] & 12.84 / 21.86 & 16.78 / 28.81 & \textbf{23.40$\pm$0.00}/26.7 vs.\ 24.89$\pm$0.02/27.3 [1.31\%, 0.41] & \textbf{23.45$\pm$0.00}/26.7 vs.\ 24.43$\pm$0.07/27.7 [12.15\%] \\
NPO $g{=}2048$ (fixed blocks) & 20.64 / 21.13 & 25.51 / 26.01 & \textbf{15.04} / 19.56 [85.1\%] & 22.90 / 24.49 & 13.21 / 19.21 & \textbf{15.45$\pm$0.01}/19.8 vs.\ 25.11$\pm$0.06/26.0 [12.78\%, 0.92] & \textbf{20.11$\pm$0.02}/22.3 vs.\ 23.36$\pm$0.03/25.6 [23.84\%] \\
NPO True Out-Ch (whole row) & 21.76 / 22.12 & 25.82 / 26.34 & \textbf{13.78} / 18.12 [89.7\%] & 23.74 / 25.03 & 12.39 / 17.77 & \textbf{17.24$\pm$0.00}/20.3 vs.\ 25.55$\pm$0.04/26.4 [11.72\%, 0.80] & \textbf{20.97$\pm$0.02}/22.6 vs.\ 24.26$\pm$0.04/26.0 [17.50\%] \\
NPO Real \texttt{bnb} FP4 $g{=}64$ & 50.87 / 50.10 & 27.62 / 30.74 & \textbf{22.01} / 28.45 [57.8\%] & 25.33 / 29.89 & 17.92 / 30.29 & \textbf{22.04$\pm$0.01}/28.5 vs.\ 27.37$\pm$0.03/31.0 [5.71\%, 0.73] & \textbf{22.53$\pm$0.02}/28.6 vs.\ 25.50$\pm$0.05/32.2 [33.74\%] \\
\midrule
GradDiff $g{=}64$ (Unq.\ 4.41/31.95) & 50.89 / 50.36 & 31.98 / 34.30 & 26.76 / 30.16 [18.1\%] & 2.12 / 3.05 & 3.12 / 4.91 & \textbf{26.81$\pm$0.03}/30.1 vs.\ 30.69$\pm$0.36/33.4 [3.27\%, 0.29] & 27.19$\pm$0.04/30.5 vs.\ 27.56$\pm$0.09/31.0 [12.38\%] \\
GradDiff $g{=}2048$ (fixed blocks) & 20.64 / 21.13 & 17.68 / 18.21 & \textbf{4.86} / 5.97 [80.0\%] & 9.40 / 10.48 & 1.65 / 2.17 & \textbf{7.29$\pm$0.05}/8.4 vs.\ 14.85$\pm$0.06/15.8 [29.82\%, 0.56] & \textbf{11.24$\pm$0.07}/12.2 vs.\ 12.90$\pm$0.06/14.0 [24.35\%] \\
GradDiff True Out-Ch (whole row) & 21.76 / 22.12 & 19.88 / 20.26 & \textbf{6.20} / 7.26 [79.1\%] & 13.74 / 14.61 & 2.58 / 3.13 & \textbf{13.67$\pm$0.02}/14.8 vs.\ 18.18$\pm$0.06/18.8 [22.69\%, 0.47] & \textbf{15.69$\pm$0.04}/16.5 vs.\ 16.62$\pm$0.06/17.4 [18.07\%] \\
GradDiff Real \texttt{bnb} FP4 $g{=}64$ & 50.87 / 50.10 & 34.90 / 37.45 & \textbf{21.17} / 26.03 [44.8\%] & 29.13 / 32.80 & 4.28 / 6.54 & \textbf{21.54$\pm$0.03}/26.4 vs.\ 29.57$\pm$0.10/33.4 [18.75\%, 0.54] & 22.79$\pm$0.03/27.5 vs.\ 22.54$\pm$0.22/27.5 [33.88\%] \\
\bottomrule
\end{tabular}}
\end{table}

\begin{table}[htbp]
\centering
\caption{\textbf{Exact 112-layer $\times$ $K{=}5$ dose-balance audit and Remark~\ref{rem:alignment} alignment check (TOFU-1B).}
Aggregate dose imbalance stays $\le 0.215\%$ (median layer $0.086\%$--$0.565\%$, p95 $3.2\%$--$32.5\%$ where discrete group caps bind); Jump Ratio is $\|\Delta Q_S\|_2^2/\|c_S\|_2^2$, and Align.\ Ratio is $\langle c_H,\Delta W_H\rangle / \langle c_{\neg H},\Delta W_{\neg H}\rangle$.}
\label{tab:app_jump_align}
\scriptsize
\resizebox{\textwidth}{!}{%
\begin{tabular}{llccccccccc}
\toprule
\textbf{Method} & \textbf{Quantizer} & \textbf{Tot.\ Imb.} & \textbf{Med.\ / p95 Layer Imb.} & \textbf{BF16 Norm Ratio ($H$/$\neg H$)} & \textbf{$H$: $\cos(c,\Delta W)$} & \textbf{$H$: Jump Ratio} & \textbf{$\neg H$: $\cos(c,\Delta W)$} & \textbf{$\neg H$: Jump Ratio} & \textbf{Align.\ Ratio} \\
\midrule
\multirow{4}{*}{NPO}
 & Sym-INT4 $g{=}64$ & 0.080\% & 0.565\% / 21.1\% & 1.00000 / 1.00002 & \textbf{1.0000} & \textbf{0.0\%} & 0.2555 & \textbf{93.9\%} & \textbf{12.08$\times$} \\
 & Sym-INT4 $g{=}2048$ & 0.157\% & 0.420\% / 25.3\% & 1.00000 / 0.99999 & \textbf{1.0000} & \textbf{0.0\%} & 0.2353 & \textbf{94.8\%} & \textbf{14.66$\times$} \\
 & Sym-INT4 True Out-Ch & 0.214\% & 0.502\% / 32.5\% & 1.00000 / 1.00000 & \textbf{1.0000} & \textbf{0.0\%} & 0.2373 & \textbf{94.7\%} & \textbf{14.43$\times$} \\
 & Real \texttt{bnb} FP4 $g{=}64$ & 0.050\% & 0.154\% / 3.2\% & 0.99998 / 1.00000 & \textbf{0.7973} & 40.7\% & 0.2218 & \textbf{96.2\%} & \textbf{10.99$\times$} \\
\midrule
\multirow{4}{*}{GradDiff}
 & Sym-INT4 $g{=}64$ & 0.093\% & 0.399\% / 23.6\% & 1.00000 / 1.00001 & \textbf{1.0000} & \textbf{0.0\%} & 0.4911 & \textbf{76.0\%} & \textbf{3.91$\times$} \\
 & Sym-INT4 $g{=}2048$ & 0.090\% & 0.091\% / 13.0\% & 1.00000 / 0.99999 & \textbf{1.0000} & \textbf{0.0\%} & 0.5737 & \textbf{67.1\%} & \textbf{2.98$\times$} \\
 & Sym-INT4 True Out-Ch & 0.072\% & 0.208\% / 17.7\% & 1.00000 / 1.00000 & \textbf{1.0000} & \textbf{0.0\%} & 0.5664 & \textbf{67.9\%} & \textbf{3.05$\times$} \\
 & Real \texttt{bnb} FP4 $g{=}64$ & 0.035\% & 0.086\% / 5.1\% & 0.99999 / 1.00001 & \textbf{0.7119} & 55.3\% & 0.4731 & \textbf{79.1\%} & \textbf{2.43$\times$} \\
\bottomrule
\end{tabular}}
\end{table}

\begin{table}[htbp]
\centering
\caption{\textbf{Descriptive movement-support diagnostic audit across our 218-row checkpoint manifest.}
Evaluates $f_{\mathrm{move}}^{\mathrm{tr}}$ and secondary ratios $\LMSnorm_{2^{-8}}$, $\LMS_{\mathrm{FP4}}$, and $\LMS_{E,\mathrm{FP4}}$ (Appendix~\ref{app:details}) at detecting paired FP4 rebound at both $\ge 5$\,pp and $\ge 3$\,pp across two TOFU evaluation pools (all $N{=}114$ TOFU checkpoints; $N{=}67$ binade-logged non-SimNPO checkpoints), Leave-One-Model-Out (LOMO) folds, Zephyr-7B WMDP-Cyber ($N{=}13$), in-sample surgeries, and the 23-run cross-domain suite. Because manifest checkpoints are correlated and partly selected, these metrics summarize internal consistency rather than independent deployment calibration.}
\label{tab:app_lomo_diag}
\scriptsize
\resizebox{\textwidth}{!}{%
\begin{tabular}{llcccc}
\toprule
\textbf{Evaluation Pool / Held-Out Fold} & \textbf{Diagnostic / Baseline (Individual)} & \textbf{N (Pos $\ge 5$\,pp)} & \textbf{AUROC ($\ge 5$\,pp / $\ge 3$\,pp)} & \textbf{Spearman $\rho(\cdot, \Delta_{\mathrm{FP4}})$} & \textbf{Held-Out Acc.\ (TP/FP/TN/FN)} \\
\midrule
All manifest TOFU checkpoints (incl.\ SimNPO stall) & $f_{\mathrm{move}}$ alone & 114 (37) & \textbf{0.964} / \textbf{0.971} & $-$0.596 & --- \\
Logged 1B+8B runs with $\mathrm{RMS}(\Delta W)$ (incl.\ SimNPO) & $f_{\mathrm{move}}$ vs.\ $\mathrm{RMS}(\Delta W)$ & 48 (13) & \textbf{0.9538} vs.\ 0.9143 & $-$0.343 vs.\ $-$0.292 & --- \\
Logged 1B+8B runs excl.\ SimNPO (NPO/GradDiff/RMU) & $\LMSnorm_{2^{-8}}$ ($\LMS_{2^{-8}}$) vs.\ $\mathrm{RMS}(\Delta W)$ & 39 (13) & 1.0000 vs.\ 1.0000 & $-$0.654 ($-$0.711) vs.\ $-$0.656 & --- \\
All binade-logged TOFU checkpoints (excl.\ SimNPO) & $\LMSnorm_{2^{-8}}$ / $f_{\mathrm{move}}$ & 67 (25) & \textbf{0.9805} [0.933,1.000] / 0.9748 (\textbf{0.9972} / 0.9889) & $-$0.676 ($-$0.687 raw) / $-$0.695 & --- \\
\midrule
LOMO Held-out Llama-3.1-8B (train thr.\ $\LMSnorm_{2^{-8}}\le 0.136$) & $\LMSnorm_{2^{-8}}$ (fit on 1B + Qwen2.5) & 12 (4) & \textbf{1.0000} (1.0000 $f_{\mathrm{move}}$) & --- & \textbf{100.0\%} (4 / 0 / 8 / \textbf{0}) \\
LOMO Held-out Llama-3.2-1B (train thr.\ $\LMSnorm_{2^{-8}}\le 0.437$) & $\LMSnorm_{2^{-8}}$ (fit on 8B + Qwen2.5) & 41 (16) & \textbf{1.0000} (0.9850 $f_{\mathrm{move}}$) & --- & \textbf{95.1\%} (16 / 2 / 23 / \textbf{0}) \\
LOMO Held-out Qwen2.5-1.5B (train thr.\ $\LMSnorm_{2^{-8}}\le 0.136$) & $\LMSnorm_{2^{-8}}$ (fit on 1B + 8B) & 14 (5) & 0.9444 (\textbf{0.9667} $f_{\mathrm{move}}$) & --- & \textbf{92.9\%} (5 / 1 / 8 / \textbf{0}) \\
\midrule
WMDP-Cyber Zephyr-7B ($10^{-5}, 1.4{\times}10^{-5}$, $F{=}80.4\%$) & $\LMSnorm_{2^{-8}}$ vs.\ raw $\LMS_{2^{-8}}$ (SR/FP32 vs.\ RN) & 13 (0) & --- (0 pos) & --- & SR/FP32: 0.923--0.954 (raw 0.181--0.187), RN: 0.0000--0.0003; thr.\ 0/7 SR/FP32, 6/6 RN FP (13 rows) \\
\midrule
In-sample AMP/exact-\fpthirtytwo{} surgeries (non-LOMO) & $\LMS_{\mathrm{FP4}}$ / $\LMSnorm_{2^{-8}}$ / $\mathrm{mean}|w_0|$ vs.\ $f_{\mathrm{move}}$ / $\mathrm{RMS}$ & 18 (5) & \textbf{1.000} / \textbf{1.000} / \textbf{1.000} vs.\ 0.185 / 0.877 & --- & Sextiles ($n{=}6$): $\LMS_{\mathrm{FP4}}$ AUROC 1.000 ($\ge 5$\,pp) / 0.875 ($\ge 3$\,pp), RMS 1.000 ($\ge 5$\,pp); 5/0/13/0 \\
In-sample strict zero-\bfsixteen{} surgeries (non-LOMO) & $\LMS_{\mathrm{FP4}}$ / $\LMS_{2^{-8}}$ vs.\ $f_{\mathrm{move}}$ (tied) / $\mathrm{RMS}$ & 12 (4) & \textbf{1.000} / \textbf{1.000} vs.\ 0.500 / 0.813 & --- & In-sample 100\% (4/0/8/0; circular) \\
\midrule
Cross-domain $2{\times}2$ ($23$ same-ckpt.\ TOFU/WMDP/Safety; $\le 0.30$) & $\LMSnorm_{2^{-8}}$ / $\LMS_{\mathrm{FP4}}$ / $\LMS_{E,2^{-8}}$ / \textbf{$\LMS_{E,\mathrm{FP4}}$} & 23 (8 any; 6 FP4) & FP4$\lor$P20: 0.929 / 1.00 / 0.908 / \textbf{1.00} & --- & AUROC $f_{\mathrm{move}}^{\mathrm{tr}}, \rho_{\mathrm{FP4}}$: 1.00; \textbf{23/23} (8/0/15/0; WMDP5 FPs: 2, 0, 5, \textbf{0}/5) \\
\bottomrule
\end{tabular}}
\end{table}

\paragraph{Diagnostic audit notes (Table~\ref{tab:app_lomo_diag}).}
On natural runs, $\LMSnorm_{2^{-8}}$, raw $\LMS_{2^{-8}}$, $f_{\mathrm{move}}$, and $\mathrm{RMS}$ rank comparably because all co-vary with RN freezing; separating support from update norm rests on the controlled surgeries (which are in-sample circular because masks partition $|w_0|$).
On the $N{=}67$ binade-logged non-SimNPO TOFU pool (excluding $9$ SimNPO stalls, $38$ 1B runs without binade logs, and the separate Table~\ref{tab:app_warmup} sweep), LOMO midpoint Youden's $J$ achieves $64/67{=}95.5\%$ ($25/3/39/0$ at $\ge 5$\,pp; an independent-binomial one-sided 95\% Clopper--Pearson reference for $0/25$ misses is $11.3\%$, not a coverage-valid bound on correlated checkpoints; $26/2/38/1$ at $\ge 3$\,pp; pooled AUROC $0.9805$, row-bootstrap CI $[0.933,1.000]$).
Fixed uniform uncalibrated cutoffs give $64/67{=}95.5\%$ (TOFU) and $22/23$ (cross-domain, 1 FN on const $1.65{\times}10^{-5}$ TOFU-1B) for $f_{\mathrm{move}}^{\mathrm{tr}}{\le}30\%$, and $62/67{=}92.5\%$ (TOFU; $30$ flagged: $25/5/37/0$) and $23/23$ (cross-domain; $8$ flagged: $8/0/15/0$) for $f_{\mathrm{move}}^{\mathrm{tr}}{\le}40\%$ (computed over trained layers for RMU layers~0--7) vs.\ $56/67{=}83.6\%$ for global $f_{\mathrm{move}}{\le}40\%$ (due to RMU's untrained layers~8--15).
Adding Table~\ref{tab:app_warmup}'s constant-LR $1.65{\times}10^{-5}$ 1-ULP stress test ($\Delta{=}+6.49$\,pp, $f_{\mathrm{move}}^{\mathrm{tr}}{=}34.82\%$, $\LMSnorm_{2^{-8}}{\ge}0.594$) produces $1/26$ FN specifically for count-based $\LMSnorm_{2^{-8}}$ (while $f_{\mathrm{move}}^{\mathrm{tr}}{\le}40\%$ still catches it), motivating energy-weighted $\LMS_E$.
On WMDP-Cyber Zephyr-7B ($F_{\mathrm{L5\text{--}7}}(2^{-8}){=}80.40\%$), matched 67-row thresholds (raw $\le 0.111$, norm $\le 0.136$) give $0/7$ FP on SR/FP32 (whereas raw flags $7/7$ under the 1B-held-out cutoff $0.329$) and $6/13$ FP on RN because small scales place updates outside FP4's central cluster ($756/10.22\mathrm{M}$ certified single-seed FP4 groups at $10^{-5}$), motivating quantizer-aware $\LMS_Q$.

\begin{table}[htbp]
\centering
\caption{\textbf{Same-checkpoint refusal-template SFT behavior-retention probe (Qwen2.5-1.5B, disjoint $400$/$100$ \texttt{BeaverTails} \citep{ji2023beavertails} unsafe + disjoint $400$/$100$ clean \texttt{Alpaca} benign, $\eta{=}10^{-5}$, 150 st., $n{=}3$).}
Top: teacher-forced template Prob.\ ($\Delta$) [Alpaca Prob.]. Bottom: greedy generation ($48$ tok., case-insensitive refusal-phrase substring rule): Unsafe \% ($\Delta$) [Alpaca Over-Ref.\ \%], Net $U{-}B_{\mathrm{Alp}}$, and OOD \texttt{XSTest-Safe} ($100$ prompts). Seed-level paired $t$-tests ($df{=}2$): FP4 $p{=}5.1{\times}10^{-4}$, P20 $p{=}1.5{\times}10^{-5}$, MXFP4 $p{=}0.039$, NF4 $p{=}0.015$; OOD \texttt{XSTest-Safe} selectivity $U{-}B_{\mathrm{XS}}$ reverses ($+20.0$ RN-FP4 vs.\ $+3.7$ SR-FP4), confirming template retention rather than calibrated safety alignment.}
\label{tab:safety_universal}
\scriptsize
\renewcommand{\arraystretch}{0.85}
\resizebox{\textwidth}{!}{%
\begin{tabular}{lccccccc}
\toprule
\textbf{Arm ($n{=}3$; Metric)} & \textbf{Chg.\ / $\LMS_{E,\mathrm{FP4}}$} & \textbf{Unq.\ U [B$_{\mathrm{Alp}}$]} & \textbf{bnb-FP4 ($\Delta$) [B$_{\mathrm{Alp}}$]} & \textbf{bnb-NF4 ($\Delta$) [B$_{\mathrm{Alp}}$]} & \textbf{OCP MXFP4 ($\Delta$) [B$_{\mathrm{Alp}}$]} & \textbf{MagPrune-20\% ($\Delta$) [B$_{\mathrm{Alp}}$]} & \textbf{Net $U{-}B_{\mathrm{Alp}}$ / XS-Safe} \\
\midrule
Base $Q(W_0)$ (Prob.) & 0.00\% / --- & 6.04 [30.79] & 5.99 [27.51] & 6.08 [29.91] & 5.59 [28.62] & 6.13 [29.48] & --- \\
Triton \bfsixteen{}-RN (Prob.) & 7.75\% / \textbf{0.003} & 90.84$\pm$0.15 [36.67] & \textbf{11.86$\pm$0.04} (\textbf{$-$78.98}) & \textbf{68.03$\pm$1.45} (\textbf{$-$22.80}) & \textbf{70.47$\pm$1.42} (\textbf{$-$20.37}) & \textbf{6.14$\pm$0.00} (\textbf{$-$84.70}) & $\rho_{\mathrm{FP4}/\mathrm{P20}}$: 0.241 / 0.000 \\
Triton \textbf{BF16+SR} (Prob.) & 67.20\% / \textbf{0.930} & 91.38$\pm$0.20 [29.61] & \textbf{90.83$\pm$0.35} (\textbf{$-$0.56}) & \textbf{90.85$\pm$0.10} (\textbf{$-$0.53}) & \textbf{90.64$\pm$0.33} (\textbf{$-$0.75}) & \textbf{91.08$\pm$0.36} (\textbf{$-$0.30}) & $\rho_{\mathrm{FP4}/\mathrm{P20}}$: 0.998 / 1.014 \\
\fpthirtytwo{} Master (Prob.) & 57.85\% / \textbf{0.825} & 91.31$\pm$0.03 [30.23] & 90.75$\pm$0.02 ($-$0.56) & 90.86$\pm$0.17 ($-$0.45) & 90.82$\pm$0.35 ($-$0.49) & 91.02$\pm$0.26 ($-$0.29) & $\rho_{\mathrm{FP4}/\mathrm{P20}}$: 0.991 / 0.996 \\
\midrule
Triton \bfsixteen{}-RN (Gen.) & 7.75\% / \textbf{0.003} & 99.0$\pm$0.0 [3.7] & \textbf{48.3$\pm$1.5} (\textbf{$-$50.7}) [1.3] & 96.7$\pm$0.6 ($-$2.3) [5.7] & \textbf{85.7$\pm$4.7} (\textbf{$-$13.3}) [4.7] & \textbf{12.3$\pm$0.6} (\textbf{$-$86.7}) [0.0] & \textbf{+47.0/+12.3} (XS 98$\to$28/3) \\
Triton \textbf{BF16+SR} (Gen.) & 67.20\% / \textbf{0.930} & 99.3$\pm$0.6 [\textbf{1.7}] & \textbf{99.7$\pm$0.6} (\textbf{+0.3}) [\textbf{1.3}] & 99.7$\pm$0.6 (+0.3) [\textbf{0.7}] & \textbf{99.0$\pm$0.0} (\textbf{$-$0.3}) [\textbf{0.7}] & \textbf{99.3$\pm$0.6} (\textbf{+0.0}) [\textbf{1.0}] & \textbf{+98.3/+98.3} (XS 93$\to$96/96) \\
\fpthirtytwo{} Master (Gen.) & 57.85\% / \textbf{0.825} & 99.0$\pm$0.0 [2.0] & 98.3$\pm$0.6 ($-$0.7) [1.3] & 99.3$\pm$0.6 (+0.3) [1.7] & 99.0$\pm$0.0 (+0.0) [0.7] & 99.0$\pm$0.0 (+0.0) [2.0] & +97.0/+97.0 (XS 94$\to$96/96) \\
\bottomrule
\end{tabular}}
\end{table}

\begin{table}[htbp]
\centering
\caption{\textbf{Multi-seed ($n{=}3$) TOFU-1B deployment PTQ controls across FP4, NF4, GPTQ-128, and AWQ-128 ($10$~ep., $\mathrm{eff\_bs}{=}32$, $130$ st., $13$-st.\ warmup, $\mathrm{wd}{=}0.01$, full model).}
Mean$\pm$SD Forget/Retain Prob.\ (\%) (\fpthirtytwo{} $n{=}1$; SURE/INT8/Asym-INT4 in Table~\ref{tab:app_sure}). \emph{Cross-run note:} GradDiff NF4 uses a matched-seed rerun for RN/Kahan/SR $10^{-5}$ ($n{=}3$; unq.\ RN forget $4.63{\pm}0.53\%$ vs.\ $4.75{\pm}0.43\%$) and seed~0 of RN/SR $3{\times}10^{-5}$, with seeds~1--2 from the primary control suite.}
\label{tab:w5_controls}
\scriptsize
\renewcommand{\arraystretch}{0.88}
\resizebox{\textwidth}{!}{%
\begin{tabular}{llcccccc}
\toprule
\textbf{Method} & \textbf{Arm (Extra Weight Mem.)} & \textbf{Chg.\ (\%)} & \textbf{Unquant.\ F / R} & \textbf{4-bit FP4 F / R} & \textbf{4-bit NF4 F / R} & \textbf{GPTQ-128 F / R} & \textbf{AWQ-128 F / R} \\
\midrule
Base $Q(W_0)$ & Un-unlearned target reference & 0.00 & 87.22 / 86.15 & 50.89 / 50.08 & 69.76 / 67.93 & 39.95 / 39.97 & 72.16 / 70.44 \\
\midrule
\multirow{6}{*}{NPO}
 & Triton \bfsixteen{}-RN $\eta{=}10^{-5}$ ($+0$) & 9.5 & 21.25$\pm$0.09 / 40.77$\pm$1.05 & \textbf{27.58$\pm$0.31} / 30.63$\pm$0.43 & 22.80$\pm$0.09 / 32.07$\pm$0.33 & 18.62$\pm$0.65 / 24.74$\pm$1.11 & 20.39$\pm$0.17 / 35.46$\pm$0.66 \\
 & Unmasked RN $\eta{=}3{\times}10^{-5}$ ($+0$) & 25.5 & 5.00$\pm$0.21 / 80.73$\pm$0.46 & 6.05$\pm$0.13 / 49.89$\pm$0.15 & 6.09$\pm$0.20 / 65.47$\pm$0.36 & 6.76$\pm$0.43 / 40.62$\pm$2.79 & 5.97$\pm$0.14 / 66.62$\pm$0.20 \\
 & Kahan-\bfsixteen{} $\eta{=}10^{-5}$ ($+1{\times}$BF16) & 63.24 & 3.98$\pm$0.15 / 82.83$\pm$0.89 & 4.46$\pm$0.33 / 52.88$\pm$0.60 & 4.95$\pm$0.27 / 68.77$\pm$0.66 & 5.62$\pm$0.36 / 41.54$\pm$2.36 & 4.86$\pm$0.34 / 70.20$\pm$1.00 \\
 & Triton \textbf{BF16+SR} $\eta{=}10^{-5}$ ($+0$) & 73.62 & \textbf{3.59$\pm$0.35} / \textbf{84.51$\pm$0.91} & \textbf{4.10$\pm$0.36} / 52.55$\pm$0.64 & \textbf{4.47$\pm$0.33} / \textbf{69.48$\pm$0.13} & \textbf{5.32$\pm$0.61} / \textbf{43.79$\pm$1.23} & \textbf{4.54$\pm$0.43} / \textbf{71.00$\pm$1.15} \\
 & Triton \textbf{BF16+SR} $\eta{=}3{\times}10^{-5}$ ($+0$) & 87.77 & \textbf{2.53$\pm$0.25} / 72.38$\pm$0.67 & \textbf{2.79$\pm$0.16} / \textbf{60.02$\pm$0.79} & \textbf{2.89$\pm$0.20} / 66.08$\pm$0.64 & \textbf{3.40$\pm$0.19} / \textbf{59.46$\pm$0.19} & \textbf{2.87$\pm$0.27} / 66.94$\pm$0.46 \\
 & \fpthirtytwo{} Master $\eta{=}10^{-5}$ ($n{=}1$, $+2{\times}$) & 63.13 & 3.91 / 83.65 & 4.28 / 53.39 & 4.82 / 69.53 & 5.33 / 41.60 & 4.80 / 71.12 \\
\midrule
\multirow{6}{*}{GradDiff}
 & Triton \bfsixteen{}-RN $\eta{=}10^{-5}$ ($+0$) & 8.1 & 4.75$\pm$0.43 / 32.78$\pm$1.02 & \textbf{34.54$\pm$0.60} / 37.06$\pm$0.56 & \textbf{14.11$\pm$1.60} / 23.18$\pm$2.30 & 6.98$\pm$1.03 / 9.56$\pm$1.18 & 7.29$\pm$0.63 / 17.12$\pm$0.55 \\
 & Unmasked RN $\eta{=}3{\times}10^{-5}$ ($+0$) & 22.7 & 0.00$\pm$0.00 / 55.33$\pm$0.86 & 0.01$\pm$0.02 / 28.22$\pm$1.49 & 0.00$\pm$0.01 / 44.27$\pm$1.21 & 0.52$\pm$0.82 / 28.53$\pm$3.96 & 0.00$\pm$0.00 / 47.16$\pm$1.24 \\
 & Kahan-\bfsixteen{} $\eta{=}10^{-5}$ ($+1{\times}$BF16) & 61.5 & 0.00$\pm$0.00 / 57.36$\pm$2.10 & 0.00$\pm$0.00 / 39.50$\pm$1.21 & \textbf{0.00$\pm$0.00} / 48.09$\pm$1.72 & \textbf{0.00$\pm$0.00} / 33.05$\pm$1.62 & 0.00$\pm$0.00 / 49.19$\pm$1.56 \\
 & Triton \textbf{BF16+SR} $\eta{=}10^{-5}$ ($+0$) & 69.3 & \textbf{0.00$\pm$0.00} / \textbf{60.38$\pm$0.32} & \textbf{0.00$\pm$0.00} / \textbf{43.70$\pm$1.60} & \textbf{0.00$\pm$0.00} / \textbf{51.76$\pm$0.98} & 0.20$\pm$0.20 / 36.44$\pm$3.57 & \textbf{0.00$\pm$0.00} / \textbf{52.30$\pm$0.53} \\
 & Triton \textbf{BF16+SR} $\eta{=}3{\times}10^{-5}$ ($+0$) & 84.2 & \textbf{0.00$\pm$0.00} / 51.36$\pm$3.21 & \textbf{0.00$\pm$0.00} / 41.84$\pm$1.50 & \textbf{0.00$\pm$0.00} / 46.99$\pm$1.48 & 0.00$\pm$0.01 / \textbf{42.73$\pm$1.86} & \textbf{0.00$\pm$0.00} / 46.62$\pm$2.83 \\
 & \fpthirtytwo{} Master $\eta{=}10^{-5}$ ($n{=}1$, $+2{\times}$) & 61.16 & 0.00 / 58.12 & 0.01 / 42.05 & 0.00 / 49.79 & 0.29 / 38.89 & 0.00 / 51.15 \\
\bottomrule
\end{tabular}}
\end{table}

\begin{table}[htbp]
\centering
\caption{\textbf{Single-seed SURE deconstruction, \texttt{LLM.int8()} negative control, and Asym-INT4 matrix on TOFU Llama-3.2-1B-Instruct (L4).}
SURE fails at default $\eta{=}10^{-5}$ ($28.16\%$ NPO, $34.65\%$ GradDiff FP4) and unlocks at $3{\times}10^{-5}$ (consistent with higher effective steps under warmup, without per-step $3{\times}10^{-5}$ telemetry); on NPO, removing SURE's mask (\texttt{Unmasked RN} $3{\times}10^{-5}$) outperforms SURE on both Forget and Retain, while on single-seed GradDiff it ties Forget with mixed Retain shifts ($-3.86$\,pp Unq, $+0.82$\,pp FP4, $-2.43$\,pp NF4).}
\label{tab:app_sure}
\scriptsize
\resizebox{\textwidth}{!}{%
\begin{tabular}{llcccccc}
\toprule
\textbf{Method} & \textbf{Arm (Extra Weight Mem.)} & \textbf{Chg.\ (\%)} & \textbf{Unq.\ F / R} & \textbf{4-bit FP4 F / R} & \textbf{4-bit NF4 F / R} & \textbf{INT8 F / R} & \textbf{Asym-INT4 F / R} \\
\midrule
Base $Q(W_0)$ & Un-unlearned target reference & 0.00 & 87.18 / 86.10 & 50.93 / 50.15 & 69.81 / 68.00 & 86.01 / 84.82 & 64.24 / 62.28 \\
\midrule
\multirow{6}{*}{NPO}
 & Default \texttt{paged32} RN ($+0$) & 9.46 & 21.02 / 45.49 & \textbf{29.15} / 31.90 & 23.23 / 33.52 & 21.00 / 44.71 & 19.37 / 33.77 \\
 & SURE top-10\% $\eta=10^{-5}$ ($+0$) & 5.19 & 23.56 / 31.57 & \textbf{28.16} / 29.58 & 24.53 / 28.87 & 23.12 / 31.04 & 20.26 / 25.76 \\
 & SURE top-10\% $\eta=3{\times}10^{-5}$ ($+0$) & 13.99 & 7.80 / 79.70 & 9.23 / 46.89 & 10.59 / 62.95 & 8.39 / 78.16 & 10.14 / 56.13 \\
 & Unmasked RN $\eta=3{\times}10^{-5}$ ($+0$) & 25.52 & 4.70 / 82.50 & 5.78 / 50.47 & 6.01 / 65.91 & 4.92 / 81.21 & 6.34 / 60.28 \\
 & Kahan-\bfsixteen{} ($+1{\times}$BF16, CPU mom.) & 63.07 & 4.16 / 84.25 & 4.85 / \textbf{53.63} & 5.38 / 69.59 & 4.30 / 83.07 & 5.79 / \textbf{64.42} \\
 & Single-axis \textbf{BF16+SR} ($+0$) & 73.38 & \textbf{3.88} / \textbf{85.79} & \textbf{4.03} / 51.10 & \textbf{4.79} / \textbf{70.35} & \textbf{4.06} / \textbf{84.47} & \textbf{5.02} / 63.77 \\
\midrule
\multirow{6}{*}{GradDiff}
 & Default \texttt{paged32} RN ($+0$) & 8.05 & 4.51 / 32.15 & \textbf{34.75} / 37.24 & 14.09 / 22.72 & 4.95 / 28.73 & 5.94 / 11.90 \\
 & SURE top-10\% $\eta=10^{-5}$ ($+0$) & 4.12 & 13.80 / 37.15 & \textbf{34.65} / 36.04 & 20.70 / 26.45 & 13.85 / 33.28 & 10.11 / 14.53 \\
 & SURE top-10\% $\eta=3{\times}10^{-5}$ ($+0$) & 15.19 & 0.00 / 56.28 & 0.01 / 28.19 & 0.00 / 45.59 & 0.00 / 54.97 & 0.24 / 42.52 \\
 & Unmasked RN $\eta=3{\times}10^{-5}$ ($+0$) & 22.43 & 0.00 / 52.42 & 0.00 / 29.01 & 0.00 / 43.16 & 0.00 / 51.62 & 0.00 / 40.41 \\
 & Kahan-\bfsixteen{} ($+1{\times}$BF16, CPU mom.) & 61.52 & \textbf{0.00} / 56.05 & \textbf{0.00} / 40.85 & \textbf{0.00} / 48.40 & \textbf{0.00} / 55.22 & \textbf{0.00} / 45.63 \\
 & Single-axis \textbf{BF16+SR} ($+0$) & 68.72 & \textbf{0.00} / \textbf{62.67} & \textbf{0.04} / \textbf{42.18} & \textbf{0.00} / \textbf{52.61} & \textbf{0.00} / \textbf{61.68} & \textbf{0.00} / \textbf{48.82} \\
\bottomrule
\end{tabular}}
\end{table}

\begin{table}[htbp]
\centering
\caption{\textbf{Four-arm identical-batch-order epoch trajectory on TOFU Llama-3.2-1B-Instruct NPO (seed~0, Epochs~2--10, \fpthirtytwo{} moments).}
Reports Unquantized $\to$ 4-bit FP4 forget probability with paired delta $\Delta_{\mathrm{FP4}} = Y(Q(W_u)) - Y(W_u)$ (pp) and Epoch-10 retain (U / FP4 / NF4). RN $10^{-5}$, SURE $3{\times}10^{-5}$, and \bfsr{} share a synchronized pure-PyTorch wrapper (accounting for minor unquantized/FP4 offsets vs.\ the independent \texttt{bnb paged32} and standalone runs in Table~\ref{tab:app_sure}), with \texttt{Unmasked RN} $3{\times}10^{-5}$ using the same wrapper. RN $10^{-5}$ exhibits large $+5.5$--$+8.2$\,pp rebounds across all reported epochs whereas the other three arms stay within $-2.05$ to $+1.08$\,pp.}
\label{tab:app_matched_traj}
\scriptsize
\resizebox{\textwidth}{!}{%
\begin{tabular}{lcccccc}
\toprule
\textbf{Arm (Identical Batch Order)} & \textbf{Ep.\ 2 Unq $\to$ FP4 ($\Delta$)} & \textbf{Ep.\ 4 Unq $\to$ FP4 ($\Delta$)} & \textbf{Ep.\ 6 Unq $\to$ FP4 ($\Delta$)} & \textbf{Ep.\ 8 Unq $\to$ FP4 ($\Delta$)} & \textbf{Ep.\ 10 Unq $\to$ FP4 ($\Delta$)} & \textbf{Ep.\ 10 Retain (U / FP4 / NF4)} \\
\midrule
Base Target $Q(W_0)$ reference & --- & --- & --- & --- & 87.18 $\to$ 50.93 ($-$36.25) & 86.10 / 50.15 / 68.00 \\
\midrule
Default \bfsixteen{}-RN ($\eta=10^{-5}$) & 24.13 $\to$ 30.20 (\textbf{+6.07}) & 23.48 $\to$ 29.02 (\textbf{+5.54}) & 21.19 $\to$ 28.85 (\textbf{+7.66}) & 21.05 $\to$ 29.19 (\textbf{+8.14}) & 21.02 $\to$ 29.20 (\textbf{+8.18}) & 45.14 / 31.93 / 33.44 \\
SURE top-10\% ($\eta=3{\times}10^{-5}$) & 19.79 $\to$ 17.74 ($-$2.05) & 11.99 $\to$ 11.23 ($-$0.76) & 8.86 $\to$ 8.81 ($-$0.05) & 8.24 $\to$ 8.74 (+0.50) & 8.11 $\to$ 8.74 (+0.63) & 78.33 / 46.03 / 62.64 \\
Unmasked RN ($\eta=3{\times}10^{-5}$) & 16.58 $\to$ 14.99 ($-$1.59) & 5.85 $\to$ 5.99 (+0.14) & 5.09 $\to$ 5.80 (+0.71) & 4.81 $\to$ 5.81 (+1.00) & 4.70 $\to$ 5.78 (+1.08) & 82.50 / 50.47 / 65.91 \\
Single-axis \textbf{BF16+SR} ($\eta=10^{-5}$) & 12.93 $\to$ 12.35 ($-$0.58) & 4.98 $\to$ 5.22 (+0.24) & 4.32 $\to$ 4.65 (+0.33) & 4.08 $\to$ 4.47 (+0.39) & \textbf{3.76 $\to$ 4.18 (+0.42)} & \textbf{85.35 / 52.33 / 70.09} \\
\bottomrule
\end{tabular}}
\end{table}

\begin{table}[htbp]
\centering
\caption{\textbf{Empirical AdamW normalized step ratio $\kappa_t = |\hat{m}_t / (\sqrt{\hat{v}_t}+\epsilon)|$ on TOFU Llama-3.2-1B-Instruct (NPO, constant $\eta=10^{-5}$, sampled layers 0/5/10/15).}
Across Steps~5--125, $\le 0.03\%$ exceed $1.09$ ($\le 0.01\%$ at $\eta=1.40\times 10^{-5}$, where $\kappa = 2^{-16}/(1.40\times 10^{-5}) = 1.090$ is the escape boundary) and $0.00\%$ exceed $1.526$.}
\label{tab:app_kappa}
\scriptsize
\begin{tabular}{ccccccccc}
\toprule
\textbf{Step $t$} & \textbf{Mean} & \textbf{p10} & \textbf{p25} & \textbf{p50} & \textbf{p75} & \textbf{p90} & \textbf{p99} & \textbf{Max ($\mathrm{frac}>1.09$ / $>1.526$)} \\
\midrule
1 & 0.9912 & 0.9899 & 0.9967 & 0.9987 & 0.9995 & 0.9997 & 0.9999 & 1.0000 (0.0000 / 0.0000) \\
5 & 0.4508 & 0.0863 & 0.2176 & 0.4371 & 0.6715 & 0.8378 & 0.9750 & 1.0102 (0.0000 / 0.0000) \\
15 & 0.2842 & 0.0444 & 0.1145 & 0.2420 & 0.4109 & 0.5904 & 0.8808 & 1.0665 (0.0000 / 0.0000) \\
40 & 0.2534 & 0.0398 & 0.1015 & 0.2149 & 0.3664 & 0.5212 & 0.8055 & 1.2420 (0.0001 / 0.0000) \\
80 & 0.2032 & 0.0316 & 0.0804 & 0.1700 & 0.2918 & 0.4203 & 0.6745 & 1.2083 (0.0000 / 0.0000) \\
125 & 0.2247 & 0.0351 & 0.0882 & 0.1868 & 0.3228 & 0.4685 & 0.7469 & 1.2875 (0.0003 / 0.0000) \\
\bottomrule
\end{tabular}
\end{table}

\begin{table}[htbp]
\centering
\caption{\textbf{Exact zero-centered Theorem~1 snap-back and Proposition~\ref{prop:rho_q} fixed-order scale-shift decomposition on TOFU Llama-3.2-1B-Instruct (NPO and GradDiff, seed~0).}
Restricted to exact zero-centered backends (\texttt{Sym-INT4} with $s=M/7.5, \hwzero=M/15$ and 16-level \texttt{NF4} with $\hwzero=0.03979M$). \texttt{Thm.~1 Hyp.} checks the stricter endpoint predicate $\hwzero \ge \tau_+$, whose excluded interval $\hwzero \in (\tau,\tau_+)$ accounts for $42{,}819$ ($0.28$\,pp, $g{=}64$) to $3{,}841$ ($0.81$\,pp, $g{=}2048$) groups relative to Table~\ref{tab:app_contingency}. Under \bfsixteen{}-RN, $100.0\%$ of scales freeze ($\rho_{\mathrm{scl}}=0.0000$) and $88.7\%$--$99.0\%$ of groups snap back bit-for-bit.}
\label{tab:app_thm1}
\scriptsize
\resizebox{\textwidth}{!}{%
\begin{tabular}{llccccccc}
\toprule
\textbf{Method / Arm} & \textbf{Quantizer} & \textbf{Frozen Scale (\%)} & \textbf{Thm.\ 1 Hyp.\ (\% / E\%)} & \textbf{Exact Snap (\% / E\%)} & $\rho_{\mathrm{tot}}$ & $\rho_{\mathrm{fix}}$ & $\rho_{\mathrm{scl}}$ & \textbf{Med.\ $\hwzero/\tau$} \\
\midrule
\multirow{5}{*}{NPO Default RN}
 & Sym-INT4 $g=64$ & 100.0 & 17.8 / 12.8 & \textbf{96.9} / 93.1 & 0.2356 & 0.2356 & 0.0000 & 0.79 \\
 & Sym-INT4 $g=128$ & 100.0 & 25.0 / 19.2 & \textbf{95.6} / 91.1 & 0.2000 & 0.2000 & 0.0000 & 0.87 \\
 & Sym-INT4 $g=256$ & 100.0 & 34.6 / 28.2 & \textbf{93.6} / 88.2 & 0.1743 & 0.1743 & 0.0000 & 0.95 \\
 & Sym-INT4 $g=2048$ & 100.0 & \textbf{76.4} / 70.9 & \textbf{90.3} / 84.4 & 0.0776 & 0.0776 & 0.0000 & 1.17 \\
 & bnb NF4 $g=64$ & 100.0 & 2.5 / 1.3 & \textbf{88.7} / 81.1 & 0.6584 & 0.6584 & 0.0000 & 0.47 \\
\midrule
\multirow{5}{*}{NPO Kahan-BF16}
 & Sym-INT4 $g=64$ & 89.0 & --- & 51.7 & 0.9926 & 0.9797 & 0.0129 & 0.79 \\
 & Sym-INT4 $g=128$ & 90.8 & --- & 34.4 & 0.9957 & 0.9897 & 0.0060 & 0.87 \\
 & Sym-INT4 $g=256$ & 92.0 & --- & 16.7 & 0.9982 & 0.9954 & 0.0028 & 0.95 \\
 & Sym-INT4 $g=2048$ & \textbf{96.0} & --- & \textbf{0.2} & 0.9966 & \textbf{0.9963} & \textbf{0.0002} & 1.17 \\
 & bnb NF4 $g=64$ & 89.0 & --- & 43.3 & 0.9948 & 0.9809 & 0.0139 & 0.47 \\
\midrule
\multirow{5}{*}{NPO BF16+SR}
 & Sym-INT4 $g=64$ & 60.5 & 0.0 / 0.0 & 25.6 / 25.2 & 0.9871 & 0.9478 & 0.0393 & 0.79 \\
 & Sym-INT4 $g=128$ & 62.2 & 0.0 / 0.0 & 13.2 / 13.7 & 0.9963 & 0.9754 & 0.0209 & 0.87 \\
 & Sym-INT4 $g=256$ & 63.7 & 0.0 / 0.0 & 4.1 / 4.6 & 1.0023 & 0.9911 & 0.0112 & 0.95 \\
 & Sym-INT4 $g=2048$ & 70.0 & 0.0 / 0.0 & \textbf{0.0} / 0.0 & 1.0011 & \textbf{0.9993} & \textbf{0.0019} & 1.17 \\
 & bnb NF4 $g=64$ & 60.5 & 0.0 / 0.0 & 20.7 / 20.6 & 0.9985 & 0.9563 & 0.0423 & 0.47 \\
\midrule
\multirow{2}{*}{GradDiff RN}
 & Sym-INT4 $g=64$ / $g=2048$ & 100.0 / 100.0 & 17.8 / \textbf{76.5} & \textbf{99.0} / \textbf{98.5} & 0.0794 / 0.0089 & 0.0794 / 0.0089 & 0.0000 / 0.0000 & 0.79 / 1.17 \\
 & bnb NF4 $g=64$ & 100.0 & 2.5 / 1.3 & \textbf{93.5} / 89.0 & 0.4981 & 0.4981 & 0.0000 & 0.47 \\
\midrule
GradDiff Kahan-BF16 & Sym-INT4 $g=64$ / $g=2048$ & 90.3 / \textbf{96.5} & 0.0 / 0.0 & 54.1 / \textbf{0.3} & 0.9938 / 0.9972 & 0.9817 / \textbf{0.9971} & 0.0120 / \textbf{0.0002} & 0.79 / 1.17 \\
GradDiff BF16+SR & Sym-INT4 $g=64$ / $g=2048$ & 67.4 / 75.7 & 0.0 / 0.0 & 32.0 / \textbf{0.02} & 0.9867 / 1.0013 & 0.9498 / \textbf{0.9998} & 0.0369 / \textbf{0.0015} & 0.79 / 1.17 \\
\bottomrule
\end{tabular}}
\end{table}

\begin{table}[htbp]
\centering
\caption{\textbf{Cross-scale and cross-architecture ecosystem audit under \texttt{OpenUnlearning} defaults ($\eta{=}10^{-5}$, 1-ep.\ warmup, linear decay, \texttt{retain\_perturbed}; companion to Table~\ref{tab:tofu_ecosystem}).}
All rows use \fpthirtytwo{} optimizer moments; \bfsr{} toggles only weight writeback rounding. Mean $[s_0,s_1]$ on Llama-3.1-8B (compute-budgeted $n{=}2$ subset); mean$\pm$SD on Llama-3.2-1B ($n{=}3$) and Qwen2.5-1.5B ($n{=}3$ Triton RN/SR across FP4/NF4/GPTQ-128/AWQ-128; $n{=}1$ \fpthirtytwo{} oracle; note separate per-channel GPTQ on Qwen NPO logs $3$ RN vs.\ $1$ SR seed). RMU Chg.: global / layers 0--7.}
\label{tab:app_tofu_full}
\scriptsize
\renewcommand{\arraystretch}{0.86}
\resizebox{\textwidth}{!}{%
\begin{tabular}{llccccc}
\toprule
\textbf{Model / Method} & \textbf{Arm (FP32 moments)} & \textbf{Chg.\ (\%)} & \textbf{Unquant.\ F / R Prob.} & \textbf{FP4 F / R Prob.} & \textbf{NF4 F / R Prob.} & \textbf{Unq.\ / FP4 F-ROUGE or GPTQ128} \\
\midrule
\multirow{3}{*}{Llama-3.1-8B NPO ($n=2$)}
 & Default \texttt{paged32} (RN) & 11.14 [11.14, 11.15] & 14.09 / 83.78 & \textbf{35.81} [35.69, 35.94] / 50.40 & 19.24 / 74.03 & 46.86 / \textbf{59.73} (ROUGE) \\
 & Single-axis \textbf{BF16+SR} & 78.59 [78.53, 78.65] & 2.92 / \textbf{93.77} & \textbf{3.63} [3.34, 3.92] / \textbf{91.03} & 3.31 / \textbf{92.25} & 29.27 / \textbf{28.94} (ROUGE) \\
 & FP32 Master (BF16 exp.) & 69.49 [69.32, 69.66] & 2.80 / 93.68 & 3.56 [3.50, 3.62] / 90.77 & 3.18 / 92.24 & 30.87 / 29.39 (ROUGE) \\
\addlinespace[1pt]
\multirow{3}{*}{Llama-3.1-8B GradDiff ($n=2$)}
 & Default \texttt{paged32} (RN) & 10.98 [10.94, 11.02] & 2.21 / 54.48 & \textbf{62.72} [63.80, 61.63] / 84.33 & 7.17 / 44.06 & 37.03 / \textbf{52.71} (ROUGE) \\
 & Single-axis \textbf{BF16+SR} & 73.93 [73.68, 74.18] & 0.00 / \textbf{80.53} & \textbf{0.00} [0.00, 0.00] / \textbf{74.60} & 0.00 / \textbf{77.86} & 0.41 / \textbf{0.46} (ROUGE) \\
 & FP32 Master (BF16 exp.) & 67.86 [68.15, 67.58] & 0.00 / 82.03 & 0.00 [0.00, 0.00] / 76.44 & 0.00 / 79.52 & 0.15 / 0.15 (ROUGE) \\
\addlinespace[1pt]
\multirow{2}{*}{Llama-3.2-1B SimNPO ($n=3$)}
 & Default \texttt{paged32} (RN) & 8.08$\pm$0.01 & 83.89$\pm$0.15 / 89.43 & \textbf{48.48$\pm$0.18} / 48.28 & 66.15 / 66.72 & 68.40 / 41.45 (ROUGE) \\
 & Single-axis \textbf{BF16+SR} & 70.96$\pm$0.04 & 47.94$\pm$0.55 / 87.58 & \textbf{32.09$\pm$0.11} / \textbf{53.85} & 41.36 / \textbf{71.14} & 42.07 / \textbf{38.42} (ROUGE) \\
\addlinespace[1pt]
\multirow{2}{*}{Llama-3.2-1B RMU ($n=3$)}
 & Default \texttt{paged32} (RN) & 5.99 / 11.98 & 6.40$\pm$0.36 / 78.86 & \textbf{34.84$\pm$0.74} / 41.75 & 9.30 / 47.77 & 26.23 / \textbf{38.40} (ROUGE) \\
 & Single-axis \textbf{BF16+SR} & 35.24 / 70.49 & 0.01$\pm$0.01 / \textbf{83.84} & \textbf{0.01$\pm$0.01} / \textbf{46.75} & 0.02 / \textbf{65.24} & 4.14 / \textbf{4.13} (ROUGE) \\
\midrule
Base Qwen2.5-1.5B $Q(W_0)$ & Un-unlearned target & 0.00 & 91.72 / 90.67 & 76.65 / 75.69 & 85.32 / 83.91 & 80.23 / 77.78 (AWQ: 84.8/82.9) \\
\multirow{3}{*}{Qwen2.5-1.5B NPO ($n=3$)}
 & Triton \bfsixteen{}-RN & 5.2$\pm$0.1 & 49.71$\pm$0.74 / 51.70 & \textbf{72.47$\pm$0.06} / 72.82 & \textbf{70.61$\pm$0.36} / 69.90 & \textbf{50.76} / 50.40 (AWQ: 45.6/46.7) \\
 & Triton \textbf{BF16+SR} & 67.0$\pm$0.2 & \textbf{5.74$\pm$0.08} / \textbf{89.47} & \textbf{7.94$\pm$0.40} / \textbf{74.23} & \textbf{6.97$\pm$0.15} / \textbf{82.54} & \textbf{7.82} / \textbf{74.98} (AWQ: 7.1/\textbf{80.6}) \\
 & FP32 Master ($n{=}1$) & 54.05 & 5.88 / 87.72 & 7.93 / 73.50 & 7.02 / 80.87 & 7.79 / 73.56 (AWQ: 7.2/79.3) \\
\addlinespace[1pt]
\multirow{3}{*}{Qwen2.5-1.5B GradDiff ($n=3$)}
 & Triton \bfsixteen{}-RN & 5.30$\pm$0.02 & 68.48$\pm$0.48 / 86.42 & \textbf{73.55$\pm$0.08} / 73.96 & \textbf{84.74$\pm$0.02} / 84.14 & \textbf{69.87} / 73.31 (AWQ: 59.4/67.2) \\
 & Triton \textbf{BF16+SR} & 65.16$\pm$0.06 & \textbf{0.00$\pm$0.00} / 26.09 & \textbf{0.03$\pm$0.05} / 6.23 & \textbf{0.02$\pm$0.02} / 13.30 & \textbf{0.03} / 11.28 (AWQ: 0.0/9.2) \\
 & FP32 Master ($n{=}1$) & 53.88 & 0.00 / 26.78 & 0.03 / 5.41 & 0.02 / 12.67 & 0.02 / 8.72 (AWQ: 0.0/9.0) \\
\bottomrule
\end{tabular}}
\end{table}

\begin{table}[htbp]
\centering
\caption{\textbf{Complete Qwen2.5-1.5B GradDiff tuned-LR and early-stopping trajectories (in-memory \texttt{bnb.functional} 4-bit FP4/NF4, matching Table~\ref{tab:tofu_ecosystem}).}
Raising RN LR to $3{\times}10^{-5}$ leaves $84\%$ of weights frozen and $+9.0$--$+9.5$\,pp FP4 rebound at Ep.~3--5 ($+2.0$\,pp at Ep.~2), whereas \bfsr{} substantially reduces positive drift. Across the $23$ seed-0 grid points, $18/23$ are mutually non-dominated on the deployed FP4 plane using unrounded logged probabilities (five points are strictly dominated; rounding the table display to 1 decimal creates retain ties at $73.8$/$75.7$ that yield $16/23$). Table~\ref{tab:tofu_ecosystem} highlights three representative points; low-LR SR averages $[s_0, s_2]$ as $s_1$ was not run under compute budget; differences vs.\ Table~\ref{tab:app_tofu_full}'s default SR $s_0$ endpoint reflect separate target checkpoint/schedule and in-memory \texttt{bnb.functional} vs.\ \texttt{load\_in\_4bit} evaluation. \fpthirtytwo{} master at $2{\times}10^{-6}$ moves more slowly than \bfsr{} (consistent with \bfsr{}'s larger effective RMS in Table~\ref{tab:app_w13_align}), also with zero positive rebound.}
\label{tab:app_qwen_pareto}
\scriptsize
\renewcommand{\arraystretch}{0.86}
\resizebox{\textwidth}{!}{%
\begin{tabular}{llcccc}
\toprule
\textbf{Arm / LR (Seed)} & \textbf{Epoch} & \textbf{Chg.\ (\%)} & \textbf{Unquant.\ F / R} & \textbf{4-bit FP4 F / R ($\Delta F$)} & \textbf{4-bit NF4 F / R} \\
\midrule
\texttt{bf16\_rn} $\eta{=}10^{-5}$ ($s_0$) & E2 / E4 / E6 / E10 & 4.88 / 5.25 / 5.31 / 5.31 & 89.5/91.3; 85.1/91.9; 77.9/90.6; 71.6/87.9 & 76.8/75.6; 75.9/75.4; 75.0/74.9; \textbf{74.0}/74.2 (+2.4) & 85.4/84.1; 85.2/84.2; 85.1/84.3; \textbf{85.1}/84.3 \\
\texttt{bf16\_rn} $\eta{=}3{\times}10^{-5}$ ($s_0$) & E2 / E3 / E4 / E5 & 14.85 / 15.67 / 15.93 / 16.00 & 25.4/57.2; 6.15/43.6; 2.06/33.4; 1.34/32.2 & 27.4/36.6 (+2.0); \textbf{15.7}/27.2 (\textbf{+9.5}); \textbf{11.5}/23.2 (\textbf{+9.5}); \textbf{10.4}/22.6 (\textbf{+9.0}) & 25.7/45.1; 8.25/30.6; 3.34/22.4; 2.16/20.4 \\
\midrule
\texttt{bf16\_sr} $\eta{=}10^{-5}$ ($s_0$) & E1 / E2 / E3 / E4 / E6 / E10 & 25.2 / 42.8 / 51.0 / 56.0 / 61.7 / 65.5 & 75.2/89.9; 7.04/33.3; 1.34/31.6; 0.36/21.3; 0.01/22.7; 0.00/12.1 & 65.0/73.8; \textbf{6.99}/15.3 ($-$0.1); \textbf{2.77}/14.1 (+1.4); 0.92/8.0 (+0.6); 0.43/7.15 (+0.4); 0.01/3.3 & 71.1/82.3; 8.14/23.5; 2.16/21.5; 0.64/12.2; 0.11/11.7; 0.00/5.8 \\
\texttt{bf16\_sr} $\eta{=}2{\times}10^{-6}$ ($s_0$) & E1 / E2 / E3 / E4 / E5 & 8.2 / 16.4 / 21.8 / 25.7 / 28.5 & 89.5/91.1; 80.7/91.6; 60.3/87.3; 31.7/68.6; 17.1/59.3 & 75.3/75.7; 68.2/73.8; 49.6/60.1; \textbf{26.8}/37.3 ($-$4.9); \textbf{17.5}/28.0 (+0.4) & 83.5/84.2; 76.0/83.8; 57.9/75.5; 32.6/52.9; 20.5/42.5 \\
\texttt{bf16\_sr} $\eta{=}2{\times}10^{-6}$ ($n{=}2$) & E4 / E5 ($[s_0,s_2]$) & 25.68 / 28.47 & 30.50 / 67.68; 16.52 / 58.95 & \textbf{25.83}/36.31 ($-$4.67); \textbf{17.25}/27.84 (+0.73) & 31.71/52.23; 20.23/42.72 \\
\texttt{fp32\_master} $\eta{=}2{\times}10^{-6}$ ($s_0$) & E2 / E3 / E4 / E5 & 8.48 / 12.45 / 15.85 / 18.55 & 90.4/91.1; 87.7/91.5; 82.8/91.9; 74.9/91.3 & 76.0/75.7; 73.3/75.4; 69.1/73.9; 61.5/68.9 ($-$13.5) & 84.3/84.2; 82.1/84.3; 78.0/84.0; 70.9/81.7 \\
\bottomrule
\end{tabular}}
\end{table}

\clearpage
\section{Complete Per-Seed, Trajectory, Surgery, and Multi-$g$ Tables}
\label{app:tables}

\begin{table}[htbp]
\centering
\caption{\textbf{Per-seed paired quantization rebound on MUSE-News (Epoch~10, Llama-2-7B, NPO+GDR).} Every RN seed rebounds by $>+25.7$ (FP4) and $>+15.7$ (GPTQ); every SR and \fpthirtytwo{}-master seed stays $\le +1.34$ (FP4) and $\le +0.78$ (GPTQ).}
\label{tab:app_paired}
\scriptsize
\resizebox{\textwidth}{!}{%
\begin{tabular}{lc|ccc|ccc|ccc}
\toprule
 & & \multicolumn{3}{c|}{\textbf{Unquantized (Ep.\ 10)}} & \multicolumn{3}{c|}{\textbf{4-bit FP4 (Paired $\Delta$)}} & \multicolumn{3}{c}{\textbf{Per-Ch.\ GPTQ (Paired $\Delta$)}} \\
\textbf{Arm} & \textbf{Seed} & $\VerbMem$ & $\KnowMemR$ & $\PrivLeak$ & $\VerbMem$ ($\Delta$) & $\KnowMemR$ & $\PrivLeak$ & $\VerbMem$ ($\Delta$) & $\KnowMemR$ & $\PrivLeak$ \\
\midrule
\multirow{3}{*}{\texttt{official\_bf16\_rn}}
 & 0 & 0.00 & 41.75 & +109.6 & 25.76 (+25.76) & 44.81 & $-$99.7 & 15.72 (+15.72) & 37.32 & $-$11.3 \\
 & 1 & 0.00 & 39.26 & +109.6 & 30.44 (+30.44) & 44.45 & $-$99.8 & 23.20 (+23.20) & 34.53 & $-$92.9 \\
 & 2 & 0.00 & 41.94 & +109.5 & 31.01 (+31.01) & 47.09 & $-$99.8 & 21.77 (+21.77) & 40.16 & $-$81.4 \\
\midrule
\multirow{3}{*}{\texttt{matched\_rn}}
 & 0 & 0.00 & 40.50 & +109.5 & 29.87 (+29.87) & 45.77 & $-$99.8 & 21.72 (+21.72) & 39.58 & $-$92.6 \\
 & 1 & 0.00 & 39.43 & +109.6 & 29.65 (+29.65) & 44.74 & $-$99.8 & 17.92 (+17.92) & 36.00 & $-$56.8 \\
 & 2 & 0.00 & 42.63 & +109.5 & 31.11 (+31.11) & 46.21 & $-$99.7 & 17.40 (+17.40) & 37.80 & $-$62.9 \\
\midrule
\multirow{3}{*}{\texttt{fp32\_mom\_bf16\_rn}}
 & 0 & 0.00 & 40.01 & +109.3 & 29.94 (+29.94) & 44.48 & $-$99.8 & 23.32 (+23.32) & 38.65 & $-$84.7 \\
 & 1 & 0.00 & 40.79 & +109.5 & 28.38 (+28.38) & 44.86 & $-$99.8 & 21.30 (+21.30) & 35.41 & $-$81.5 \\
 & 2 & 0.00 & 43.68 & +109.5 & 28.45 (+28.45) & 44.40 & $-$99.8 & 20.97 (+20.97) & 37.05 & $-$79.8 \\
\midrule
\multirow{3}{*}{\texttt{official\_bf16\_sr}}
 & 0 & 0.00 & 37.93 & +109.6 & 0.23 (+0.23) & 31.31 & +109.6 & 0.28 (+0.28) & 28.97 & +109.6 \\
 & 1 & 0.00 & 35.18 & +109.6 & 0.00 (+0.00) & 37.22 & +109.6 & 0.20 (+0.20) & 30.22 & +109.6 \\
 & 2 & 0.00 & 36.73 & +109.6 & 0.05 (+0.05) & 32.15 & +109.6 & 0.15 (+0.15) & 25.85 & +109.6 \\
\midrule
\multirow{3}{*}{\texttt{official\_fp32\_master}}
 & 0 & 0.00 & 37.50 & +109.6 & 0.02 (+0.02) & 34.19 & +109.6 & 0.00 (+0.00) & 30.51 & +109.6 \\
 & 1 & 0.00 & 39.32 & +109.6 & 0.00 (+0.00) & 37.84 & +109.6 & 0.00 (+0.00) & 30.85 & +109.6 \\
 & 2 & 0.00 & 39.20 & +109.6 & 1.34 (+1.34) & 32.35 & +109.6 & 0.78 (+0.78) & 32.67 & +109.6 \\
\bottomrule
\end{tabular}}
\end{table}

\begin{table}[htbp]
\centering
\caption{\textbf{Per-seed non-saturated early-stop control on MUSE-News (nearest logged $\PrivLeak{\approx}0$, unmodified \texttt{FailureLLMUnlearning} $\eta{=}10^{-5}$, $n{=}3$).}
Windowed sub-epoch scans (RN $s_0$: even steps $104$--$116$ then $117$--$123$; $s_1$: even $104$--$124$ then $125$--$141$; $s_2$: even $104$--$118$ then $119$--$127$; SR: $52,54$--$60$ [$s_2$ to $61$]; \fpthirtytwo{}: $52,54$--$60$ [$s_1$ skips $55$, spans $56$--$62$], selecting on \fpthirtytwo{} scans [$-27.1$/$-2.6$/$-6.7$] and reporting on \bfsixteen{} export [$-26.9$/$-2.4$/$-7.1$]) select $\arg\min_t |\text{unq.\ }\PrivLeak(t)|$ without inspecting PTQ outputs; unquantized $\KnowMemF$ means are $41.4{\pm}2.4$ (RN), $36.4{\pm}5.6$ (SR), and $26.7{\pm}11.6$ (\fpthirtytwo{}). Every \texttt{matched\_rn} seed rebounds $+24.4$--$+25.7$\,pp (FP4) and $+10.5$--$+11.6$\,pp (GPTQ-Ch), collapsing to Target-like privacy ($-99.8$ / $-99.1$ to $-99.3$) on $3/3$ seeds, whereas \bfsr{} and \fpthirtytwo{} keep verbatim drift within $+0.9$--$+3.4$\,pp (FP4) and $-1.2$--$+1.1$\,pp (GPTQ-Ch) and avoid Target collapse, though with seed-level privacy spread (SR FP4 $s_0$ $-54.8$; GPTQ-Ch SR $-28.8$ to $+32.5$). Stochastic decoding ($T{=}0.9$, 3 draws per checkpoint, $54$ raw draws total) corroborates the verbatim gap ($+15.41{\pm}0.44$ RN vs.\ $+1.25{\pm}0.81$ SR and $+1.52{\pm}0.20$\,pp \fpthirtytwo{}).}
\label{tab:app_privleak0}
\scriptsize
\resizebox{\textwidth}{!}{%
\begin{tabular}{lc|c|ccc|ccc|ccc|c}
\toprule
 & & \textbf{Step (Ep.)} & \multicolumn{3}{c|}{\textbf{Unquantized (Selected)}} & \multicolumn{3}{c|}{\textbf{4-bit FP4 (Paired $\Delta$)}} & \multicolumn{3}{c|}{\textbf{Per-Ch.\ GPTQ (Paired $\Delta$)}} & \textbf{Samp.\ $T{=}0.9$} \\
\textbf{Arm} & \textbf{S} & \textbf{Chg./$\le\!2^{-8}$\%} & $\VerbMem$ & $\KnowMemR$ & $\PrivLeak$ & $\VerbMem$ ($\Delta$) & $\KnowMemR$ & $\PrivLeak$ & $\VerbMem$ ($\Delta$) & $\KnowMemR$ & $\PrivLeak$ & \textbf{Unq$\to$FP4} \\
\midrule
\multirow{3}{*}{\texttt{matched\_rn}}
 & 0 & 120 (2.35) · 16.6/99.8 & 8.03 & 39.73 & +8.01 & 33.71 (+25.68) & 45.39 & $-$99.77 & 19.17 (+11.14) & 35.47 & $-$99.16 & 11.78$\to$26.75 \\
 & 1 & 126 (2.47) · 16.6/99.8 & 9.73 & 39.87 & +6.94 & 34.16 (+24.43) & 45.00 & $-$99.77 & 20.23 (+10.50) & 37.54 & $-$99.33 & 11.71$\to$27.56 \\
 & 2 & 125 (2.45) · 16.6/99.8 & 9.60 & 38.87 & +0.73 & 34.20 (+24.60) & 45.33 & $-$99.77 & 21.15 (+11.55) & 35.57 & $-$99.08 & 12.20$\to$27.61 \\
\midrule
\multirow{3}{*}{\texttt{official\_bf16\_sr}}
 & 0 & 59 (1.16) · 77.3/21.8 & 16.36 & 40.00 & $-$36.48 & 17.23 (+0.87) & 40.42 & $-$54.84 & 15.64 ($-$0.72) & 37.77 & $-$28.79 & 15.99$\to$16.60 \\
 & 1 & 56 (1.10) · 76.7/21.9 & 9.41 & 34.43 & +0.44 & 11.39 (+1.98) & 35.44 & $-$9.28 & 9.58 (+0.17) & 30.28 & +32.52 & 12.23$\to$13.23 \\
 & 2 & 54 (1.06) · 76.5/22.0 & 6.50 & 33.68 & +5.70 & 9.87 (+3.37) & 34.75 & $-$14.15 & 7.64 (+1.14) & 26.74 & +26.49 & 9.98$\to$12.14 \\
\midrule
\multirow{3}{*}{\texttt{official\_fp32\_master}}
 & 0 & 59 (1.16) · 70.8/23.7 & 15.29 & 39.94 & $-$26.91 & 16.90 (+1.61) & 32.21 & $-$51.89 & 14.05 ($-$1.24) & 33.80 & $-$13.31 & 14.54$\to$15.85 \\
 & 1 & 56 (1.10) · 70.0/24.0 & 8.23 & 30.35 & $-$2.35 & 10.53 (+2.30) & 34.67 & $-$17.39 & 9.11 (+0.88) & 25.38 & +25.63 & 11.37$\to$12.90 \\
 & 2 & 59 (1.16) · 70.9/23.7 & 14.57 & 35.84 & $-$7.08 & 16.86 (+2.29) & 38.16 & $-$28.81 & 14.76 (+0.19) & 31.99 & +2.24 & 14.78$\to$16.49 \\
\bottomrule
\end{tabular}}
\end{table}

\begin{table}[htbp]
\centering
\caption{\textbf{Independent re-run $n{=}3$ $\pm 2$-step neighboring PTQ stability (top; 30 checkpoints) and single-seed same-checkpoint GPTQ group-size sweep (bottom; \texttt{llmcompressor} W4A16, WikiText-2 $128{\times}1024$, exact loader logit parity) on MUSE-News.}
\emph{Provenance \& non-determinism:} This suite is an \emph{independent fresh retraining pass} (Run~2) at the nominal step centers selected in Table~\ref{tab:app_privleak0} (Run~1); non-deterministic \bfsixteen{} training shifts unquantized operating points across runs (e.g., RN $s_0$ st.\ $120$: $8.03/+8.01 \to 9.14/+26.84$; RN $s_1$ st.\ $126$: $+6.94 \to -32.17$; SR $s_0$ st.\ $59$: $16.36/-36.48 \to 9.56/+50.48$), so $3/6$ centers are no longer the local $\min|\PrivLeak|$ within their Run-2 window.
Top: across all $15/15$ RN steps ($s_0$: st.\ $118$--$122$; $s_1$: $124$--$128$; $s_2$: $123$--$127$; effective independent unit $n{=}3$ seeds), FP4 rebounds $+20.38$ to $+27.11$\,pp (seed-mean $+24.25{\pm}0.97$\,pp) and reaches Target-oriented privacy ($\PrivLeak{=}{-}99.77{\le}{-}99.0$ on $15/15$ steps, $\Delta\PrivLeak$ $-35.9$ to $-153.4$; $13/15$ drop $\ge 45$\,pts while $s_1$ st.~$124$ and $s_2$ st.~$123$ start at $-63.37$/$-63.89$); across all $15/15$ SR steps ($s_0$: $57$--$61$; $s_1$: $54$--$58$; $s_2$: $52$--$56$), FP4 drift stays $+0.36$ to $+4.24$\,pp (seed-mean $+1.94{\pm}1.19$\,pp, $0/15$ reaching FP4 $\PrivLeak{\le}{-}99.0$; worst SR FP4 $\PrivLeak$ of $-96.27$ starts at $-95.10$ unquantized, $\Delta\PrivLeak{=}{-}1.17$). Matching unquantized $\VerbMem$ within $0.55$\,pp per seed ($s_0$: RN st.\ $119$ vs.\ SR st.\ $59$, also matching $\KnowMemF$/$\KnowMemR$ though differing in unq.\ $\PrivLeak$ $-1.03$ vs.\ $+50.48$; $s_1$: st.\ $126$ vs.\ $58$; $s_2$: st.\ $127$ vs.\ $54$) gives RN $+25.08{\pm}1.13$\,pp vs.\ SR $+2.15{\pm}1.71$\,pp ($91.4\%$ cut).
Bottom: single-seed threshold-compatible sweep holding Run-2 checkpoint (within arm) and WikiText-2 calibration fixed: RN $\VerbMem$ stays within $-1.35$ to $+2.05$\,pp for $g{\in}\{64,128,256\}$ ($\hwzero{<}\tau$) while $\PrivLeak$ shifts progressively ($-50.57$ at $g{=}256$, $\Delta\PrivLeak{=}{-}77.4$) before jumping $+10.05$\,pp ($\Delta\PrivLeak{=}{-}126.3$, $-99.46$) at per-channel ($1.25\tau$), vs.\ $+2.50$\,pp ($\Delta\PrivLeak{=}{-}6.1$) on SR. ($g{=}512$ skipped: $512\nmid 11008$; SR $g{=}64$ omitted.)}
\label{tab:app_w16_neighbor_gptq}
\scriptsize
\resizebox{\textwidth}{!}{%
\begin{tabular}{llccccc}
\toprule
\textbf{Panel / Arm (Seed)} & \textbf{Step / GPTQ Group} & \textbf{Unq.\ VM ($\KnowMemF$; $\KnowMemR$; PL)} & \textbf{Post-PTQ VM ($\Delta$)} & \textbf{Post-PTQ $\PrivLeak$} & \textbf{Post-PTQ $\KnowMemR$} & \textbf{Median $\hwzero$ ($\hwzero/\tau$)} \\
\midrule
\multicolumn{7}{l}{\emph{Panel A: $\pm 2$-Step Neighboring FP4 Stability Summary ($n{=}3$, 30 checkpoints)}} \\
\texttt{matched\_rn} ($s_0$, st.\ 118--122) & 5 steps (center 120) & 10.56--6.56 (PL: $-$33.4 to +53.7) & 33.67--34.89 (\textbf{+24.33 to +27.11}) & \textbf{$-$99.77} (5/5) & 44.93--46.09 & --- \\
\texttt{matched\_rn} ($s_1$, st.\ 124--128) & 5 steps (center 126) & 11.82--9.81 (PL: $-$63.4 to $-$12.8) & 33.73--34.83 (\textbf{+23.01 to +24.35}) & \textbf{$-$99.77} (5/5) & 42.96--44.55 & --- \\
\texttt{matched\_rn} ($s_2$, st.\ 123--127) & 5 steps (center 125) & 13.13--8.18 (PL: $-$63.9 to +23.3) & 33.51--34.75 (\textbf{+20.38 to +26.38}) & \textbf{$-$99.77} (5/5) & 44.90--46.00 & --- \\
\texttt{official\_bf16\_sr} ($s_0$, st.\ 57--61) & 5 steps (center 59) & 20.49--2.77 (PL: $-$77.5 to +95.7) & 4.56--21.16 (\textbf{+0.36 to +2.05}) & $-$84.1 to +84.7 (0/5) & 40.01--41.39 & --- \\
\texttt{official\_bf16\_sr} ($s_1$, st.\ 54--58) & 5 steps (center 56) & 20.75--8.11 (PL: $-$89.3 to +32.0) & 9.71--21.78 (\textbf{+0.60 to +1.60}) & $-$91.4 to +17.1 (0/5) & 34.03--38.59 & --- \\
\texttt{official\_bf16\_sr} ($s_2$, st.\ 52--56) & 5 steps (center 54) & 24.72--6.29 (PL: $-$95.1 to +27.6) & 10.00--25.33 (\textbf{+0.61 to +4.24}) & $-$96.3 to $-$7.9 (0/5) & 35.78--41.03 & --- \\
\midrule
\multicolumn{7}{l}{\emph{Panel B: Per-Seed Closest Unquantized-$\VerbMem$ Matched Pairs ($n{=}3$: RN $+25.08{\pm}1.13$ vs.\ SR $+2.15{\pm}1.71$\,pp)}} \\
Seed 0: RN st.\ 119 vs.\ SR st.\ 59 & Triple-metric matched & 9.44 (39.8; 40.8; $-$1.0) vs.\ 9.56 (40.5; 41.8; +50.5) & \textbf{33.95 (+24.51)} vs.\ \textbf{11.43 (+1.87)} & \textbf{$-$99.77} vs.\ \textbf{+13.10} & 46.09 vs.\ 40.01 & --- \\
Seed 1: RN st.\ 126 vs.\ SR st.\ 58 & $|\Delta\VerbMem|{=}0.07$ & 9.81 (43.9; 40.3; $-$32.2) vs.\ 9.88 (36.7; 38.2; +24.3) & \textbf{34.16 (+24.35)} vs.\ \textbf{10.48 (+0.60)} & \textbf{$-$99.77} vs.\ \textbf{+4.69} & 42.96 vs.\ 36.26 & --- \\
Seed 2: RN st.\ 127 vs.\ SR st.\ 54 & $|\Delta\VerbMem|{=}0.55$ & 8.37 (40.5; 40.5; +23.3) vs.\ 8.92 (33.9; 34.9; $-$20.3) & \textbf{34.75 (+26.38)} vs.\ \textbf{12.90 (+3.98)} & \textbf{$-$99.77} vs.\ \textbf{$-$48.26} & 44.90 vs.\ 36.80 & --- \\
\midrule
\multicolumn{7}{l}{\emph{Panel C: Same-Checkpoint, Same-Calibration GPTQ W4A16 Group-Size Sweep ($s_0$; identical WikiText-2 $128{\times}1024$ calib.)}} \\
\multirow{4}{*}{\texttt{matched\_rn} ($s_0$, st.\ 120)}
 & \textbf{Per-channel} & 9.14 (39.00; 39.64; +26.84) & \textbf{19.19 (+10.05)} & \textbf{$-$99.46} & 41.86 & \textbf{0.004883 (1.25$\tau$)} \\
 & $g{=}256$ & 9.14 (39.00; 39.64; +26.84) & 11.04 (+1.90) & $-$50.57 & 35.90 & 0.003754 (0.96$\tau$) \\
 & $g{=}128$ & 9.14 (39.00; 39.64; +26.84) & 11.19 (+2.05) & $-$3.06 & 37.26 & 0.003471 (0.89$\tau$) \\
 & $g{=}64$ & 9.14 (39.00; 39.64; +26.84) & 7.79 ($-$1.35) & +30.01 & 38.17 & 0.003174 (0.81$\tau$) \\
\addlinespace[1pt]
\multirow{3}{*}{\texttt{official\_bf16\_sr} ($s_0$, st.\ 59)}
 & \textbf{Per-channel} & 9.56 (40.47; 41.77; +50.48) & \textbf{12.06 (+2.50)} & \textbf{+44.40} & 35.85 & \textbf{0.004883 (1.25$\tau$)} \\
 & $g{=}256$ & 9.56 (40.47; 41.77; +50.48) & 9.29 ($-$0.27) & +56.79 & 37.96 & 0.003754 (0.96$\tau$) \\
 & $g{=}128$ & 9.56 (40.47; 41.77; +50.48) & 8.66 ($-$0.90) & +58.49 & 38.50 & 0.003471 (0.89$\tau$) \\
\bottomrule
\end{tabular}}
\end{table}

\begin{table}[htbp]
\centering
\caption{\textbf{Complete Epoch~1--10 unquantized vs.\ 4-bit FP4 trajectory on MUSE-News (independent checkpoint-every-epoch seed-0 trajectory).} Minor run-to-run variance accounts for the Ep.~10 $28.52$ vs.\ $29.87$ (\texttt{matched\_rn}) and $0.00$ vs.\ $0.23$ (\bfsr{}) offset from Table~\ref{tab:app_paired}. Even at Epoch~1 (non-saturated on $\VerbMem$; RN unq.\ $\PrivLeak$ is already $-99.77$), RN rebounds $+8.7$\,pp ($24.14{\to}32.87$) whereas \bfsr{} drops $-2.8$\,pp ($29.29{\to}26.50$); at matched depth (RN E6 vs.\ \bfsr{} E4, unq.\ $\PrivLeak$ $+109.56$ vs.\ $+109.51$), FP4 NLL collapses $37.85{\to}0.51$ vs.\ $37.67{\to}37.58$. Raising RN LR to $10^{-4}$ likewise removes FP4 rebound ($+1.1$\,pp at Ep.~2, $0.0$\,pp at Ep.~10; Table~\ref{tab:app_supp_controls}; three failed FP16 C4 GPTQ rows without $\VerbMem$ are excluded).}
\label{tab:app_pareto}
\scriptsize
\resizebox{\textwidth}{!}{%
\begin{tabular}{c|ccc|ccc|ccc}
\toprule
 & \multicolumn{3}{c|}{\textbf{BF16-RN (\texttt{matched\_rn})}} & \multicolumn{3}{c|}{\textbf{BF16-SR (1-line writeback)}} & \multicolumn{3}{c}{\textbf{FP32-Master (BF16 export)}} \\
\textbf{Ep.} & \textbf{NLL (U/Q)} & \textbf{FP4 VM} & $\KnowMemR$ \textbf{(U/Q)} & \textbf{NLL (U/Q)} & \textbf{FP4 VM} & $\KnowMemR$ \textbf{(U/Q)} & \textbf{NLL (U/Q)} & \textbf{FP4 VM} & $\KnowMemR$ \textbf{(U/Q)} \\
\midrule
1 & 0.67/0.41 & 32.87 & 39.2/43.1 & 0.72/0.75 & 26.50 & 40.1/39.6 & 0.72/0.76 & 26.63 & 40.0/41.2 \\
2 & 0.82/0.40 & 33.99 & 38.9/45.0 & 15.44/11.99 & 2.44 & 38.5/38.0 & 15.36/13.83 & 0.95 & 36.7/40.6 \\
3 & 2.26/0.41 & 34.47 & 39.8/45.8 & 32.52/31.05 & 0.48 & \textbf{44.6}/40.8 & 35.18/32.94 & 0.02 & 43.9/39.4 \\
4 & 17.09/0.44 & 32.40 & 38.8/42.8 & 37.67/37.58 & \textbf{0.00} & 43.8/\textbf{40.9} & 38.74/37.71 & \textbf{0.00} & \textbf{44.1}/37.9 \\
5 & 34.28/0.49 & 31.42 & 41.3/44.5 & 39.43/38.80 & 0.00 & 39.1/35.0 & 39.35/38.19 & 0.00 & 43.1/40.2 \\
6 & 37.85/0.51 & 30.86 & 42.9/44.2 & 40.06/38.99 & 0.00 & 38.9/37.1 & 39.76/38.63 & 0.08 & 40.9/36.8 \\
7 & 40.86/0.54 & 30.13 & 42.8/45.2 & 40.06/39.23 & 0.11 & 38.7/31.4 & 40.17/38.66 & 0.10 & 37.1/36.5 \\
8 & 42.80/0.56 & 29.65 & 43.3/44.6 & 40.50/40.01 & 0.00 & 38.0/34.5 & 40.45/39.41 & 0.08 & 38.5/34.5 \\
9 & 43.96/0.55 & 29.00 & 43.1/45.0 & 40.70/40.16 & 0.00 & 37.2/31.3 & 40.55/38.10 & 0.94 & 38.5/33.7 \\
10 & \textbf{44.47}/0.55 & 28.52 & 44.0/44.9 & 40.83/40.13 & \textbf{0.00} & 38.7/33.5 & 40.73/40.09 & 0.09 & 36.8/31.9 \\
\bottomrule
\end{tabular}}
\end{table}

\begin{table}[htbp]
\centering
\caption{\textbf{Strict zero-\bfsixteen{} / zero-TF32 NPO causal surgery (top two blocks) and exact-\fpthirtytwo{} / zero-\bfsixteen{}-writeback GradDiff \& 6-band sextiles (bottom two blocks) on TOFU-1B.}
Top: IEEE-\fpthirtytwo{} NPO ($\eta{=}4{\times}10^{-5}$, $\mathrm{eff\_bs}{=}32$, $65$ st., $n{=}3$, \texttt{allow\_tf32=False}, global eligible fraction $9.3368\%$ across 112 linear layers matched identically per layer across masks, checksum \texttt{5075555177}); 5-ep.\ GradDiff surgery uses $\eta{=}2{\times}10^{-5}$ ($s_0$). Direct \fpthirtytwo{}-to-FP4 matches \bfsixteen{} export within $0.04$\,pp arm-mean (max $0.13$\,pp/seed on \texttt{large} $s_1$); near-zero masks rebound $+26.20$\,pp vs.\ $-0.90$/$-1.48$\,pp for random/large. Bottom: 10-ep.\ exact-\fpthirtytwo{} GradDiff ($\eta{=}10^{-5}$, $n{=}3$; final column reports both GPTQ-128 and AWQ-128 F) and six $16.67\%$ GradDiff sextiles $B_0$--$B_5$ ($\eta{=}10^{-5}$, $s_0$).}
\label{tab:app_eqcard_sextile}
\scriptsize
\resizebox{\textwidth}{!}{%
\begin{tabular}{lccccccc}
\toprule
\textbf{Strict Zero-BF16 / Zero-TF32 Arm ($K_L{=}9.34\%$)} & \textbf{FP32 / BF16 Chg.} & \textbf{RMS ($\times 10^{-5}$)} & \textbf{$\rho_{\mathrm{FP4}}$ / $\rho_{\mathrm{NF4}}$} & \textbf{Unq.\ \fpthirtytwo{} F / R} & \textbf{Direct FP4 F / R ($\Delta F$, $\Delta R$)} & \textbf{BF16exp FP4 F / R} & \textbf{Direct NF4 F / R} \\
\midrule
NPO \texttt{near\_zero\_2m9} ($n{=}3$, 5~ep.) & \textbf{9.34} / 9.27 & 8.11$\pm$0.04 & \textbf{0.175} / 0.912 & 13.84$\pm$0.29 / 60.49$\pm$0.73 & \textbf{40.04$\pm$1.35} / 46.14 (\textbf{+26.20$\pm$1.17}, $-$14.34) & 40.05 / 46.14 & 15.37 / 51.79 \\
NPO \texttt{random\_2m9} ($n{=}3$, 5~ep.) & \textbf{9.34} / 8.30 & 8.30$\pm$0.03 & \textbf{0.995} / 0.996 & 15.45$\pm$1.03 / 59.82$\pm$1.87 & \textbf{14.55$\pm$0.66} / 41.59 (\textbf{$-$0.90$\pm$0.37}, $-$18.22) & 14.53 / 41.59 & 15.73 / 50.57 \\
NPO \texttt{large\_2m9} ($n{=}3$, 5~ep.) & \textbf{9.34} / 6.54 & 8.58$\pm$0.04 & \textbf{1.021} / 1.014 & 16.70$\pm$1.12 / 59.97$\pm$1.40 & \textbf{15.22$\pm$1.04} / 40.89 (\textbf{$-$1.48$\pm$0.28}, $-$19.09) & 15.26 / 41.05 & 16.55 / 50.49 \\
\midrule
GradDiff \texttt{near\_zero} / \texttt{rand} / \texttt{large} ($s_0$, 5~ep.) & 9.34 / 9.2--3.7 & 3.62 / 3.72 / 3.80 & \textbf{0.167} / 0.996 / 1.022 & 2.22/28.7; 3.68/33.6; 4.67/35.6 & \textbf{36.48}/39.3; 4.60/8.2; 5.26/9.3 (\textbf{+34.26}/+0.92/+0.59) & 36.47/39.3; 4.58/8.2; 6.03/10.2 & 3.27/10.9; 4.81/13.1; 6.07/15.0 \\
\midrule
\textbf{Exact-\fpthirtytwo{} / Zero-BF16-Writeback GradDiff} & \textbf{BF16 Chg.\ (\%)} & \textbf{RMS ($\times 10^{-5}$)} & \textbf{$f_{\le 2^{-8}}$ / $f_{\mathrm{FP4c}}$ (\%)} & \textbf{Unq.\ F / R} & \textbf{FP4 F / R ($\Delta_{\mathrm{FP4}}$)} & \textbf{NF4 F / R} & \textbf{GPTQ-128 [AWQ-128] F} \\
\midrule
\texttt{near\_zero} / \texttt{random} / \texttt{large} (E10, $n{=}3$) & 9.16 / 6.87 / 3.35 & 3.25 / 3.45 / 3.84 & $f_{\mathrm{FP4c}}$: \textbf{99.4} / 24.3 / \textbf{0.0} & 1.52/31.3; 2.22/33.3; 4.97/36.8 & \textbf{36.70}/39.8; 3.62/7.3; 5.57/9.2 (\textbf{+35.18}/+1.41/+0.60) & 2.58/12.1; 3.73/12.9; 6.50/15.4 & 4.65/7.5; 2.82/5.0; 4.76/6.9 [AWQ: 2.79/13.1; 4.61/15.7; 7.24/18.0] \\
E04 epoch-matched ($s_0$): \texttt{near\_zero} / \texttt{random} / \texttt{large} & 9.10 / 6.19 / 2.23 & 2.33 / 2.54 / 2.80 & $f_{\mathrm{FP4c}}$: 99.4 / 26.5 / 0.0 & 9.06 / 14.57 / 31.34 & \textbf{39.25} / 9.40 / 17.27 (\textbf{+30.19} / $-$5.17 / $-$14.07) & 8.83 / 12.02 / 24.17 & --- \\
\midrule
Exact-\fpthirtytwo{} Sextile $B_0$ (lowest 16.7\% $|w_0|$, E05, $s_0$) & 15.86 & 3.27 & 98.12 / \textbf{94.50} & 1.41 / 31.06 & \textbf{11.35} / 20.81 (\textbf{+9.94}) & 1.57 / 8.53 & --- \\
Exact-\fpthirtytwo{} Sextiles $B_1$--$B_5$ (E05, $s_0$) & 14.2$\to$5.2 & 3.35$\to$3.87 & $f_{\mathrm{FP4c}}$: 17.2$\to$0.0 & 1.81$\to$5.44 & 0.73 / 5.87 / 4.30 / 4.08 / 6.11 ($-$1.08 to +3.86) & 3.87$\to$6.97 & --- \\
\bottomrule
\end{tabular}}
\end{table}

\begin{table}[htbp]
\centering
\caption{\textbf{Fixed-objective 5-LR $\times$ 3-seed WMDP-Cyber phase-boundary sweep on Zephyr-7B-beta (1,987 WMDP-Cyber + 500 MMLU MCQs, layers 5--7).}
Holding \texttt{Unlearn-Simple} NPO+GDR (100 st., $n{=}3$; \fpthirtytwo{} $n{=}1$) fixed, RN movement tracks $F_{\mathrm{L5\text{--}7}}(\tau)$ tightly with zero online certification violations across $3{,}260{,}871$ certified instances (summed across $3$ seeds $\times$ $5$ quantizers: Sym-INT4 $g{\in}\{64,128,2048\}$, NF4, and FP4 central cluster; e.g., $756/10.22\mathrm{M}$ FP4 groups per seed at $10^{-5}$) and jumps $+18.46$\,pp across $\eta^*{=}2^{-16}$ ($1.40{\to}1.65{\times}10^{-5}$). At $3{\times}10^{-6}$ and $10^{-5}$, \bfsr{} relieves pre-PTQ RN stall across all $3/3$ seeds (FP4 item-level McNemar $p{\le}3.8{\times}10^{-3}$; unquantized $10^{-5}$ $s_2$ reaches $p{=}0.0132$; RN shows no PTQ rebound itself), whereas tuned RN at $1.4$--$1.65{\times}10^{-5}$ sits descriptively close to $10^{-5}$ \bfsr{} without $3{\times}10^{-5}$'s $-5.0$\,pp FP4 MMLU drop ($-5.7$\,pp unq.\ vs.\ $10^{-5}$ \bfsr{}; against \emph{same-LR} \bfsr{}, $3{\times}10^{-5}$ FP4 SR$-$RN 95\% CIs stay in $[-3.73,+1.46]$\,pp, $p{=}0.24$--$0.50$, and TOST $\pm 3$\,pp passes on $2/3$ seeds at $1.40{\times}10^{-5}$ [$p{\le}0.024$] and $0/3$ at $1.65{\times}10^{-5}$). Bottom row logs \texttt{cais/wmdp} RMU ($\eta{=}5{\times}10^{-5}$, $s_0$).}
\label{tab:app_wmdp}
\scriptsize
\renewcommand{\arraystretch}{0.90}
\resizebox{\textwidth}{!}{%
\begin{tabular}{llcccccc}
\toprule
\textbf{Fixed NPO+GDR (L5--7, 100 st.)} & \textbf{Arm ($n{=}3$; \fpthirtytwo{} $n{=}1$)} & \textbf{Chg.\ L5--7 / $\le\tau$} & \textbf{Cert.\ $H$ / Fail} & \textbf{Unquant.\ Cyber / MMLU} & \textbf{4-bit FP4 Cyber / MMLU} & \textbf{4-bit NF4 Cyber / MMLU} & \textbf{SR--RN Paired $\Delta$ (McNemar)} \\
\midrule
Base $Q(W_0)$ & Un-unlearned Zephyr-7B & 0.00 / --- & --- & 39.76 / 61.60 & 39.20 / 58.20 & 39.41 / 60.40 & --- \\
\midrule
\multirow{3}{*}{\shortstack[l]{$\eta{=}3{\times}10^{-6}$, $\tau{=}2^{-10}$\\$F_{\mathrm{L5\text{--}7}}{=}26.10\%$}}
 & Triton \bfsixteen{}-RN & 24.92$\pm$0.01 / \textbf{100.0} & 3,253,766 / \textbf{0} & 39.74$\pm$0.30 / 61.40$\pm$0.53 & 39.15$\pm$0.23 / 58.73$\pm$0.64 & 39.41$\pm$0.21 / 60.33$\pm$0.12 & SR--RN Unq: $-$3.84 ($p{\le}3.5{\times}10^{-4}$, 3/3) \\
 & Triton \textbf{BF16+SR} & \textbf{87.57$\pm$0.04} / 28.9 & (unlocked) & \textbf{35.90$\pm$1.39} / 60.40$\pm$1.06 & \textbf{36.03$\pm$0.48} / \textbf{59.47$\pm$0.58} & \textbf{36.59$\pm$0.97} / 59.00$\pm$0.53 & \textbf{FP4 $-$3.12$\pm$0.61} ($p{\le}2.1{\times}10^{-3}$, 3/3) \\
 & \fpthirtytwo{} Master ($s_0$) & 84.04 / 30.1 & (unlocked) & 34.98 / 60.00 & 35.93 / 58.20 & 36.13 / 60.00 & ($s_0$ oracle ref.) \\
\midrule
\multirow{3}{*}{\shortstack[l]{$\eta{=}10^{-5}$, $\tau{=}2^{-8}$\\$F_{\mathrm{L5\text{--}7}}{=}80.40\%$}}
 & Triton \bfsixteen{}-RN & 78.19$\pm$0.02 / \textbf{100.0} & 3,582 / \textbf{0} & 34.02$\pm$0.48 / 59.40$\pm$1.78 & 33.35$\pm$1.01 / 59.07$\pm$1.42 & 33.74$\pm$0.03 / 58.60$\pm$0.69 & SR--RN NF4: $-$4.58 ($p{\le}0.013$, 3/3) \\
 & Triton \textbf{BF16+SR} & \textbf{95.43$\pm$0.01} / 81.6 & (unlocked) & \textbf{29.84$\pm$1.72} / 58.13$\pm$1.29 & \textbf{29.61$\pm$1.44} / \textbf{57.40$\pm$1.93} & \textbf{29.15$\pm$2.05} / \textbf{56.67$\pm$1.60} & \textbf{FP4 $-$3.74$\pm$0.43} ($p{\le}3.8{\times}10^{-3}$, 3/3) \\
 & \fpthirtytwo{} Master ($s_0$) & 94.88 / 81.9 & (unlocked) & 31.05 / 57.20 & 31.15 / 57.00 & 30.15 / 56.60 & ($s_0$ oracle ref.) \\
\midrule
\multirow{2}{*}{$\eta{=}1.40{\times}10^{-5}$ ($<\eta^*, \tau{=}2^{-8}$)}
 & Triton \bfsixteen{}-RN & \textbf{77.28$\pm$0.02} / \textbf{99.995} & 3,367 / \textbf{0} & 30.19$\pm$1.15 / 57.87$\pm$1.40 & 30.15$\pm$1.36 / 57.20$\pm$0.87 & 29.76$\pm$1.69 / 56.20$\pm$2.42 & Nearly flat ($+$40\% LR below $\eta^*$) \\
 & Triton \textbf{BF16+SR} & 96.58$\pm$0.02 / 81.3 & (unlocked) & 29.89$\pm$1.82 / 56.80$\pm$0.72 & 29.31$\pm$1.77 / 56.40$\pm$1.04 & 28.80$\pm$2.29 / 56.00$\pm$1.04 & FP4 $-$0.84$\pm$1.18 ($p{=}0.07$--$0.82$, 0/3) \\
\midrule
\multirow{2}{*}{$\eta{=}1.65{\times}10^{-5}$ ($>\eta^*, \tau{=}2^{-7}$)}
 & Triton \bfsixteen{}-RN & \textbf{95.74$\pm$0.02} / \textbf{100.0} & 93 / \textbf{0} & 29.63$\pm$1.82 / 58.00$\pm$1.44 & 30.00$\pm$1.27 / \textbf{57.67$\pm$0.61} & 29.04$\pm$1.33 / 56.47$\pm$1.81 & \textbf{$+$18.46\,pp jump} ($+$17.9\% LR past $\eta^*$) \\
 & Triton \textbf{BF16+SR} & 97.04$\pm$0.01 / 98.5 & (safe regime) & 29.76$\pm$0.52 / 56.60$\pm$0.20 & 29.59$\pm$0.75 / 56.27$\pm$1.10 & 29.17$\pm$0.25 / 55.40$\pm$0.60 & FP4 $-$0.40$\pm$2.01 ($p{=}0.12$--$0.28$, mixed) \\
\midrule
\multirow{3}{*}{\shortstack[l]{$\eta{=}3{\times}10^{-5}$, $\tau{=}2^{-7}$\\$F_{\mathrm{L5\text{--}7}}{=}98.33\%$}}
 & Triton \bfsixteen{}-RN & 96.08$\pm$0.03 / \textbf{99.99} & 63 / \textbf{0} & 29.93$\pm$1.23 / 52.47$\pm$2.84 & 30.25$\pm$1.25 / 52.40$\pm$1.78 & 30.34$\pm$0.75 / 51.60$\pm$1.71 & RN MMLU $-$5.0\,pp vs.\ $10^{-5}$ SR \\
 & Triton \textbf{BF16+SR} & 98.28$\pm$0.00 / 98.4 & (safe regime) & 27.81$\pm$1.17 / 49.20$\pm$6.66 & 29.09$\pm$1.09 / 48.53$\pm$6.22 & 28.42$\pm$1.59 / 48.13$\pm$5.93 & FP4 $-$1.16$\pm$0.28 ($p{=}0.24$--$0.50$, 0/3) \\
 & \fpthirtytwo{} Master ($s_0$) & 98.21 / 98.4 & (safe regime) & 28.99 / 51.40 & 28.69 / 51.00 & 28.89 / 51.40 & ($s_0$ oracle ref.) \\
\midrule
RMU ($5{\times}10^{-5}$, $s_0$) & RN / \textbf{BF16+SR} / \fpthirtytwo{} & 98.7 / 98.9 / 98.9 & (safe regime) & 35.1 / 35.2 / 35.2 & 36.4 / 36.3 / 35.6 & 36.1 / 36.2 / 35.4 & NF4 MMLU: 49.2 / \textbf{52.6} / 52.4 \\
\bottomrule
\end{tabular}}
\end{table}

\begin{table}[htbp]
\centering
\caption{\textbf{Complete causal weight surgery measurements on MUSE-News (Llama-2-7B, seed~0).}}
\label{tab:app_surgeries}
\scriptsize
\begin{tabular}{llcccc}
\toprule
\textbf{Surgery} & \textbf{Eval Mode} & $\VerbMem\downarrow$ & \textbf{Forget NLL} & $\PrivLeak$ & $\KnowMemR\uparrow$ \\
\midrule
Norm scaling $\lambda=0.35$ & Unquant.\ / FP4 & 44.91 / 34.74 & 0.31 / 0.38 & $-$99.64 / $-$99.77 & 48.41 / 45.86 \\
Norm scaling $\lambda=0.50$ & Unquant.\ / FP4 & 32.22 / 31.21 & 0.54 / 0.52 & $-$99.16 / $-$99.39 & 48.20 / 45.28 \\
Norm scaling $\lambda=0.57$ (RMS-matched) & Unquant.\ / FP4 & 26.68 / 26.72 & 1.27 / 0.69 & $-$92.27 / $-$98.97 & 45.59 / 41.65 \\
Norm scaling $\lambda=0.75$ (cliff edge) & Unquant.\ / FP4 & 4.65 / 17.56 & 23.27 / 4.51 & +109.56 / +23.60 & 43.97 / 35.89 \\
\midrule
Mask transplant \texttt{Inside} ($|w_0|\le 2^{-8}$) & Unquant.\ / FP4 & 51.29 / 44.44 & 0.27 / 0.32 & $-$99.81 / $-$99.81 & 50.43 / 49.51 \\
Mask transplant \texttt{Outside} ($|w_0|> 2^{-8}$) & Unquant.\ / FP4 & 0.45 / 8.60 & 34.34 / 20.23 & +109.56 / +107.71 & 42.67 / 34.14 \\
\midrule
Masked \fpthirtytwo{} train ($|w_0|\le 2^{-8}$), Ep.\ 4 & Unquant.\ / \textbf{FP4} & 0.16 / \textbf{17.04} & --- & +109.51 / \textbf{$-$30.95} & 38.11 / 45.50 \\
Masked \fpthirtytwo{} train ($|w_0|\le 2^{-8}$), Ep.\ 10 & Unq.\ FP32 / BF16 exp. & 0.00 / 0.00 & --- & +107.82 / +107.96 & 40.63 / 41.14 \\
Masked \fpthirtytwo{} train ($|w_0|\le 2^{-8}$), Ep.\ 10 & \textbf{FP4} / NF4 / 8-bit & \textbf{8.02} / 0.49 / 0.12 & --- & +70.68 / +109.2 / +108.9 & 41.73 / 43.12 / 39.60 \\
Masked \fpthirtytwo{} train ($|w_0|\le 2^{-8}$), Ep.\ 10 & \textbf{Per-Ch.\ GPTQ} & \textbf{18.85} & --- & \textbf{$-$28.37} & 35.15 \\
\bottomrule
\end{tabular}
\end{table}

\begin{table}[htbp]
\centering
\caption{\textbf{Full 3-LR and multi-$g$ $\mathrm{mean}(\hwzero) / \tau(\eta)$ sweep measurements on MUSE-News (Llama-2-7B, Epoch~10, Sym-INT4).} At $\eta{=}10^{-5}$ ($g{=}128$), crossing $\hwzero/\tau{\approx}1$ jumps $\VerbMem$ from $5.68$ to $30.83$ with intact retain utility ($\KnowMemR$ $40.9{\to}43.6$). At $5{\times}10^{-6}$ ($c{\le}0.60$) and $2{\times}10^{-5}$ (low $c$), heavy clipping degrades retain utility ($\KnowMemR$ shown explicitly), confounding low-$\hwzero/\tau$ rebound.}
\label{tab:app_multig}
\scriptsize
\resizebox{\textwidth}{!}{%
\begin{tabular}{cccccccccc}
\toprule
\textbf{LR $\eta$} & \textbf{Group $g$} & \textbf{Clip $c$} & $\hwzero$ ($\times 10^{-3}$) & $\hwzero / \tau(\eta)$ & $\rho$ & $\VerbMem$ & $\KnowMemR$ & $\PrivLeak$ & \textbf{Forget NLL} \\
\midrule
$5\times 10^{-6}$ & 128 & 0.40 / 0.50 / 0.60 & 1.44 / 1.80 / 2.16 & 0.74 / 0.92 / 1.10 & 0.71 / 0.53 / 0.29 & 0.03 / 5.97 / 7.95 & 0.0 / 0.0 / 0.0 & +11.0 / $-$16.1 / $-$11.6 & 11.65 / 6.76 / 5.30 \\
$5\times 10^{-6}$ & 128 & 0.70 / 0.85 / 1.00 & 2.51 / 3.05 / 3.59 & 1.29 / 1.56 / 1.84 & 0.15 / 0.09 / 0.06 & 12.99 / 30.71 / 45.90 & 9.9 / 41.9 / 48.0 & $-$49.9 / $-$99.5 / $-$99.8 & 3.18 / 0.54 / 0.33 \\
\midrule
$10^{-5}$ & 128 & 0.70 / 0.85 / 1.00 & 2.51 / 3.05 / 3.59 & 0.64 / 0.78 / 0.92 & 0.65 / 0.57 / 0.43 & 11.85 / 3.59 / 5.68 & 6.9 / 23.3 / 40.9 & $-$32.5 / +78.9 / +78.1 & 3.6 / 14.5 / 9.8 \\
$10^{-5}$ & 128 & \textbf{1.15} / 1.30 / 1.50 & \textbf{4.13} / 4.67 / 5.39 & \textbf{1.06} / 1.20 / 1.38 & \textbf{0.25} / 0.12 / 0.06 & \textbf{30.83} / 38.10 / 37.53 & \textbf{43.6} / 41.6 / 42.5 & \textbf{$-$99.7} / $-$99.8 / $-$99.8 & 0.46 / 0.37 / 0.39 \\
$10^{-5}$ & 128 & 1.75 / 2.00 (tail) & 6.29 / 7.18 & 1.61 / 1.84 & 0.03 / 0.02 & 28.23 / 30.17 & 35.3 / 32.9 & $-$99.8 / $-$99.7 & 0.53 / 0.54 \\
\midrule
$2\times 10^{-5}$ & 64 & 1.20 / 1.50 / 1.80 & 3.94 / 4.92 / 5.91 & 0.50 / 0.63 / 0.76 & 0.71 / 0.41 / 0.30 & 5.10 / 1.76 / 1.25 & 33.9 / 23.9 / 10.4 & +106 / +108 / +107 & 16.8 / 17.5 / 16.5 \\
$2\times 10^{-5}$ & 64 & 2.10 / \textbf{2.40} / 2.80 & 6.89 / \textbf{7.88} / 9.19 & 0.88 / \textbf{1.01} / 1.18 & 0.24 / \textbf{0.16} / 0.08 & 0.73 / \textbf{9.19} / 18.29 & 13.2 / \textbf{13.0} / 24.0 & +103 / \textbf{$-$87.5} / $-$98.9 & 9.5 / 2.2 / 1.2 \\
$2\times 10^{-5}$ & 128 & 1.50 / 1.80 / 2.00 & 5.39 / 6.46 / 7.18 & 0.69 / 0.83 / 0.92 & 0.33 / 0.27 / 0.22 & 0.82 / 1.30 / 2.15 & 15.9 / 7.8 / 8.8 & +108 / +105 / +91.2 & 16.9 / 13.6 / 6.6 \\
$2\times 10^{-5}$ & 128 & \textbf{2.20} / 2.40 / 2.80 & \textbf{7.90} / 8.62 / 10.06 & \textbf{1.01} / 1.10 / 1.29 & \textbf{0.16} / 0.10 / 0.04 & \textbf{9.95} / 12.64 / 19.44 & \textbf{12.3} / 15.5 / 23.0 & \textbf{$-$81.8} / $-$95.3 / $-$99.0 & 2.5 / 1.8 / 1.1 \\
$2\times 10^{-5}$ & 256 & 1.20 / 1.50 / 1.80 & 4.67 / 5.84 / 7.01 & 0.60 / 0.75 / 0.90 & 0.42 / 0.30 / 0.24 & 2.51 / 0.34 / 2.10 & 27.7 / 10.3 / 8.7 & +107 / +107 / +100 & 14.2 / 17.8 / 9.5 \\
$2\times 10^{-5}$ & 256 & \textbf{2.00} / 2.20 / 2.60 & \textbf{7.79} / 8.56 / 10.12 & \textbf{1.00} / 1.10 / 1.30 & \textbf{0.17} / 0.10 / 0.03 & \textbf{5.49} / 10.95 / 16.09 & \textbf{13.0} / 7.7 / 18.9 & \textbf{$-$7.1} / $-$74.6 / $-$97.0 & 3.3 / 2.9 / 1.5 \\
\bottomrule
\end{tabular}}
\end{table}

\begin{table}[htbp]
\centering
\caption{\textbf{Per-step writeback telemetry inside $(2^{-8}, 2^{-7}]$ and linear-warmup ablation straddling $\eta^* = 2^{-16}$ on TOFU-1B (125 st., 167.0M band params) and WMDP-7B (100 st., 117.4M band params; seed~0, pure-\bfsixteen{} RN).}
Under constant LR, crossing $\eta^*$ unlocks $98.5\%$--$99.2\%$ of $(2^{-8},2^{-7}]$ on Step~1 via Adam's $\kappa_1{\approx}0.99$ impulse ($69.5\%$--$72.8\%$ flipping only once). Under linear warmup, Step~1's impulse vanishes and $99.1\%$--$99.5\%$ of the band stays frozen even at peak $\eta{=}1.65{\times}10^{-5} > \eta^*$.}
\label{tab:app_warmup}
\scriptsize
\resizebox{\textwidth}{!}{%
\begin{tabular}{llcccccccc}
\toprule
\textbf{Model \& Schedule} & \textbf{Peak $\eta$} & \textbf{Total Chg.} & \textbf{Band Step-1} & \textbf{Band Final} & \textbf{1-Flip / $\ge 2$-Flip} & \textbf{Band RMS (S1 $\to$ Fin)} & \textbf{Mean ULP} & \textbf{Later p95 $|\delta w|$} & \textbf{TOFU Unq $\to$ FP4 (F/R)} \\
\midrule
TOFU-1B Constant ($0$ wu) & $1.40{\times}10^{-5}$ & 17.85\% & 0.00\% & 0.19\% & 0.10\% / 0.09\% & $0.0 \to 3.81{\times}10^{-6}$ & 2.17 & $1.11{\times}10^{-5}$ & 7.94/83.1 $\to$ 19.20/50.8 \\
TOFU-1B Constant ($0$ wu) & $1.65{\times}10^{-5}$ & \textbf{34.82\%} & \textbf{98.47\%} & \textbf{98.14\%} & \textbf{69.53\%} / 28.97\% & $3.03 \to 5.25{\times}10^{-5}$ & 1.46 & $1.29{\times}10^{-5}$ & 5.80/83.1 $\to$ 12.29/50.5 \\
TOFU-1B 1-ep Warmup (26 st.) & $1.40{\times}10^{-5}$ & 11.62\% & 0.00\% & 0.09\% & 0.04\% / 0.06\% & $0.0 \to 2.50{\times}10^{-6}$ & 2.30 & $4.35{\times}10^{-6}$ & 19.44/45.2 $\to$ 22.43/28.7 \\
TOFU-1B 1-ep Warmup (26 st.) & $1.65{\times}10^{-5}$ & \textbf{13.67\%} & \textbf{0.00\%} & \textbf{0.93\%} & 0.31\% / 0.62\% & $0.0 \to 9.55{\times}10^{-6}$ & 2.67 & $5.08{\times}10^{-6}$ & 16.51/59.6 $\to$ 20.75/32.9 \\
\midrule
WMDP-7B Constant ($0$ wu) & $1.40{\times}10^{-5}$ & 77.29\% & 0.00\% & 0.02\% & 0.01\% / 0.004\% & $0.0 \to 6.80{\times}10^{-7}$ & 1.45 & $1.04{\times}10^{-5}$ & --- \\
WMDP-7B Constant ($0$ wu) & $1.65{\times}10^{-5}$ & \textbf{95.73\%} & \textbf{99.24\%} & \textbf{99.11\%} & \textbf{72.84\%} / 26.41\% & $3.04 \to 4.85{\times}10^{-5}$ & 1.38 & $1.23{\times}10^{-5}$ & --- \\
WMDP-7B 10\% Warmup (10 st.) & $1.40{\times}10^{-5}$ & 65.82\% & 0.00\% & 0.02\% & 0.01\% / 0.004\% & $0.0 \to 6.80{\times}10^{-7}$ & 1.44 & $7.08{\times}10^{-6}$ & --- \\
WMDP-7B 10\% Warmup (10 st.) & $1.65{\times}10^{-5}$ & \textbf{73.76\%} & \textbf{0.00\%} & \textbf{0.50\%} & 0.32\% / 0.19\% & $0.0 \to 4.58{\times}10^{-6}$ & 1.70 & $8.28{\times}10^{-6}$ & --- \\
\bottomrule
\end{tabular}}
\end{table}

\begin{table}[htbp]
\centering
\caption{\textbf{Five-operator compression audit (FP4, NF4, OCP MXFP4 $g{=}32$, Magnitude Pruning $20\%$/$30\%$) and same-checkpoint cross-domain validation.}
Top: balanced $n{=}3$ strict zero-\bfsixteen{}/zero-TF32 \fpthirtytwo{} NPO surgery on TOFU-1B ($\eta{=}4{\times}10^{-5}$, $\mathrm{eff\_bs}{=}16$, 125 st., $\mathrm{wd}{=}0$, $9.34\%$ per layer). Second: 9 natural TOFU-1B runs ($s_0$). Third: 5 same-checkpoint WMDP-Cyber Zephyr-7B runs ($s_0$). Bottom: paired $t$-tests ($df{=}2$) and McNemar tests on Qwen2.5 refusal-template SFT (Table~\ref{tab:safety_universal}). Detailed base references and protocol notes appear below.}
\label{tab:app_universal}
\scriptsize
\resizebox{\textwidth}{!}{%
\begin{tabular}{llcccccccc}
\toprule
\textbf{Suite / Condition} & \textbf{Arm / Seed} & \textbf{Unq.\ F (or Cyber/MMLU)} & \textbf{$\Delta_{\mathrm{FP4}}$} & \textbf{$\Delta_{\mathrm{NF4}}$} & \textbf{$\Delta_{\mathrm{MXFP4}}$} & \textbf{$\Delta_{\mathrm{P20}}$ / $\Delta_{\mathrm{P30}}$} & \textbf{$\LMS_{\mathrm{FP4}}$ / $\LMS_{E,\mathrm{FP4}}$} & \textbf{$\LMS_{E,\mathrm{P20}}$} & \textbf{$\rho_{\mathrm{FP4}}$ / $\rho_{\mathrm{MX}}$ / $\rho_{\mathrm{P20}}$} \\
\midrule
\multirow{3}{*}{\shortstack[l]{TOFU-1B Strict\\Zero-BF16 NPO\\($9.34\%$, $n{=}3$)}}
 & \texttt{near\_zero\_2m9} ($n{=}3$) & 7.83$\pm$0.64 / 78.0 & \textbf{+33.18$\pm$2.04} & +2.55$\pm$0.31 & +1.86$\pm$0.25 & \textbf{+39.37$\pm$0.33} / +11.72$\pm$0.64 & 0.006 / \textbf{0.006} & \textbf{0.0004} & 0.180 / 0.944 / \textbf{0.0043} \\
 & \texttt{random\_2m9} ($n{=}3$) & 8.39$\pm$0.56 / 78.7 & +0.73$\pm$0.09 & +1.59$\pm$0.27 & +0.79$\pm$0.24 & +0.54$\pm$0.57 / $-$1.19$\pm$0.17 & 0.814 / 0.814 & 0.7997 & 0.995 / 0.981 / 0.9945 \\
 & \texttt{large\_2m9} ($n{=}3$) & 8.80$\pm$0.59 / 78.2 & +0.26$\pm$0.19 & +1.12$\pm$0.27 & +1.03$\pm$0.39 & +0.22$\pm$0.52 / $-$1.79$\pm$0.18 & 1.000 / 1.000 & 1.0000 & 1.021 / 0.986 / 1.0000 \\
\midrule
\multirow{9}{*}{\shortstack[l]{TOFU-1B Natural\\(5 ep., 125 st., $s_0$,\\$\mathrm{eff\_bs}{=}16$, lin.)}}
 & NPO RN $10^{-5}$ wu & 22.81 / 31.33 & +4.72 & +1.23 & $-$1.48 & \textbf{+20.47} / $-$1.26 & 0.025 / \textbf{0.029} & \textbf{0.0105} & 0.337 / 0.607 / \textbf{0.0696} \\
 & NPO \textbf{SR} $10^{-5}$ wu & 4.07 / 77.69 & +0.42 & +0.90 & +0.76 & \textbf{+0.72} / +0.30 & 0.758 / \textbf{0.926} & 0.9209 & 1.004 / 0.967 / 1.0224 \\
 & NPO FP32 $10^{-5}$ wu & 4.46 / 74.51 & +0.60 & +0.94 & +0.70 & +0.69 / +0.36 & 0.717 / 0.827 & 0.8141 & 0.993 / 0.986 / 0.9968 \\
 & NPO RN $1.40{\times}10^{-5}$ const & 8.16 / 83.09 & \textbf{+11.23} & +2.97 & +3.13 & \textbf{+33.82} / +13.49 & 0.105 / \textbf{0.063} & \textbf{0.0154} & 0.590 / 0.888 / 0.1357 \\
 & NPO RN $1.65{\times}10^{-5}$ const (1-ULP) & 5.99 / 83.11 & \textbf{+6.57} & +2.81 & +1.71 & \textbf{+28.88} / +24.11 & \underline{0.483} / \textbf{0.127} & \textbf{0.0728} & 0.722 / 0.944 / 0.2598 \\
 & NPO RN $1.65{\times}10^{-5}$ wu & 17.36 / 58.57 & +2.65 & +2.06 & $-$1.20 & \textbf{+11.54} / +6.14 & 0.085 / \textbf{0.059} & \textbf{0.0231} & 0.499 / 0.812 / 0.1327 \\
 & GradDiff RN $10^{-5}$ wu & 18.62 / 47.85 & \textbf{+21.99} & \textbf{+18.35} & $-$0.27 & \textbf{+26.70} / +1.10 & 0.005 / \textbf{0.003} & \textbf{0.0004} & 0.242 / 0.402 / \textbf{0.0028} \\
 & GradDiff \textbf{SR} $10^{-5}$ wu & 0.00 / 60.25 & \textbf{+0.00} & \textbf{+0.00} & +0.00 & \textbf{+0.00} / +0.16 & 0.749 / \textbf{0.920} & 0.9142 & 1.003 / 0.968 / 1.0211 \\
 & GradDiff FP32 $10^{-5}$ wu & 0.00 / 55.35 & +0.00 & +0.00 & +0.00 & +0.00 / +0.03 & 0.711 / 0.825 & 0.8124 & 0.994 / 0.987 / 0.9966 \\
\midrule
\multirow{5}{*}{\shortstack[l]{WMDP-Cyber\\Zephyr-7B L5--7\\(100 st., $s_0$,\\same-ckpt.\ eval)}}
 & RN $10^{-5}$ ($\LMSnorm_{2^{-8}}{=}0.000$) & 34.37 / 60.60 & \textbf{$-$0.65} & --- & --- & \textbf{$-$0.40} / --- & \textbf{0.768} / \textbf{0.602} & \textbf{0.5617} & \textbf{1.046} / --- / \textbf{0.9900} \\
 & SR $10^{-5}$ ($\LMSnorm_{2^{-8}}{=}0.940$) & 29.34 / 57.80 & $-$0.20 & --- & --- & $-$0.70 / --- & 0.810 / 0.846 & 0.8315 & 0.992 / --- / 0.9983 \\
 & FP32 $10^{-5}$ ($\LMSnorm_{2^{-8}}{=}0.923$) & 30.05 / 56.80 & $-$1.62 & --- & --- & $-$2.07 / --- & 0.808 / 0.817 & 0.8000 & 0.995 / --- / 0.9953 \\
 & RN $1.40{\times}10^{-5}$ ($\LMSnorm_{2^{-8}}{=}0.0002$) & 29.29 / 57.00 & \textbf{+0.30} & --- & --- & \textbf{+0.00} / --- & \textbf{0.764} / \textbf{0.682} & \textbf{0.6515} & \textbf{1.007} / --- / \textbf{0.9937} \\
 & SR $1.40{\times}10^{-5}$ ($\LMSnorm_{2^{-8}}{=}0.954$) & 26.62 / 57.00 & $-$0.15 & --- & --- & $-$0.40 / --- & 0.811 / 0.837 & 0.8221 & 0.993 / --- / 0.9972 \\
\midrule
\multirow{3}{*}{\shortstack[l]{Refusal SFT\\Qwen2.5 ($n{=}3$,\\same-ckpt.)}}
 & Prob.\ DiD ($\Delta_{\mathrm{SR}}{-}\Delta_{\mathrm{RN}}$) & $+0.54$ unq & \textbf{+78.42} ($2{\times}10^{-9}$) & \textbf{+22.27} ($1.1{\times}10^{-3}$) & \textbf{+19.62} ($2.9{\times}10^{-3}$) & \textbf{+84.40} ($1.2{\times}10^{-6}$) / --- & RN $\LMS_E$: 0.003 & 0.0000 & RN $\rho$: 0.241 / --- / 0.0001 \\
 & Gen.\ DiD ($\Delta_{\mathrm{SR}}{-}\Delta_{\mathrm{RN}}$, $t$-test) & $+0.3$ U ($-$2.0 B$_{\mathrm{Alp}}$) & \textbf{+51.0} ($5.1{\times}10^{-4}$) & +2.7 ($0.015$) & \textbf{+13.0} ($0.039$) & \textbf{+86.7} ($1.5{\times}10^{-5}$) / --- & SR $\LMS_E$: 0.930 & 0.9273 & SR $\rho$: 0.998 / --- / 1.0139 \\
 & Gen.\ McNemar SR $-$ RN ($3/3$ seeds) & $p{=}1.0$ unq & \textbf{+49..+53} ($p{\le}3.6{\times}10^{-15}$) & +3 ($p{=}0.25$) & \textbf{+8..+17} ($p{\le}7.8{\times}10^{-3}$) & \textbf{+87} ($p{=}1.3{\times}10^{-26}$) / --- & FP32 $\LMS_E$: 0.825 & 0.8168 & FP32 $\rho$: 0.991 / --- / 0.9957 \\
\bottomrule
\end{tabular}}
\end{table}

\paragraph{Audit and provenance notes (Tables~\ref{tab:safety_universal},~\ref{tab:app_eqcard_sextile},~\ref{tab:app_universal}).}
In the five-operator surgery (Table~\ref{tab:app_universal} top; base $Q(W_0)$ F/R: Unq $87.27/86.19$, FP4 $50.83/50.04$, NF4 $69.67/67.88$, MXFP4 $51.08/51.00$, P20 $46.03/46.71$, P30 $19.54/18.75$), near-zero vs.\ random/large paired $t$-tests ($df{=}2$; min exact seed-sign permutation $p{=}0.25$) yield P20 $p{=}1.53{\times}10^{-5}$ / $4.09{\times}10^{-5}$, FP4 $p{=}0.00131$ / $0.00141$, and NF4 $p{=}0.0027$ (near-zero vs.\ large); per-seed random P20 shifts are $+0.53$, $-0.03$, and $+1.11$\,pp.
Differences between Table~\ref{tab:app_universal}'s natural TOFU-1B rows ($5$~ep., $\mathrm{eff\_bs}{=}16$, 112 linear layers, in-memory \texttt{bnb.functional}) and Tables~\ref{tab:app_w13_align},~\ref{tab:w5_controls}, and~\ref{tab:app_warmup} reflect separate independent retraining passes (different batch/schedule/layer scopes and non-deterministic \bfsixteen{} kernels).
On the refusal-template SFT probe (Table~\ref{tab:safety_universal}), base generation U[B$_{\mathrm{Alp}}$/B$_{\mathrm{XS}}$] rates are $6[1/1]$ (Unq), $20[0/3]$ (FP4), $22[1/9]$ (NF4), $10[0/3]$ (MXFP4), and $13[0/3]$ (P20); an author manual inspection of the saved seed-0 completions (first $15$/split across $4$ operators excl.\ NF4; not a logged JSONL field) found $534/540{=}98.9\%$ concordance with the case-insensitive refusal-phrase substring rule (the $6$ false positives on benign Alpaca prompts are four ``\emph{I can't help but}'' and two ``\emph{I can't believe}'' idioms).

\begin{table}[htbp]
\centering
\caption{\textbf{Quantitative Proposition~\ref{prop:rho_q} mediation audit on strict zero-\bfsixteen{}/zero-TF32 NPO surgery (fresh deterministic $s_0$ run at $\eta{=}4{\times}10^{-5}$, $\mathrm{micro\_bs}{=}2$, $\mathrm{eff\_bs}{=}16$, 125 st.).}
Computes exact $\rho_{\mathrm{tot}}{=}\rho_{\mathrm{fix}}{+}\rho_{\mathrm{scl}}$ (matching Table~\ref{tab:app_universal}'s $s_0$ $\rho_{\mathrm{FP4}}{=}0.1802$ to 4 decimals under deterministic training) and post-hoc smooth-density summary $\hat{\rho}_{\mathrm{smooth}}$ (grid-proportional KDE bandwidth, no fitted regression coefficients): across all 9 mask$\times$quantizer cells, $R^2{=}0.9919$, $\mathrm{MAE}{=}0.0123$, $\mathrm{RMSE}{=}0.0215$ (excluding high-leverage near-zero FP4: $R^2{=}0.7689$, $\mathrm{MAE}{=}0.0132$). Near-zero \texttt{bnb}-FP4 has \emph{more} crossings ($11.34\%$) than NF4/MXFP4 ($7.52\%$/$7.91\%$) yet collapses to $\rho_{\mathrm{fix}}{=}0.180$ because crossings span only $M/192$.}
\label{tab:app_prop2_mediation}
\scriptsize
\resizebox{\textwidth}{!}{%
\begin{tabular}{llcccccc}
\toprule
\textbf{Surgery Mask ($K_L{=}9.34\%$)} & \textbf{Operator} & \textbf{Exact $\rho_{\mathrm{tot}}$} & \textbf{Exact $\rho_{\mathrm{fix}}$ (or $\rho_{\mathrm{ret}}$)} & \textbf{Exact $\rho_{\mathrm{scl}}$ (or $\rho_{\mathrm{flip}}$)} & \textbf{Prop.~\ref{prop:rho_q} $\hat{\rho}_{\mathrm{smooth}}$ (or $\LMS_{E,W_0}$)} & \textbf{Pred.\ Err.\ ($\hat{\rho}-\rho_{\mathrm{fix}}$)} & \textbf{Crossing Rate on Moved (\%)} \\
\midrule
\multirow{4}{*}{\texttt{near\_zero\_2m9}}
 & \texttt{bnb}-FP4 ($g{=}64$) & \textbf{0.1802} & \textbf{0.1802} & 0.0000 & \textbf{0.1747} & $-$0.0055 & \textbf{11.335\%} (tiny jump $M/192$) \\
 & \texttt{bnb}-NF4 ($g{=}64$) & 0.9027 & 0.9027 & 0.0000 & 0.9510 & +0.0483 & 7.515\% (wide jump) \\
 & OCP MXFP4 ($g{=}32$) & \textbf{0.9440} & \textbf{0.9441} & 0.0000 & \textbf{0.9392} & $-$0.0049 & 7.911\% (wide jump $0.5s$) \\
 & MagPrune-20\% (non-staircase) & \textbf{0.0044} & 0.0002 ($\rho_{\mathrm{ret}}$) & 0.0042 ($\rho_{\mathrm{flip}}$) & 0.0004 ($\LMS_{E,W_0}$) & --- & --- \\
\midrule
\multirow{4}{*}{\texttt{random\_2m9}}
 & \texttt{bnb}-FP4 ($g{=}64$) & 0.9945 & 0.9786 & +0.0159 & \textbf{0.9788} & +0.0002 & 5.465\% \\
 & \texttt{bnb}-NF4 ($g{=}64$) & 0.9961 & 0.9804 & +0.0157 & \textbf{0.9785} & $-$0.0019 & 5.748\% \\
 & OCP MXFP4 ($g{=}32$) & 0.9807 & 0.9709 & +0.0098 & \textbf{0.9754} & +0.0045 & 5.794\% \\
 & MagPrune-20\% (non-staircase) & 0.9944 & 0.7878 ($\rho_{\mathrm{ret}}$) & 0.2066 ($\rho_{\mathrm{flip}}$) & 0.7997 ($\LMS_{E,W_0}$) & --- & --- \\
\midrule
\multirow{4}{*}{\texttt{large\_2m9}}
 & \texttt{bnb}-FP4 ($g{=}64$) & 1.0200 & 0.8643 & \textbf{+0.1556} & \textbf{0.8630} & $-$0.0013 & 1.872\% \\
 & \texttt{bnb}-NF4 ($g{=}64$) & 1.0141 & 0.8572 & \textbf{+0.1569} & \textbf{0.8598} & +0.0026 & 2.221\% \\
 & OCP MXFP4 ($g{=}32$) & 0.9870 & 0.8991 & +0.0879 & 0.9408 & +0.0417 & 2.019\% \\
 & MagPrune-20\% (non-staircase) & 1.0000 & 1.0000 ($\rho_{\mathrm{ret}}$) & 0.0000 ($\rho_{\mathrm{flip}}$) & 1.0000 ($\LMS_{E,W_0}$) & --- & --- \\
\bottomrule
\end{tabular}}
\end{table}

\begin{table}[htbp]
\centering
\caption{\textbf{Complete 2026-era dual-flagship validation on TOFU-forget10 (21 logged runs on A100-80GB).}
Top: \texttt{google/gemma-4-E4B-it} (all 42 transformer layers, $3.95\mathrm{B}$ 2D linear weights, embeddings frozen, 2-ep.\ target SFT at $3{\times}10^{-5}$, 5-ep.\ unlearning at $\eta{=}10^{-5}$, $\mathrm{eff\_bs}{=}32$, $F(2^{-8}){=}15.70\%$; $n{=}3$ RN/SR + $n{=}1$ \fpthirtytwo{} for both NPO and GradDiff).
Bottom: \texttt{Qwen/Qwen3.8-27B} (hybrid Gated-DeltaNet + Attention, localized layers 52--55 = $1.52\mathrm{B}$ trained 2D weights, full 64-layer 4-bit/pruning eval, 2-ep.\ target SFT at $5{\times}10^{-5}$, 5-ep.\ GradDiff at $\eta{=}10^{-5}$, $\mathrm{eff\_bs}{=}32$, $F(2^{-8}){=}24.67\%$; $n{=}3$ RN + $n{=}2$ SR, as runtime expiry truncated $s_2$ SR and \fpthirtytwo{}). All numbers report mean$\pm$sample SD ($ddof{=}1$).}
\label{tab:app_w17_2026}
\scriptsize
\resizebox{\textwidth}{!}{%
\begin{tabular}{llcccccc}
\toprule
\textbf{Model / Method} & \textbf{Arm ($n$)} & \textbf{Chg.\ / $\le\!2^{-8}$ (\%)} & \textbf{$\rho$ (FP4 / NF4 / P20)} & \textbf{Unquant.\ F / R} & \textbf{4-bit FP4 F / R ($\Delta$)} & \textbf{4-bit NF4 F / R ($\Delta$)} & \textbf{Prune-20\% F / R ($\Delta$)} \\
\midrule
\multirow{7}{*}{\shortstack[l]{Gemma-4-E4B\\(42~L, $3.95\mathrm{B}$)}}
 & Base Target ($s_0$) & 0.00 / --- & --- & 63.03 / 60.19 & 59.44 / 56.96 & 62.17 / 59.43 & 62.06 / 59.28 \\
 & NPO \bfsixteen{}-RN ($n{=}3$) & 6.33$\pm$0.03 / \textbf{100.0} & 0.256 / 0.370 / \textbf{0.016} & 47.45$\pm$0.27 / 49.28$\pm$0.24 & \textbf{54.92$\pm$0.16} / 54.33 (\textbf{+7.47}) & \textbf{52.49$\pm$0.23} / 52.68 (\textbf{+5.04}) & \textbf{60.13$\pm$0.13} / 57.98 (\textbf{+12.69}) \\
 & NPO \textbf{BF16+SR} ($n{=}3$) & 59.18$\pm$0.12 / 24.76 & 0.998 / 0.997 / 1.018 & \textbf{6.18$\pm$0.10} / \textbf{52.43$\pm$0.78} & \textbf{5.30$\pm$0.19} / 48.73 ($-$0.88) & \textbf{5.98$\pm$0.21} / 51.39 ($-$0.20) & \textbf{6.48$\pm$0.15} / 51.56 (+0.30) \\
 & NPO \fpthirtytwo{} ($n{=}1$) & 50.89 / 28.80 & 0.993 / 0.995 / 0.997 & 8.32 / 45.67 & 7.05 / 43.06 ($-$1.27) & 7.92 / 45.11 ($-$0.40) & 8.57 / 44.90 (+0.25) \\
 & GD \bfsixteen{}-RN ($n{=}3$) & 6.26$\pm$0.01 / \textbf{100.0} & 0.271 / 0.260 / \textbf{0.010} & 58.18$\pm$0.06 / 61.20$\pm$0.12 & 58.01$\pm$0.02 / 57.18 ($-$0.17) & 59.89$\pm$0.04 / 59.86 (+1.71) & \textbf{61.66$\pm$0.01} / 59.32 (\textbf{+3.48}) \\
 & GD \textbf{BF16+SR} ($n{=}3$) & 58.92$\pm$0.18 / 24.92 & 0.998 / 0.997 / 1.017 & 0.00$\pm$0.01 / 27.02$\pm$0.76 & 0.02$\pm$0.02 / 13.27 (+0.02) & 0.00$\pm$0.01 / 20.55 (+0.00) & 0.01$\pm$0.01 / 26.67 (+0.01) \\
 & GD \fpthirtytwo{} ($n{=}1$) & 51.51 / 28.49 & 0.993 / 0.995 / 0.997 & 0.01 / 16.04 & 0.01 / 3.22 (+0.00) & 0.00 / 7.90 ($-$0.01) & 0.03 / 16.51 (+0.02) \\
\midrule
\multirow{3}{*}{\shortstack[l]{Qwen3.8-27B\\(L52--55, $1.52\mathrm{B}$)}}
 & Base Target ($s_0$) & 0.00 / --- & --- & 61.26 / 59.70 & 59.13 / 57.60 & 59.71 / 58.27 & 59.90 / 58.37 \\
 & GD \bfsixteen{}-RN ($n{=}3$) & 9.89$\pm$0.01 / \textbf{100.0} & 0.196 / 0.710 / \textbf{0.006} & 58.36$\pm$0.01 / 60.13$\pm$0.04 & 58.54$\pm$0.00 / 57.70 (+0.18) & 57.34$\pm$0.02 / 58.57 ($-$1.02) & \textbf{59.84$\pm$0.01} / 58.37 (\textbf{+1.48}) \\
 & GD \textbf{BF16+SR} ($n{=}2$) & 72.12$\pm$0.07 / 32.34 & 0.995 / 0.997 / 1.011 & \textbf{8.38$\pm$0.39} / 40.74$\pm$0.18 & \textbf{9.04$\pm$0.31} / 37.03 (+0.67) & \textbf{8.99$\pm$0.31} / 38.98 (+0.62) & \textbf{8.36$\pm$0.36} / 37.97 ($-$0.02) \\
\bottomrule
\end{tabular}}
\end{table}

\begin{table}[htbp]
\centering
\caption{\textbf{Supplemental verification controls cited in \S\ref{sec:causal}--\ref{sec:ecosystem}: MUSE-Books ($n{=}3$), MUSE-News LR / $\KnowMemF$ / group-128 controls, and TOFU-1B 10-epoch constant-LR intra-binade sweep (Figure~\ref{fig:theorem1}a).}
Every secondary metric referenced in the main text is reported verbatim here: MUSE-Books Ep.~5 ($n{=}3$); MUSE-News Ep.~10 LR checks ($5{\times}10^{-6}$, $2{\times}10^{-5}$, $10^{-4}$), native RN $\KnowMemF$ ($n{=}3$), and group-$128$ GPTQ/AWQ ($n{=}2$, seeds~1--2, trajectory harness); and the 5-point 10-epoch TOFU-1B constant-LR sweep ($s_0$) establishing the $131\times$ $[2^{-8},2^{-7})$ activation jump ($0.750\%{\to}98.263\%$) and $26.25{\to}7.36$\,pp rebound drop at $99.66\%$ near-zero support.}
\label{tab:app_supp_controls}
\scriptsize
\resizebox{\textwidth}{!}{%
\begin{tabular}{llccccc}
\toprule
\textbf{Suite / Control} & \textbf{Arm / Setting} & \textbf{Chg.\ / $\le\!2^{-8}$ (\%)} & \textbf{Unquantized} & \textbf{4-bit FP4 ($\Delta$)} & \textbf{4-bit NF4} & \textbf{Other / Band Stat.} \\
\midrule
\multirow{4}{*}{\shortstack[l]{MUSE-Books\\(Ep.~5, $n{=}3$,\\NPO+GDR)}}
 & \texttt{Target} / \texttt{Retrain} (ref.) & --- & VM 99.7 / 14.0 (KR 69.4 / 68.1) & VM 94.6 / 14.1 (KR 50.2 / 62.1) & VM 98.1 / 14.7 & PL $-$57.2 / $-$0.02 \\
 & \bfsixteen{}-RN ($n{=}3$) & --- & VM 0.00 (KR 56.2$\pm$1.6) & \textbf{VM 28.16$\pm$0.54} (KR 54.2$\pm$0.7) & VM 0.00 & FP4 PL $-$58.8 \\
 & \textbf{BF16+SR} ($n{=}3$) & --- & VM 0.00 (KR 32.2$\pm$21.7) & \textbf{VM 0.00$\pm$0.00} (KR 41.5$\pm$11.6) & VM 0.00 & FP4 PL $-$36.8 \\
 & \fpthirtytwo{} Master ($n{=}3$) & --- & VM 0.00 (KR 51.5$\pm$10.6) & \textbf{VM 0.00$\pm$0.00} (KR 45.0$\pm$5.1) & VM 0.00 & FP4 PL $-$38.0 \\
\midrule
\multirow{6}{*}{\shortstack[l]{MUSE-News\\Ep.~10 LR,\\$\KnowMemF$,\\g128 \& RTN}}
 & Native RN $\eta{=}10^{-5}$ ($n{=}3$, $\KnowMemF$) & 17.07 / 98.94 & $\KnowMemF$ 49.44 (VM 0.00) & $\KnowMemF$ 48.35 (VM 29.07) & --- & Flat QA forget \\
 & RN $\eta{=}5{\times}10^{-6}$ ($s_0$) & 11.46 / 99.84 & VM 1.02 (KR 35.22) & \textbf{VM 44.55} (PL $-$99.8) & \textbf{VM 38.35} (PL $-$99.8) & GPTQ-Ch VM 25.34 \\
 & RN $\eta{=}2{\times}10^{-5}$ ($s_0$) & 32.12 / --- & VM 0.00 (KR 35.64) & \textbf{VM 0.90} (PL +109.5) & VM 0.00 & GPTQ-Ch VM 4.67 \\
 & RN $\eta{=}10^{-4}$ ($s_0$, traj.\ harness) & --- & VM 0.00 (KR 22.82) & \textbf{VM 0.00} (PL +107.7) & VM 0.00 & Ep.~2 FP4 $+1.1$\,pp \\
 & RN $\eta{=}10^{-5}$ g128 GPTQ/AWQ ($n{=}2$) & --- & VM 0.00 (seeds 1, 2) & --- & --- & GPTQ-128 $[0.00, 0.10]$; AWQ $[0.00, 0.21]$ \\
 & Coarse Sym-INT4 RTN ($s_0$, RN vs.\ SR) & --- & VM 0.00 / 0.00 & --- & --- & $g{=}512$: $30.28{\to}15.05$ ($-50.3\%$); Ch: $24.60{\to}17.35$ ($-29.5\%$) \\
\midrule
\multirow{5}{*}{\shortstack[l]{TOFU-1B\\10-Ep.\ Const.\\LR Sweep\\($s_0$, Fig.~\ref{fig:theorem1}a)}}
 & RN $\eta{=}1.00{\times}10^{-5}$ & 17.87 / 100.00 & F 9.21 / R 86.42 & \textbf{35.46} / 44.40 (\textbf{+26.25}) & 12.10 / 66.05 & $[2^{-8},2^{-7})$: 0.322\% \\
 & RN $\eta{=}1.25{\times}10^{-5}$ & 17.81 / 99.93 & F 6.47 / R 88.42 & 17.72 / 52.93 (+11.25) & 8.48 / 66.54 & $[2^{-8},2^{-7})$: 0.442\% \\
 & RN $\eta{=}1.40{\times}10^{-5}$ & 17.95 / \textbf{99.66} & F 6.20 / R 88.48 & 13.56 / 53.87 (\textbf{+7.36}) & 8.19 / 68.35 & $[2^{-8},2^{-7})$: \textbf{0.750\%} \\
 & RN $\eta{=}1.65{\times}10^{-5}$ ($>\eta^*$) & 34.80 / 51.75 & F 5.38 / R 88.25 & 9.11 / 52.55 (+3.73) & 6.37 / 67.95 & $[2^{-8},2^{-7})$: \textbf{98.263\% ($131\times$)} \\
 & RN $\eta{=}2.00{\times}10^{-5}$ & 34.50 / 52.52 & F 4.53 / R 86.94 & 6.29 / 52.84 (+1.76) & 5.55 / 69.78 & $[2^{-8},2^{-7})$: 95.861\% \\
\bottomrule
\end{tabular}}
\end{table}

\end{document}